\documentclass[journal,compsoc]{IEEEtran}
\usepackage[latin9]{inputenc}
\usepackage{color}
\usepackage{array}
\usepackage{verbatim}
\usepackage{float}
\usepackage{units}
\usepackage{textcomp}
\usepackage{multirow}
\usepackage{amsmath}
\usepackage{amssymb}
\usepackage{graphicx}

\makeatletter

\providecommand{\tabularnewline}{\\}
\floatstyle{ruled}
\newfloat{algorithm}{tbp}{loa}
\providecommand{\algorithmname}{Algorithm}
\floatname{algorithm}{\protect\algorithmname}

\ifCLASSOPTIONcompsoc
  \usepackage[nocompress]{cite}
\else
  \usepackage{cite}
\fi
\ifCLASSINFOpdf
\else
\fi
\usepackage[font=footnotesize]{caption}

\usepackage{arydshln}
\@ifundefined{showcaptionsetup}{}{%
 \PassOptionsToPackage{caption=false}{subfig}}
\usepackage{subfig}
\makeatother

\begin{document}
\title{A Bayesian Filter for Multi-view 3D Multi-object\\ 
Tracking with Occlusion Handling}
\author{Jonah Ong, Ba-Tuong Vo, Ba-Ngu Vo, Du Yong Kim and Sven Nordholm
\IEEEcompsocitemizethanks{\IEEEcompsocthanksitem J. Ong, B.T. Vo, B.N. Vo, and S. Nordholm are with the Department of Electrical and Computer Engineering, Curtin University, Bentley, WA 6102, Australia.\protect\\  
% note need leading \protect in front of \\ to get a newline within \thanks as % \\ is fragile and will error, could use \hfil\break instead.
E-mail: \{j.ong1, ba-tuong.vo, ba-ngu.vo, s.nordholm\}@curtin.edu.au
\IEEEcompsocthanksitem D.Y. Kim is with the School of Engineering, RMIT University, Melbourne, Australia.\protect\\ % <-this % stops a space
E-mail: duyong.kim@rmit.edu.au}
}

\IEEEtitleabstractindextext{
\begin{abstract}
This paper proposes an online multi-camera multi-object tracker that only requires monocular detector training, independent of the multi-camera configurations, allowing seamless extension/deletion of cameras  without retraining effort. The proposed algorithm has a linear complexity in the total number of detections across the cameras, and hence scales gracefully with the number of cameras. It operates in the 3D world frame, and provides 3D trajectory estimates of the objects. The key innovation is a high fidelity yet tractable 3D occlusion model, amenable to optimal Bayesian multi-view multi-object filtering, which seamlessly integrates, into a single Bayesian recursion, the sub-tasks of track management, state estimation, clutter rejection, and occlusion/misdetection handling. The proposed algorithm is evaluated on the latest WILDTRACKS dataset, and demonstrated to work in very crowded scenes on a new dataset.
\end{abstract}
}

\maketitle

\section{Introduction }

\IEEEPARstart{T}{he} interest of visual tracking is to jointly estimate
an unknown time-varying number of object trajectories from a stream
of images \cite{poiesi2013multi}. The challenges of visual tracking
are the random appearance/disappearance of the objects, false positives/negatives,
and data association uncertainty \cite{shitrit2014multi}. Multiple
object tracking (MOT) algorithms can operate online to produce current
estimates as data arrives, or in batch which delay the estimation
until further data is available \cite{xu2016multi,kim2019labeled}.
In principle, batch algorithms are more accurate than online as they
allow better data integration into the estimates \cite{berclaz2011multiple,milan2014continuous,shitrit2014multi,wang2016tracking}.
Online algorithms, however, tend to be faster and hence better suited
for time-critical applications \cite{breitenstein2010online,babenko2010robust,hoseinnezhad2012visual,henriques2014high,kim2019labeled}. 

The common sub-tasks, traditionally performed by separate modules
in a MOT system are track management, state estimation, clutter rejection,
and occlusion/misdetection handling. Track management involves the
initiation, termination and identification of trajectories of individual
objects, while state estimation is concerned with determining the
state vectors of the trajectories. Problems such as track loss, track
fragmentation and identity switching are caused by false negatives
that can arise from occlusions when objects of interest are visually
blocked from a sensor, or from misdetections when the sensor/detector
fails to register objects of interest. On the other hand, false positives
can lead to false tracks and identity switching. Hence, occlusion/misdetection
handling and clutter rejection are critical for improving tracking
performance.

While occlusion handling is just as challenging compared with the
other sub-tasks, theoretical developments are far and few \cite{peng2015robust}.
This is due mainly to the complex object-to-object and object-to-background
relationships, as well as computational tractability because, theoretically,
all possible partitions of the set of objects need to be considered
\cite{kim2019labeled}. In a single-view setting, useful \emph{a priori
}information about the objects of interest are exploited to resolve
occlusions \cite{andriyenko2011analytical,milan2014continuous,henriques2014high,shitrit2014multi}.
However, there are fundamental limitations on what can be achieved
with single-view data. In contrast, a multi-view setting naturally
allows exploiting complementary information from the data to resolve
occlusions since an object occluded in one view may not be occluded
in another \cite{dockstader2001multiple}. Furthermore, from an information
theoretic standpoint, data from diverse views will reduce the uncertainty
on the set of objects of interest, thereby improving overall tracking
performance. Given the proliferation of cameras in today's world,
it is imperative to develop effective means for making the best of
the information-rich multi-view data sources, not only for occlusion
handling, but ultimately to achieve better visual tracking. 

\begin{figure}
\subfloat[]{\includegraphics[width=0.44\columnwidth]{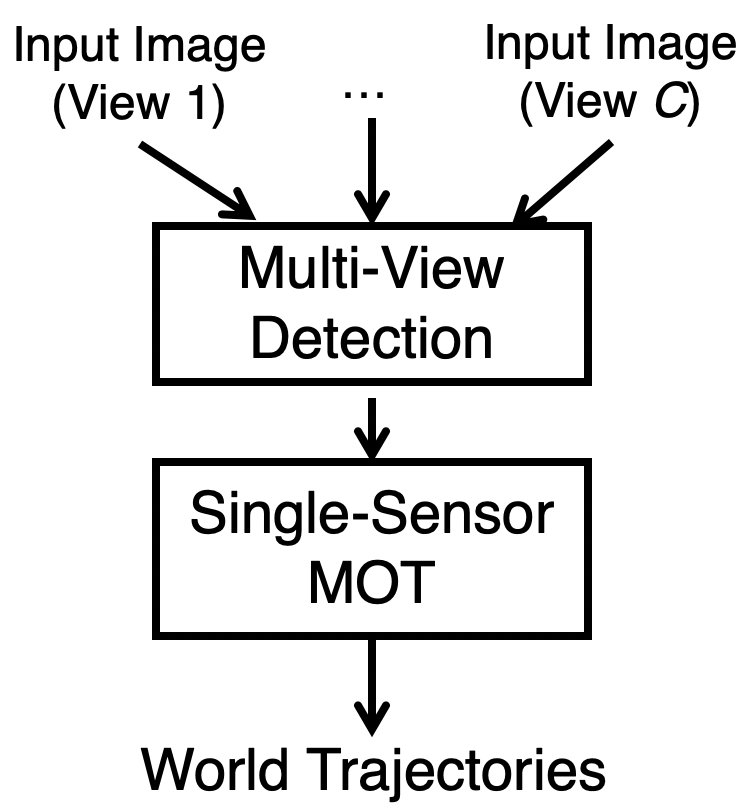}

}\subfloat[]{\includegraphics[width=0.56\columnwidth]{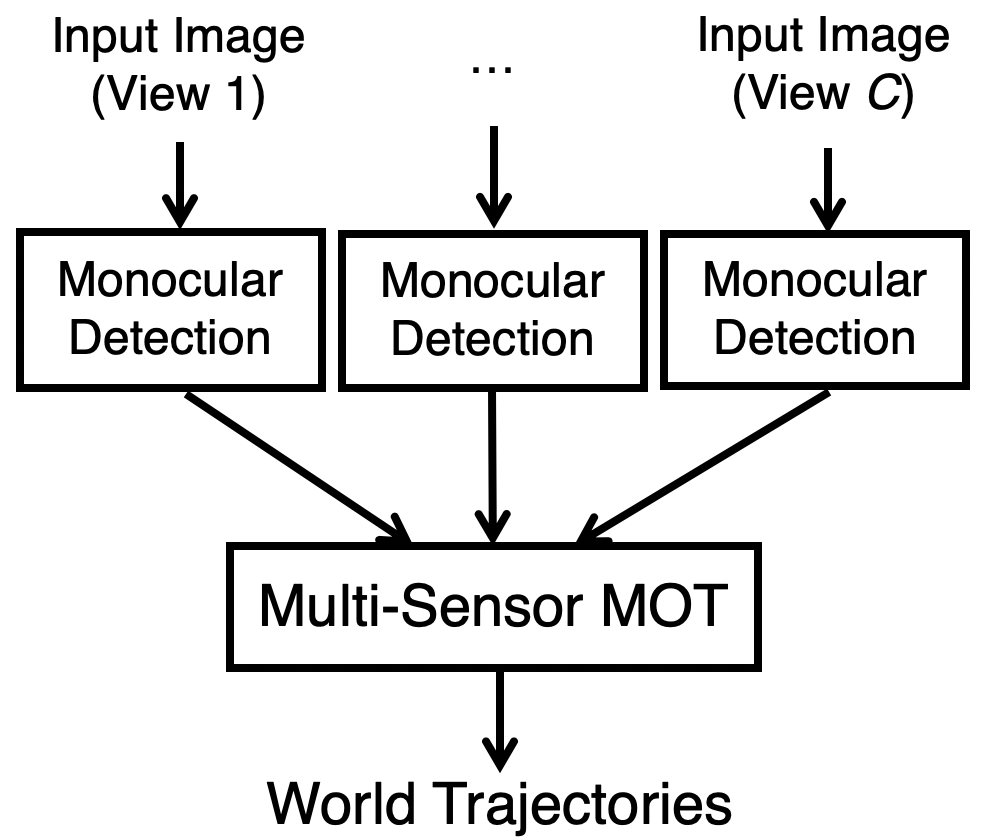}

}

\caption{Multi-view Architectures: (a) Multi-view Detection + Single-sensor
Multi-object Tracking \cite{chavdarova2018wildtrack}; (b) Monocular
Detection + Multi-sensor Multi-object Tracking.}
\label{Fig_1}
\end{figure}

The perennial challenge in multi-view visual MOT is the high-dimensional
data association problem between the detections and objects, across
different views/cameras \cite{fleuret2008multicamera,peng2015robust}.
Two common architectures for multi-view MOT are shown in Fig. \ref{Fig_1}.
So far the best solutions are batch algorithms with the architecture
in Fig. \ref{Fig_1} (a). These solutions are based on: generative
modeling and dynamic programming \cite{fleuret2008multicamera}; convolutional
neural network (CNN) multi-camera detection (MCD), trained on multi-view
datasets \cite{chavdarova2017deep}, followed by track management
\cite{chavdarova2018wildtrack}; and MCD via multi-view CNN training
combined with Conditional Random Fields (CRF) models to exploit multi-camera
geometry (followed by track management) \cite{baque2017deep}. These
MCD based MOT solutions, which produce trajectories on the ground
plane, have been shown to outperform previous works \cite{chavdarova2017deep},
and demonstrated remarkable performance in crowded scenarios \cite{baque2017deep}.
Note that such data-centric MCDs require retraining when the multi-camera
system is extended/reconfigured, and that training/learning is expensive
as the input space is very high-dimensional due to the large number
of possible combinations across the cameras \cite{domke2013learning}.
In practice, it is desirable for a multi-view MOT system to produce
trajectories in 3D world frame, online, and requires no retraining
for multi-camera extension/reconfiguration (including camera failures)
so as to operate uninterrupted. %
\begin{comment}
since the model has to be trained on copious datasets and to As the
computational cost and requirement for the MCD can be quite demanding,
in this paper, however, we aim to achieve the same performance as
the aforementioned approaches with the interest of proposing an online
and efficient framework that casts the problem of multi-view MOT into
a recursive Bayesian paradigm in which every image from each view
is pre-processed independently using state-of-the-art monocular detector.
high dimensional learning... learning in high dimensional space
\end{comment}
{} 

\noindent 

\noindent \vspace*{-0.8cm}

This paper proposes a model-centric, online multi-view visual MOT
solution that only requires monocular detector training, independent
of the multi-camera configurations, via the architecture of Fig. \ref{Fig_1}
(b). Hence, no retraining of the detectors is needed when the multi-camera
system is extended/reconfigured. More importantly, our algorithm has
a linear complexity in the total number of detections, thereby scales
gracefully with the number of cameras. The algorithm intrinsically
operates in the 3D world frame by exploiting multi-camera geometry,
allowing it to track people jumping and falling, suitable for applications
such as sports analytics, age care, school environment monitoring,
etc. We validate the proposed method on the latest WILDTRACKS dataset
on ground plane and show comparable results with Deep-Occlusion+KSP+ptrack
\cite{chavdarova2018wildtrack}. To evaluate tracking performance
in the 3D world frame, we develop a new dataset with varying degrees
of difficulties on scenarios with very closely spaced people, with
addition/deletion of cameras during operation, and with people jumping
and falling.

\begin{figure}
\centering

\includegraphics[width=0.7\columnwidth]{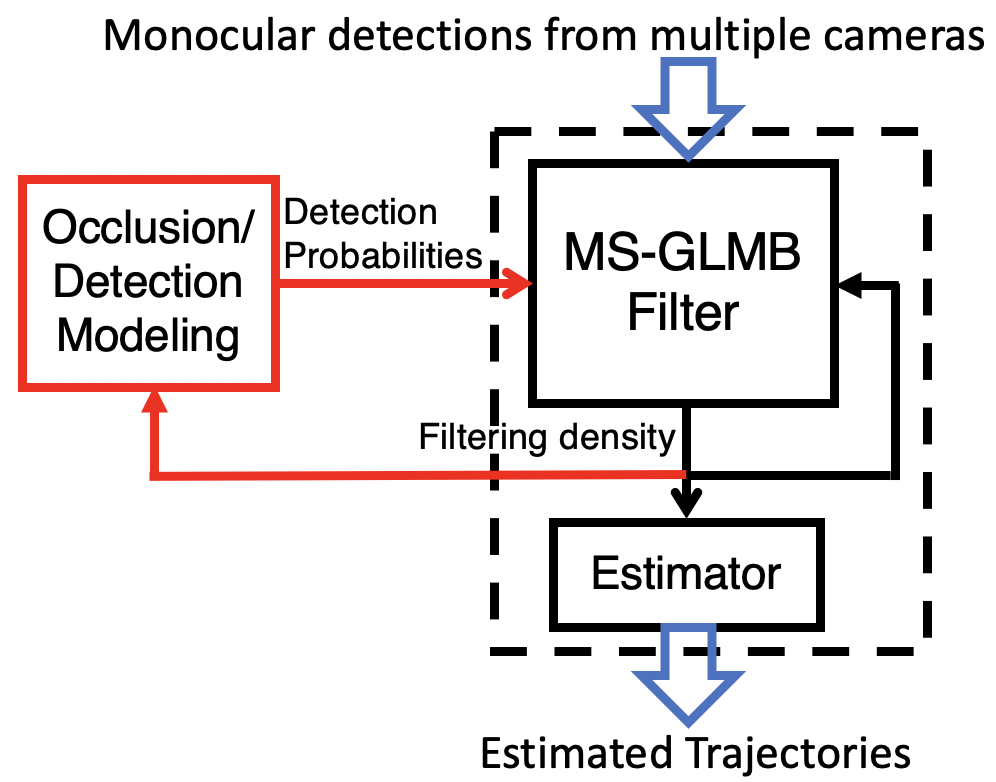}

\caption{MV-GLMB-OC filter Processing Chain. Monocular detections from multiple
cameras are fed into the filter, which outputs the filtering density.
This output is fed into: the estimator to generate track estimates;
and back into the filter to process detections at the next time. The
Occlusion Model (red) is an add-on that takes the filter output and
compute the detection probabilities for the filter on-the-fly.}
\label{MVGLMB_OC_proc_chain}
\end{figure}

The key innovation is a high fidelity yet tractable 3D occlusion model,
amenable to Bayesian multi-sensor multi-object filtering \cite{vo2019multi},
which seamlessly integrates, into a single Bayesian recursion, the
sub-tasks of track management, state estimation, clutter rejection,
and occlusion/misdetection handling.\textcolor{black}{{} In the Bayesian
paradigm, the multi-object filtering density captures all information
on the set of trajectories in 3D, encapsulated in the observations,
as well as dynamic and observation models. The novel occlusion model,
incorporated in the multi-object measurement likelihood function,
enables the MOT Bayesian filter to correctly maintain occluded tracks
that would have otherwise been incorrectly terminated.} The schematic
in Fig. \ref{MVGLMB_OC_proc_chain} shows the integration of the novel
occlusion model into a near-optimal multi-sensor multi-object Bayes
filter known as the Multi-Sensor Generalized Labeled Multi-Bernoulli
(MS-GLMB) filter \cite{vo2019multi}. This configuration enables the
proposed algorithm, herein referred to as Multi-View GLMB with OCclusion
modeling (MV-GLMB-OC), to address occlusions, and inherits the numerical
efficiency of the MS-GLMB filter. %
\begin{comment}
As documented in \cite{peng2015robust}, extracting detections from
all views and projecting them onto the world plane is susceptible
to reprojection errors. We show that collection of detections from
each image is formulated as a vector of sets which is then treated
as observations for an optimal multi-sensor Bayes filter that performs
state-space estimation. The Multi-Sensor Generalized Labeled Multi-Bernoulli
(MS-GLMB) filter is an analytical solution to the optimal multi-sensor
Bayes filter \cite{vo2017multi}. The derivation of the MS-GLMB filter
is justified using the theory of Labeled Random Finite Sets (LRFS)
\cite{vo2013labeled,vo2014labeled} and Finite Set Statistics (FISST)
\cite{mahler2007statistical,mahler2014advances}. The rationale is
to treat a multi-object state that is subjected to appearance and
disappearance as an LRFS such that its cardinality and spatial/kinematic
information are random. The goal is then to obtain the estimates of
the multi-object state over a discrete time step from the \emph{posterior
probability density function }(pdf) using observations from multiple
views that are subjected to reprojection errors, misdetections and
false alarms. 
\end{comment}
{} In short, our main technical contributions are: 
\begin{itemize}
\item A tractable and realistic detection model that accommodates 3D occlusion
by taking into account the Lines of Sights (LoSs) of all objects in
the scene with respect to the cameras. In contrast, conventional detection
models either neglect the LoSs of the objects or are computationally
intractable, leading to poor tracking performance in the presence
of occlusions. Our new detection model can be regarded as a generalization
of tractable conventional detection models;
\item The first Bayesian multi-view MOT filter for such detection model,
which resolves occlusion online and is scalable with the number of
sensors. Experiments show better performance than the latest multi-camera
tracking algorithm; 
\item A new dataset with full 3D annotations (not restricted to the ground
plane), in terms of position and extent in all 3 x, y, z-coordinates,
including sequences that involve changes in the z-coordinate due to
people jumping and falling. %
\begin{comment}
\begin{itemize}
\item A tractable and realistic 3D occlusion model via a probability of
detection profile that involves all current states rather than a single
current state as in existing tracking algorithms. 
\item A new Bayes optimal multi-view tracking filter that is applicable
to such detection model to resolve occlusion online, which shows better
performance than the latest multi-camera tracking algorithm. Existing
trackers from RFS or other frameworks are not applicable to such detection
model. This is the first online multi-camera/sensor tracking algorithm
that addresses occlusion in a principled manner, and is scalable with
the number of sensors. 
\item A new dataset to test the new algorithm on densely populated small
space, which existing datasets are not sufficient. 
\end{itemize}
\end{comment}
{} Instead of reporting performance for the entire scenario duration
(as done traditionally), we also introduce live or online tracking
performance evaluation over time, using the OSPA\textsuperscript{(2)}
metric \cite{beard2020solution}, to characterize the behavior of
the algorithm and demonstrate uninterrupted operation when the multi-camera
system is extended/reconfigured. 
\end{itemize}

The rest of the paper is organized as follows. Section \ref{sec:Related-Work}
presents the related work. Section \ref{sec:Bayesian-Formulation}
formulates the multi-view MOT problem, including the proposed occlusion/detection
model, and the new tractable filter with occlusion handling capability
via optimal Bayesian estimation. Section \ref{sec:Implementation}
presents the implementation of the algorithm. Section \ref{sec:Experimental-Results}
shows experimental results and discussions. Finally, some conclusions
are drawn in Section \ref{sec:Conclusion}.%

\section{Related Work \label{sec:Related-Work}}

A Deep Convolutional Neural Network (CNN) trained on large-scale high-resolution
image dataset, with efficient implementations such as Fast/Faster
R-CNN \cite{girshick2015fast,ren2015faster}, has been shown to outperform
all previous object detectors based on hand-engineered features, e.g.
the Aggregated Channel Features (ACF) object detector \cite{krizhevsky2012imagenet}.
Faster R-CNN introduces the concept of Region Proposal Network (RPN)
and exploits feature sharing together with efficient multi-scale solution
to improve test-time speed and detection accuracy, achieving real-time
detection at 5 frames-per-second (fps) \cite{ren2015faster}. Recently,
the You Only Look Once (YOLO) real-time object detector, which attains
40fps at mAP of 76.8\% (resolution of 544x544) on PASCAL VOC 2007,
has gained immense popularity \cite{redmon2017yolo9000}. In contrast
to the aforementioned techniques that rely on a sliding classifier
for every image, YOLO's impressive speed is achieved by only scanning
the image once. Additionally, spatial constraints, introduced to eliminate
unlikely bounding boxes, allow trade-offs between speed and accuracy
via a suitable score threshold \cite{redmon2016you}. The YOLO detector
can also be extended to 3D \cite{zhao2018novel}. The main drawback
is the inability to detect small objects due to the imposed spatial
constraints \cite{redmon2016you}. 

Progress in object detections facilitated the development of many
tracking-by-detection approaches that typically join the detections
together to form consistent trajectories \cite{andriluka2008people,breitenstein2010online,smeulders2013visual}.
Tracking-by-detection can be designed for batch or online operations.
Online algorithms tend to be faster and better suited for time-critical
applications, but may be prone to irrevocable errors if objects are
undetected in several frames or if detections at different times are
incorrectly joined \cite{shitrit2014multi}. Such errors can be reduced
by global trajectory optimization over batches of frames \cite{berclaz2011multiple,milan2014continuous,shitrit2014multi,wang2016tracking,xu2016multi}.
However, track loss and fragmentation can still be caused by occlusion,
which is an active area of research in itself \cite{andriluka2008people}.
In single-view/monocular settings, a popular approach to occlusion
handling is to exploit \emph{a priori }knowledge of the scene \cite{milan2014continuous,shitrit2014multi,wang2016tracking}.
Deep neural network techniques that leverage spatio-temporal information
in the images have shown to perform well in autonomous driving \cite{weng2020gnn3dmot,liang2020pnpnet}. 

In a multi-view setting, complementary information from the data can
be exploited to resolve occlusions naturally, since an object occluded
in one view may not be occluded in another view \cite{dockstader2001multiple}.
The hierarchical composition approach in \cite{xu2016multi} uses
monocular information from multiple views to construct estimates in
the ground plane. However, this approach is susceptible to reprojection
errors and ignores occlusions \cite{baque2017deep}. In \cite{otsuka2004multiview},
the author formulates an occlusion model based on 2D silhouette-based
visual angles from multiple views. %
\begin{comment}
In principle, multi-view approaches can resolve single-view detection
ambiguities, provided that the multi-stream information is utilized
jointly to produce detections.
\end{comment}
Subsequently, a simple approach is to pre-process images from individual
views (e.g. via background subtraction) from which occupancy (on the
ground plane) can be estimated using Probability of Occupancy Map
(POM) \cite{fleuret2008multicamera}. A more sophisticated approach
was proposed in \cite{peng2015robust}, which combines multi-view
Bayesian network modeling of occlusion relationship and homography
correspondence, across all views, with height-adaptive projection
(HAP) to obtain final ground plane detections \cite{peng2015robust}.
Stereo-based MOT approaches have also demonstrated improved 3D object
estimation and tracking \cite{osep2017combined,li2020joint,pedersen20203d}. 

So far, the best multi-view tracking solution is based on a multi-camera
detection (MCD) architecture that uses a CNN to train multi-view detectors
from monocular and multi-view data \cite{chavdarova2017deep}, together
with batch processing to compute global trajectories on the ground
plane \cite{chavdarova2018wildtrack}. Combined with Conditional Random
Field (CRF) modeling and Mean Field variational inference, this approach
achieves remarkable performance in crowded scenarios \cite{baque2017deep}.
This approach is more data-centric than model-centric as the multi-camera
detection relies mostly on training from data. Hence, large training
sets are required, and the learning algorithm tends to be computationally
expensive in exploring tight convergence levels, especially for high
dimensional scenarios (e.g. large number of cameras) \cite{domke2013learning}.
More examples of deeply learned multi-view approaches are found in
\cite{frossard2018end,zhang2019robust}. To the best of our knowledge,
no online MOT algorithm has produced comparable tracking performance
with these data-centric batch solutions.

In practice, it is desirable to have online algorithms whose complexity
scale linearly with the number of cameras, and do not require multi-view
training so that reconfiguration (including addition and deletion)
of cameras can be performed without interruption to the operation.
Moreover, in a multi-view context, it is more prudent to have trajectories
in the 3D world frame for applications such as sports analytics, age
care, school environment monitoring, etc. While there are solutions
to online 3D multi-view MOT with monocular data such as \cite{scheidegger2018mono,hu2019joint},
they do not scale gracefully with the number of cameras. Similar to
the mentioned batch-processing methods, these solutions are more data-centric
as they rely, respectively, on deep training for object depth information,
and motion learning. 

At the other end of the spectrum are the model-centric approaches
that rely largely on physical models of the dynamics of the objects,
the geometry and characteristics of the sensors/cameras. Such model-based
solutions to 3D online MOT with monocular data, using 2D object detections,
3D object proposals, and 3D point cloud techniques were developed,
respectively, in \cite{leibe2008coupled,osep2017combined,weng2019baseline}.
From a state-space modeling perspective, a natural choice for online
MOT is the multi-object Bayes filter \cite{mahler2003multitarget}.
Since the inception of the Random Finite Sets (RFS) framework for
multi-object state-space models, a number of multi-objects Bayesian
filters have been developed \cite{mahler2014advances,mahler2007statistical}
and applied to visual MOT problems \cite{maggio2008efficient,hoseinnezhad2012visual,kim2019labeled}.
The latest is the Generalized labeled Multi-Bernoulli (GLMB) filter,
an analytic solution to the multi-object Bayes filter that jointly
estimates the number of objects and their trajectories online \cite{vo2013labeled}.
The salient feature of this approach is that it seamlessly integrates
track management, state estimation, clutter rejection, occlusion/misdetection
handling and multiple sensor data into a single recursion \cite{kim2019labeled}.
In this article, we use this framework to develop an online 3D multi-view
MOT solution that only requires one-off monocular detector training
(or off-the-shelf monocular detectors), yet is capable of producing
comparable results with the aforementioned data-centric batch-processing
approaches.

In addition to algorithms, datasets for performance evaluation are
an important aspect of 3D multi-view MOT research. Existing multi-view
datasets include DukeMTMC \cite{ristani2016performance}, PETS 2009
S2.L1 \cite{ferryman2009pets2009}, EPFL - Laboratory, Terrace and
Passageway \cite{fleuret2008multicamera}, SALSA \cite{alameda2015salsa},
Campus \cite{xu2016multi} and EPFL-RLC \cite{chavdarova2017deep}.
However, in \cite{chavdarova2018wildtrack} the authors discussed
a number of their shortcomings and introduced a seven-camera high-definition
(HD) unscripted pedestrian dataset known as WILDTRACKS to provide
a high quality, highly crowded and cluttered evaluation scenario.
It comes with accurate joint (extrinsic and intrinsic) calibration,
and 7 series of 400 annotated frames for detection at a rate of 2
frames per second (fps). The annotations of the tracks are given both
as locations on the ground plane and 2D bounding boxes projected onto
each view. 

While WILDTRACKS is more extensive than earlier datasets, it is still
not sufficient for comprehensive 3D MOT performance evaluation. Specifically,
for actual 3D MOT applications where objects may also move vertically
(e.g. sport analytics, age care, etc.), ground plane annotations are
simply not adequate for evaluating tracking performance in full 3D,
i.e. changes in all 3 x, y, z-coordinates. To enrich the datasets
and to enable performance evaluation in full 3D, we propose the Curtin
Multi-Camera (CMC) dataset that comprises four calibrated cameras,
on scenarios of varying difficulties in crowd density and occlusion,
as well as scenarios with people jumping and falling, all with 3D
centroid-with-extent annotations, along with camera locations and
parameters. Note that in addition to extrinsic and intrinsic parameters,
we also provide the absolute camera locations needed for testing and
evaluation of model-centric solutions that exploit multi-camera geometry. 

\section{Bayesian Formulation \label{sec:Bayesian-Formulation}}

This section formulates the multi-view MOT problem (Sections \ref{subsec:Bayes-Filter}-\ref{subsec:Multi-Sensor-GLMB-Filter}),
including the proposed occlusion/detection model (Section \ref{subsec:Detection-Model}),
and the new tractable filter with occlusion handling capability (Section
\ref{subsec:Multi-view-GLMB-Update}). The notations used in this
paper are tabulated in Table \ref{notations}. 
\begin{table}
\caption{Basic Notation}
 \label{notations}

\begin{tabular}{|c|l|}
\hline 
{\footnotesize{}Symbol} & {\footnotesize{}Description}\tabularnewline
\hline 
{\footnotesize{}$a^{T}$} & {\footnotesize{}Transpose of vector/matrix $a$}\tabularnewline
{\footnotesize{}$\otimes$} & {\footnotesize{}Kronecker product (for matrices)}\tabularnewline
{\footnotesize{}$\mathrm{I}_{n}$} & {\footnotesize{}$n$-dimensional identity matrix}\tabularnewline
{\footnotesize{}$0_{n\times m}$} & \emph{\footnotesize{}n}{\footnotesize{} by }\emph{\footnotesize{}m}{\footnotesize{}
zero matrix }\tabularnewline
{\footnotesize{}$\mathrm{diag}(\cdot)$} & {\footnotesize{}Converts a vector to a diagonal matrix}\tabularnewline
{\footnotesize{}$X_{m:n}$} & {\footnotesize{}$X_{m},X_{m+1},\dots,X_{n}$}\tabularnewline
{\footnotesize{}$\langle f,g\rangle$} & {\footnotesize{}$\int f(x)g(x)dx$ }%
\begin{comment}
{\footnotesize{}where $f$ and $g$ are any functions that take $x\in\mathbb{R}$}
\end{comment}
\tabularnewline
{\footnotesize{}$h^{X}$} & {\footnotesize{}$\prod\limits _{x\in X}h(x)$ where $h^{\emptyset}=1$
}%
\begin{comment}
{\footnotesize{}$h$ is any real-valued function with $h^{\emptyset}=1$
by convention}
\end{comment}
\tabularnewline
{\footnotesize{}$\delta_{Y}[X]$} & {\footnotesize{}Kronecker delta function: $1$ if $X=Y$, $0$ otherwise}\tabularnewline
{\footnotesize{}$1_{Y}(x)$} & {\footnotesize{}Indicator function: $1$ if $x\in Y$, $0$ otherwise}\tabularnewline
{\footnotesize{}$\mathcal{N}(\:\cdot\:;\mu,P)$} & {\footnotesize{}Gaussian }\emph{\footnotesize{}pdf}{\footnotesize{}
with mean $\mu$ and covariance $P$}\tabularnewline
\hline 
\end{tabular}
\end{table}

\subsection{Bayes Filter \label{subsec:Bayes-Filter}}

We first recall the classical Bayesian filter where the state $x$
of the object, in some finite dimensional state space $\mathbb{X}$,
is modeled as a random vector. The dynamic of the state is described
by a Markov chain with transition density $f_{+}(x_{+}|x)$, i.e.
the probability density of a transition to the state $x_{+}$ at the
next time given the current state $x$. Note that for simplicity we
omit the subscript for current time and use the subscript `+' denotes
the next time step. Additionally, the current state $x$ generates
an observation $z$ described by the likelihood function $g(z|x)$,
i.e. the probability density of receiving the observation $z$ given
$x$. All information on the current the state is encapsulated in
the filtering density\footnote{The filtering densities are conditioned on the observations, which
have been omitted for notational compactness\emph{.}} $p$, which can be propagated to the next time as $p_{+}$, via the
celebrated Bayes recursion \cite{ristic2003beyond}
\begin{equation}
p_{+}\!\left(x_{+}\right)\propto g\left(z_{+}|x_{+}\right)\int f_{+}\!\left(x_{+}|x\right)p\left(x\right)dx.
\end{equation}

The multi-view MOT Bayes filter used in this work is conceptually
identical to the classical Bayes filter above by replacing: $x$ and
$x_{+}$ with the sets $\boldsymbol{X}$ and $\boldsymbol{X}_{+}$;
$p$ and $p_{+}$ with the multi-object filtering densities $\mathbf{\boldsymbol{\pi}}$
and $\mathbf{\boldsymbol{\pi}_{+}}$; $f_{+}$ and $g$ with the multi-object
transition density $\boldsymbol{f}_{+}$ and multi-object observation
likelihood $\boldsymbol{g}$; $z_{+}$ with the observation set $Z_{+}$;
and the integral with the set integral \cite{mahler2014advances},
i.e.
\begin{equation}
\boldsymbol{\pi}_{+}\!\left(\boldsymbol{X}_{+}\right)\propto\boldsymbol{g}\left(Z_{+}|\boldsymbol{\boldsymbol{X}}_{+}\right)\int\boldsymbol{f}_{+}\!\left(\boldsymbol{X}_{+}|\boldsymbol{X}\right)\boldsymbol{\pi}\left(\boldsymbol{X}\right)\delta\boldsymbol{X}.\label{e:MTBF}
\end{equation}
The sets $\boldsymbol{X}$ (and $\boldsymbol{X}_{+}$) containing
the object states at the current (and next) time, is called the current
(and next) multi-object state. Each element of the multi-object state
$\boldsymbol{X}$ is an ordered pair $\boldsymbol{x}=(x,\ell)$, where
$x\in\mathbb{X}$ is a state vector, and $\ell\triangleq(t,\alpha)$
is a unique label consisting of the object's time of birth $t$, and
an index $\alpha$ to distinguish those born at the same time \cite{vo2013labeled}.
The cardinality (number of elements) of $\boldsymbol{X}$ and $\boldsymbol{X}_{+}$
may differ due to the appearance and disappearance of objects from
one frame to the next. 

Under the Bayesian paradigm, the multi-object state is modeled as
a random finite set, i.e. a finite-set-valued random variable, characterized
by Mahler's multi-object density \cite{mahler2007statistical,mahler2014advances}
(equivalent to a probability density \cite{vo2005sequential}). The
multi-object transition density $\boldsymbol{f}_{+}$ captures the
motions as well as births and deaths of objects. The multi-object
observation likelihood $\boldsymbol{g}$ captures the detections,
false alarms, occlusions, and misdetections.

\subsection{Motion and Birth/Death Models \label{subsec:Object-Dynamics}}

An object at time $k$, represented by a state $\boldsymbol{x}=(x,\ell)$,
either survives with probability $P_{S}(\boldsymbol{x})$ and evolves
to state $\boldsymbol{x}_{+}=(x_{+},\ell_{+})$ at the next time with
transition density
\begin{equation}
\boldsymbol{f}_{S,+}(\boldsymbol{x}_{+}|\boldsymbol{x})=f_{S,+}(x_{+}|x,\ell)\delta_{\ell}[\ell_{+}],
\end{equation}
or dies with probability $1-P_{S}(\boldsymbol{x})$ \cite{vo2013labeled}.
At this next time, an object with label $\mathbb{\ell}$ is born with
probability $P_{B,+}(\ell)$, and with feature-vector $x$ distributed
according to a probability density $f_{B,+}(\cdot,\ell)$. Note that
the label of an object remains the same over time, and hence the \emph{trajectory}
of an object is a sequence of consecutive states with a common label
\cite{vo2013labeled}.

Let $\mathbb{B}_{k}$ denote the finite set of all possible labels
for objects born at time $k$, then the label space for all objects
up to time $k$ is the disjoint union $\mathbb{L}_{k}=\biguplus_{t=0}^{k}\mathbb{B}_{t}$.
For simplicity we omit the time subscript $k$, and let $\mathcal{L}\left(\boldsymbol{x}\right)$
denote the label of an $\boldsymbol{x\in}\mathbb{X}\times\mathbb{L}$.
For any finite $\boldsymbol{X\subset}\mathbb{X}\times\mathbb{L}$,
we define $\mathcal{L}\left(\boldsymbol{X}\right)\triangleq\left\{ \mathcal{L}\left(\boldsymbol{x}\right):\boldsymbol{x}\in\boldsymbol{X}\right\} $,
and the \emph{distinct label indicator} $\Delta\left(\boldsymbol{X}\right)\triangleq\delta_{\left|\boldsymbol{X}\right|}\left[\left|\mathcal{L}\left(\boldsymbol{X}\right)\right|\right]$.
At any time, the set $\boldsymbol{X}$ of (states of) objects in the
scene must have distinct labels, i.e. $\Delta\left(\boldsymbol{X}\right)=1$.
Conditional on the current set of objects, it is standard practice
to assume that objects are born or displaced at the next time, independently
of one another. The expression for the multi-object transition density
$\boldsymbol{f}_{+}$ is not needed in this work, interested readers
are referred to \cite{vo2013labeled}. 

\subsection{Multi-Sensor Observation Model \label{subsec:Multi-Sensor-Observations}}

Suppose that at time $k$, there are $C$ cameras (sensors), and a
set $\boldsymbol{X}$ of current objects. Each $\boldsymbol{x}\in\boldsymbol{X}$
is either: detected by camera $c\in\{1{\textstyle :}C\}$, with probability
$P_{D}^{\left(c\right)}\left(\boldsymbol{x};\boldsymbol{X}\!-\!\{\boldsymbol{x}\}\right)$
and generates an observation $z^{\left(c\right)}$ in the measurement
space $\mathbb{Z}^{(c)}$ with likelihood $g^{\left(c\right)}(z^{\left(c\right)}|\boldsymbol{x})$;
or missed with probability $1-P_{D}^{\left(c\right)}\left(\boldsymbol{x};\boldsymbol{X}\!-\!\{\boldsymbol{x}\}\right)$.
Note that to account for occlusions (and uncertainty in the detection
process), the probability of detecting an object $\boldsymbol{x}$
also depends on the states of other current objects $\boldsymbol{X}\!-\!\{\boldsymbol{x}\}$.
However, most MOT algorithms neglect this dependence for computational
tractability. 

The detection process also generates false positives at camera $c$,
usually characterized by an intensity function $\kappa^{\left(c\right)}$
on $\mathbb{Z}^{(c)}$. The standard model is a Poisson distribution,
with mean $\langle\kappa^{\left(c\right)},1\rangle$, for the number
of false positives, and the false positives themselves are i.i.d.
according to the probability density $\kappa^{\left(c\right)}/\langle\kappa^{\left(c\right)},1\rangle$
\cite{bar1990tracking,blackman1999design,mahler2007statistical}.
Moreover, conditional on the set $\boldsymbol{X}$ of objects, detections
are assumed to be independent from false positives, and that the set
$Z^{(c)}$ of detections and false positives at sensor $c$, are independent
from those at other sensors.

An association hypothesis (at time $k$) associating labels with detections
from camera $c$ is a mapping $\gamma^{\left(c\right)}\!\!:\!\mathbb{L}\!\rightarrow\!\{-\!1{\textstyle :}|Z^{\left(c\right)}|\}$,
such that \emph{no two distinct arguments are mapped to the same positive
value} \cite{vo2013labeled}. This property ensures each detection
comes from at most one object. Given an association hypothesis $\gamma^{\left(c\right)}$:
$\gamma^{\left(c\right)}(\ell)=-1$ means object $\ell$ does not
exist; $\gamma^{\left(c\right)}(\ell)=0$ means object $\ell$ is
not detected by camera $c$; $\gamma^{\left(c\right)}(\ell)>0$ means
object $\ell$ generates detection $z_{\gamma^{(c)}\left(\ell\right)}$
at camera $c$; and the set $\mathcal{L}(\gamma^{\left(c\right)})\triangleq\{\ell\in\mathbb{L}:\gamma^{\left(c\right)}(\ell)\geq0\}$
are the \textit{live labels} of $\gamma^{\left(c\right)}$. Under
standard assumptions, the (multi-object) likelihood for camera $c$
is given by the following sum over the space $\Gamma^{\left(c\right)}$
of association hypotheses with domain $\mathbb{L}$ and range $\{-1{\textstyle :}|Z^{\left(c\right)}|\}$
\cite{vo2013labeled}:
\begin{equation}
\boldsymbol{g}^{(c)\!}(Z^{\left(c\right)}|\boldsymbol{X})\propto\!\!\sum_{\gamma^{\left(c\right)}\in\Gamma^{\left(c\right)}}\!\!\delta_{\mathcal{L}(\gamma^{\left(c\right)\!})}[\mathcal{L}\left(\boldsymbol{X}\right)]\!\left[\psi_{\boldsymbol{X}\!-\{\cdot\}}^{(c,\gamma^{\left(c\right)})}\!\left(\cdot\right)\right]^{\!\boldsymbol{X}}\!\!,\!\!\label{e:Single_Sensor_Lkhd}
\end{equation}
where $Z^{\left(c\right)}=\{z_{1:|Z^{\left(c\right)}|}^{(c)}\}$,
and 
\begin{align}
\!\psi_{\boldsymbol{X}\!-\{\boldsymbol{x}\}}^{(c,\gamma^{\left(c\right)})}\!\negthinspace\left(\boldsymbol{x}\right)\negthinspace=\negthinspace & \begin{cases}
\!1-P_{D}^{\left(c\right)}\!\left(\boldsymbol{x};\boldsymbol{X}\!-\!\{\boldsymbol{x}\}\right)\!,\!\!\!\! & \negthinspace\!\!\!\!\gamma^{\left(c\right)}(\mathcal{L}(\boldsymbol{x}))\negthinspace=\negthinspace0\\
\!\frac{P_{D}^{\left(c\right)\!}\left(\boldsymbol{x};\boldsymbol{X}\!-\{\boldsymbol{x}\}\right)g^{\left(c\right)\!}(z_{j}^{(c)}|\boldsymbol{x})}{\kappa^{\left(c\right)}(z_{j}^{(c)})}\!, & \negthinspace\!\!\!\!\gamma^{\left(c\right)}(\mathcal{L}(\boldsymbol{x}))\negthinspace=\negthinspace j\negthinspace\!>\negthinspace0
\end{cases}\negthinspace,\!\!\label{e:Psi}
\end{align}
\textit{\emph{Note that }}$\psi_{\boldsymbol{X}\!-\{\boldsymbol{x}\}}^{(c,\gamma^{\left(c\right)})}\!\left(\boldsymbol{x}\right)$\textit{\emph{
also depends on }}$Z^{\left(c\right)}$\textit{\emph{, but we omitted
it for clarity. Interested readers are referred to}} the texts \cite{mahler2007statistical,mahler2014advances}\textit{\emph{
for the derivation/discussion.}}

\textit{\emph{A multi-sensor }}(association) hypothesis is an array
$\gamma\triangleq(\gamma^{\left(1\right)},...,\gamma^{\left(C\right)})$
of association hypotheses with the same set of live labels, denoted
as $\mathcal{L}(\gamma)$. The likelihood that $\boldsymbol{X}$ generates
the multi-sensor observation $Z\triangleq(Z^{\left(1:C\right)})$
is the product $\prod_{c=1}^{C}\boldsymbol{g}^{\left(c\right)}\!(Z^{\left(c\right)}|\boldsymbol{X})$,
which can be rewritten as \cite{vo2019multi} 
\begin{equation}
\boldsymbol{g}\left(Z|\boldsymbol{X}\right)\propto\sum_{\gamma\in\Gamma}\delta_{\mathcal{L}(\gamma)}[\mathcal{L}\left(\boldsymbol{X}\right)]\!\left[\psi_{\boldsymbol{X}\!-\{\cdot\}}^{(\gamma)}\!\left(\cdot\right)\right]^{\boldsymbol{X}},\label{e:Multi_Sensor_Lkhd}
\end{equation}
where $\Gamma$ is the set of all \textit{\emph{multi-sensor}} hypotheses,
\allowdisplaybreaks
\begin{align}
\delta_{\mathcal{L}(\gamma)}[J] & \triangleq\prod\limits _{c=1}^{C}\delta_{\mathcal{L}(\gamma^{\left(c\right)})}[J],\label{e:MS_Lkhd_Abbrev_5-1}\\
\psi_{\boldsymbol{X}\!-\{\boldsymbol{x}\}}^{(\gamma)}\left(\boldsymbol{x}\right) & \triangleq\prod\limits _{c=1}^{C}\psi_{\boldsymbol{X}\!-\{\boldsymbol{x}\}}^{(c,\gamma^{\left(c\right)})}\left(\boldsymbol{x}\right).\label{e:MS_Lkhd_Abbrev_6-1}
\end{align}

Remark: The sets of objects, observations, and possibly the number
of sensors and their parameters, may vary with time. However, for
clarity we suppressed the time index.

\subsection{Multi-Sensor GLMB Filter \label{subsec:Multi-Sensor-GLMB-Filter}}

Most of the literature on tracking assumes the probability of detection
$P_{D}^{\left(c\right)}\left(\boldsymbol{x};\boldsymbol{X}\!-\!\{\boldsymbol{x}\}\right)=P_{D}^{\left(c\right)}\left(\boldsymbol{x}\right)$,
i.e. independent of $\boldsymbol{X}\!-\!\{\boldsymbol{x}\}$. In this
case, the Bayes recursion (\ref{e:MTBF}) admits an analytical solution
based on Generalized Labeled Multi-Bernoulli (GLMB) models.

A GLMB is a multi-object density of the form \cite{vo2013labeled}
\begin{equation}
\boldsymbol{\pi}\left(\boldsymbol{X}\right)=\Delta\left(\boldsymbol{X}\right)\sum_{I,\xi}w^{\left(I,\xi\right)}\delta_{I}[\mathcal{L}\left(\boldsymbol{X}\right)]\left[p^{(\xi)}\right]^{\boldsymbol{X}},\label{e:GLMB-1}
\end{equation}
where: $I\in\mathcal{F}(\mathbb{L})$ the space of all finite subsets
of $\mathbb{L}$; $\xi\in\Xi$ the space of all (multi-sensor) association
hypotheses histories up to the current time, i.e. $\xi\triangleq\gamma_{1:k}$;
each $w^{\left(I,\xi\right)}$ is a non-negative weight such that
$\sum_{I,\xi}w^{\left(I,\xi\right)}=1$; and each $p^{\left(\xi\right)}\left(\cdot,\ell\right)$
is a probability density on $\mathbb{X}$. For convenience, we represent
a GLMB by its parameter-set
\begin{equation}
\boldsymbol{\pi}\triangleq\left\{ \left(w^{\left(I,\xi\right)},p^{\left(\xi\right)}\right):\left(I,\xi\right)\in\mathcal{F}(\mathbb{L})\times\Xi\right\} .\label{e:GLMB-(parameter set)}
\end{equation}
Each GLMB \textit{component} $\left(I,\xi\right)$ can be interpreted
as a hypothesis with probability $w^{\left(I,\xi\right)}$, and each
individual object $\ell\in I$ of this hypothesis has probability
density $p^{\left(\xi\right)}\left(\cdot,\ell\right)$.

A simple multi-object state estimate can be obtained from a GLMB by
first determining: the most probable cardinality $n^{*}$ from the
cardinality distribution \cite{vo2013labeled}

\begin{equation}
\mathrm{Prob}(|\boldsymbol{X}|=n)=\sum_{I,\xi}\delta_{n}[|I|]w^{(I,\xi)};
\end{equation}

\noindent and then the hypothesis $(I^{*},\xi^{*})$ with highest
weight such that $|I^{*}|=n^{*}$. The current state estimate for
each object $\ell\in I^{*}$ can be computed from $p^{(\xi^{*})}(\cdot,\ell)$,
e.g. the mode or mean. Alternatively, the entire trajectory of object
$\ell\in I^{*}$ can be estimated using the forward-backward algorithm,
starting from its current filtering density $p^{(\xi^{*})}(\cdot,\ell)$
and propagating backward to its time of birth \cite{nguyen2019glmb,vo2019multi}. 

Under the Bayes recursion (\ref{e:MTBF}), and the standard multi-object
model (i.e. with no occlusions, $P_{D}^{\left(c\right)}\left(\boldsymbol{x};\boldsymbol{X}\!-\!\{\boldsymbol{x}\}\right)=P_{D}^{\left(c\right)}\left(\boldsymbol{x}\right)$),
the multi-object filtering density at any time is a GLMB \cite{vo2013labeled}.
Moreover, if \eqref{e:GLMB-(parameter set)} is the current GLMB filtering
density, then the next GLMB filtering density
\begin{equation}
\boldsymbol{\pi}_{+}=\left\{ \left(w_{+}^{\left(I_{+},\xi_{+}\right)},p_{+}^{\left(\xi_{+}\right)}\right)\!:\left(I_{+},\xi_{+}\right)\in\mathcal{F}(\mathbb{L_{+}})\times\Xi_{+}\right\} ,\label{e:GLMB_next_time}
\end{equation}
can be computed via the\textit{ MS-GLMB recursion }\cite{vo2019multi}
\begin{equation}
\boldsymbol{\pi}_{+}=\Omega\left(\boldsymbol{\pi};P_{D,+}\right),\label{e:GLMB_operator}
\end{equation}
where $P_{D,+}\triangleq(P_{D,+}^{(1)},...,P_{D,+}^{(C)})$. The actual
mathematical expressions for the recursion operator\textit{ $\Omega:\boldsymbol{\pi}\mapsto\boldsymbol{\pi}_{+}$}
are not critical for our arguments, and hence omitted from this section.
Nonetheless, for completeness the definition of\textit{ $\Omega$}
is provided in Appendix \ref{subsec:MS-GLMB-recursion}. Note that\textit{
$\Omega$} also depends on the measurement $Z_{+}$, and model parameters
for birth $(P_{B,+},f_{B,+})$, death/survival $P_{S}$, motion $\boldsymbol{f}_{S,+}$,
false alarms $\kappa_{+}\triangleq(\kappa_{+}^{\left(1\right)},\ldots,\kappa_{+}^{\left(C\right)}),$
and detection $g_{+}\triangleq(g_{+}^{\left(1\right)},\ldots,g_{+}^{\left(C\right)})$
(described in Section \ref{subsec:Multi-Sensor-Observations}). However,
for our purpose it suffices to show the dependence on detection probabilities. 

While the MS-GLMB filter can applied directly to multi-view MOT, a
detection probability (of an object $\boldsymbol{x}$) that does not
depend on other objects, i.e. $\boldsymbol{X}-\{\boldsymbol{x}\}$,
is unable to capture the effect of occlusions. On the other hand,
accounting for occlusions with $P_{D}^{\left(c\right)}\left(\boldsymbol{x};\boldsymbol{X}\!-\!\{\boldsymbol{x}\}\right)$
that actually depends on $\boldsymbol{X}-\{\boldsymbol{x}\}$, results
in filtering densities that are not GLMBs. One example is the merged-measurement
model \cite{beard2015bayesian}, which involves summing over all partitions
of the set $\boldsymbol{X}$, making it intractable \cite{beard2015bayesian}.
Although the resulting filtering density can be approximated by a
GLMB, this solution is still computationally demanding and not suitable
for large number of objects \cite{beard2015bayesian}. In what follows,
we propose a new detection model that addresses occlusions and permits
efficient multi-view MOT implementations. 

\subsection{Detection Model with Occlusion\label{subsec:Detection-Model}}

For tracking in 3D, we consider the state $\boldsymbol{x}=(x,\ell)$,
where:
\begin{equation}
x=(x^{(p)},\dot{x}^{(p)},x^{(s)});\label{e:featurevector}
\end{equation}
$x^{(p)}$ is the object's position (centroid) in 3D Cartesian coordinates;
$\dot{x}^{(p)}$ is its velocity; and $x^{(s)}$ is its shape parameter.
The region in $\mathbb{R}^{3}$ occupied by an object with labeled
state $\boldsymbol{x}$ is denoted by $R(\boldsymbol{x})$. 

\begin{figure}[H]
\centering

\includegraphics[width=0.95\columnwidth]{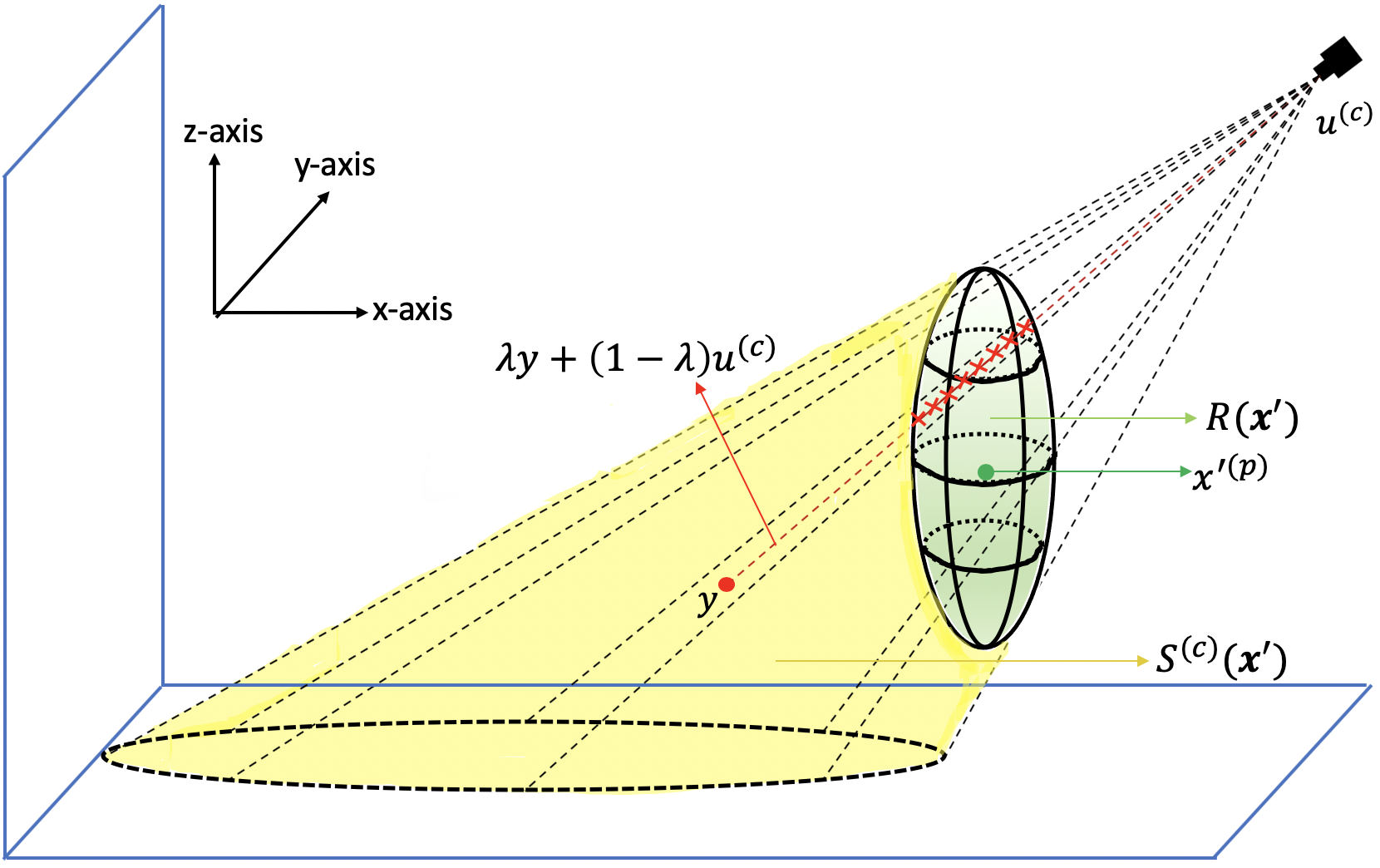}

\caption{The shadow region (in yellow) of object with labeled state $\boldsymbol{x}'$,
relative to camera $c$. }

\label{3d_pd_model}

\medskip{}
\end{figure}

Consider camera $c$ and the set $\boldsymbol{X}$ of current objects.
In this work, an object $(x,\ell)\!\in\!\boldsymbol{X}$ is regarded
as occluded from camera $c$ when its position $x^{(p)}$ is not in
the line of sight (LoS) of the camera, i.e. $x^{(p)}$ is in the \textit{shadow
regions} of the other objects in $\boldsymbol{X}$. Assuming straight
LoSs, the shadow region of an object with labeled state $\boldsymbol{x}'$,
relative to camera $c$ (see Fig. \ref{3d_pd_model}), is given by
\begin{equation}
S^{(c)}(\boldsymbol{x}')=\left\{ y\in\mathbb{R}^{3}:\overline{(u^{(c)},y)}\cap R(\boldsymbol{x}')\neq\emptyset\right\} ,\label{eq:shadow_region}
\end{equation}
where $\overline{(u^{(c)},y)}\triangleq\{\lambda y+(1-\lambda)u^{(c)}:\lambda\in[0,1]\}$
is the line segment joining the position $u^{(c)}$ of camera $c$
and $y$. Note that for an ellipsoidal region $R(\boldsymbol{x'})$,
the indicator function $1_{S^{(c)}(\boldsymbol{x}')}(\cdot)$ of its
shadow region can be computed in closed form (see Section \ref{subsec:Object-Representation,-Motion}).
\begin{comment}
ellipsoid, $R(\boldsymbol{x'})=\{y\in\mathbb{R}^{3}:(y-x^{(p)\prime})^{T}\mathrm{M}_{\boldsymbol{x'}}(y-x^{(p)\prime})\leq1\}$,
where $\mathrm{M}_{\boldsymbol{x'}}$ is a positive definite matrix
that depends on $\boldsymbol{x'}$. The indicator function $1_{S^{(c)}(\boldsymbol{x}')}(\cdot)$
of its shadow region can be computed in closed form \cite{schneider2002geometric},
see Section \ref{subsec:Object-Representation,-Motion}.
\end{comment}

\noindent 

To incorporate the effect of occlusions into the detection model,
the probability that $\boldsymbol{x}\in\boldsymbol{X}$ be detected
by camera $c$ should be close to zero when it is occluded from camera
$c$. This can be accomplished by extending the standard detection
probability so that: when $\boldsymbol{x}$ is in the LoS of camera
$c$, its detection probability is $P_{D}^{(c)}(\boldsymbol{x})$;
and when occluded by the other objects its detection probability scales
down to $\beta P_{D}^{(c)}(\boldsymbol{x})$, where $\beta$ is a
small positive number. More explicitly, 
\begin{multline}
P_{D}^{(c)}\!(\boldsymbol{x};\!\boldsymbol{X}\!\!-\negmedspace\{\boldsymbol{x}\})\negthinspace=\\
P_{D}^{(c)}(\boldsymbol{x})\Bigl(\mathcal{M}(\boldsymbol{x};\!\boldsymbol{X}\!\!-\negmedspace\{\boldsymbol{x}\})+\beta\bigl(1-\mathcal{M}(\boldsymbol{x};\!\boldsymbol{X}\!\!-\negmedspace\{\boldsymbol{x}\})\bigr)\Bigr),\label{eq:proposed_detection_model}
\end{multline}
where

\noindent \vspace*{-0.8cm}

\begin{equation}
\mathcal{M}(\boldsymbol{x};\!\boldsymbol{X}\!\!-\negmedspace\{\boldsymbol{x}\})=\prod_{\boldsymbol{x}'\boldsymbol{\in X}\!-\{\boldsymbol{x}\}\!}\left(1-1_{S^{(c)}(\boldsymbol{x}')}(\boldsymbol{x})\right)\label{eq:proposed_detection_model_2}
\end{equation}

Conditional on detection, $\boldsymbol{x}$ is observed at camera
$c$ as a bounding box $z^{(c)}\triangleq(z_{p}^{(c)},z_{e}^{(c)})$,
where $z_{p}^{(c)}$ is the center, and $z_{e}^{(c)}$ is the extent,
parameterized by the logarithms of the width (x-axis) and height (y-axis),
in image coordinates. The observed $z^{(c)}$ is a noisy version of
the box $\Phi^{(c)}(\boldsymbol{x})$ bounding the image of $R(\boldsymbol{x})$
in the camera's image plane, under the projection of the camera matrix
$\mathrm{P}_{3\times4}^{(c)}$. This matrix projects homogeneous points
in the world coordinate frame to homogeneous points in the image plane
of camera $c$, and can be obtained by standard calibration techniques
(see \cite{zhang2000flexible} for details). Note that for an ellipsoidal
region $R(\boldsymbol{x})$, the axis-aligned $\Phi^{(c)}(\boldsymbol{x})$
on the image plane can be computed analytically (see Section \ref{subsec:Object-Representation,-Motion}).
This observation process can be modeled by the likelihood
\begin{multline}
\!\!\!\!\!g^{(c)}(z^{(c)}|\boldsymbol{x})=\\
\mathcal{N}\!\left(\!z^{(c)};\Phi^{(c)}(\boldsymbol{x})+\left[\!\!\begin{array}{c}
0_{2\times1}\\
-\upsilon_{e}^{(c)}/2
\end{array}\!\!\right]\!,\textrm{diag\!}\left(\left[\!\!\begin{array}{c}
\upsilon_{p}^{(c)}\\
\upsilon_{e}^{(c)}
\end{array}\!\!\right]\right)\!\right),\label{likelihood_eq}
\end{multline}
where $\upsilon_{p}^{(c)}$ and $\upsilon_{e}^{(c)}$ are respectively
the vector of noise variances for the center and the extent (in logarithm)
of the box. This Gaussian model of the logarithms of the width and
height is equivalent to modeling the actual width and height as log-normals,
which ensures that they are non-negative. Note that these log-normals
have mean 1, and variances $e^{\upsilon_{e,1}^{(c)}}-1$ and $e^{\upsilon_{e,2}^{(c)}}-1$,
where $\upsilon_{e,1}^{(c)}$ and $\upsilon_{e,2}^{(c)}$ are the
two components of $\upsilon_{e}^{(c)}$. This means the observed width
and height are randomly scaled versions of their nominal values, with
an expected scaling factor of 1.

\subsection{Multi-view GLMB Filtering with Occlusions \label{subsec:Multi-view-GLMB-Update}}

This subsection presents a tractable GLMB approximation to the multi-view
Bayes filter to address occlusions. The proposed filter (with the
new detection model to account for occlusion) is referred to as Multi-View
GLMB with occlusion modeling (MV-GLMB-OC). 

Given the current GLMB filtering density \eqref{e:GLMB-(parameter set)},
the predicted density $\int\!\boldsymbol{f}_{+}\!\left(\boldsymbol{X}_{+}\!|\boldsymbol{X}\right)\boldsymbol{\pi}\left(\boldsymbol{X}\right)\!\delta\!\boldsymbol{X}$
in the Bayes recursion (\ref{e:MTBF}) is also a GLMB \cite{vo2013labeled},
which we denote by
\begin{equation}
\widehat{\boldsymbol{\pi}}_{+\!}\left(\boldsymbol{X}_{\!+}\right)=\Delta\left(\boldsymbol{X}_{+}\right)\sum_{I_{+},\xi}w_{+}^{\left(I_{+},\xi\right)}\delta_{I_{+}}[\mathcal{L}\left(\boldsymbol{X}_{\!+}\right)]\left[p_{+}^{(\xi)}\right]^{\boldsymbol{X}_{\!+}}\!\!,\label{deltaGLMB}
\end{equation}
where $I_{+}\in\mathcal{F}(\mathbb{L_{+}})$. Multiplying \eqref{deltaGLMB}
by the likelihood \eqref{e:MS_Lkhd_Abbrev_6-1} yields the next (unnormalized)
multi-object density
\begin{multline}
\boldsymbol{\pi}_{+}(\boldsymbol{X}_{\!+})\propto\Delta(\boldsymbol{X}_{+})\!\sum_{I_{+},\xi,\gamma_{+}}\!\!\delta_{\mathcal{L}(\gamma_{+})}\![\mathcal{L}\left(\boldsymbol{X}_{\!+}\right)]w_{+}^{\left(I_{+},\xi\right)}\\
\times\delta_{I_{+}}[\mathcal{L}\left(\boldsymbol{X}_{\!+}\right)]\!\left[p_{\boldsymbol{X}_{\!+}-\{\cdot\}}^{(\xi,\gamma_{+})}(\cdot)\right]^{\!\boldsymbol{X}_{\!+}}\!\!,\label{deltaGLMB-next_time}
\end{multline}
where 
\begin{equation}
p_{\boldsymbol{X}_{\!+}-\{\boldsymbol{x}_{+}\}}^{(\xi,\gamma_{+})}(\boldsymbol{x}_{+})=p_{+}^{(\xi)}\left(\boldsymbol{x}_{+}\right)\psi_{\boldsymbol{X}_{\!+}-\{\boldsymbol{x}_{+}\}}^{(\gamma_{+})}\!\left(\boldsymbol{x}_{+}\right).\label{deltaGLMB-1-1}
\end{equation}
As previously alluded to, the multi-object density \eqref{deltaGLMB-next_time}
is not a GLMB because $p_{\boldsymbol{X}_{\!+}-\{\boldsymbol{x}_{+}\}}^{(\xi,\gamma_{+})}$
depends on $\boldsymbol{X}_{\!+}\!-\!\{\boldsymbol{x}_{+}\}$. Nonetheless,
a good GLMB approximation of \eqref{deltaGLMB-next_time} can be obtained
by approximating $p_{\boldsymbol{X}_{\!+}-\{\boldsymbol{x}_{+}\}}^{(\xi,\gamma_{+})}$
with a density that is independent of $\boldsymbol{X}_{\!+}\!-\!\{\boldsymbol{x}_{+}\}$. 

Note that $\psi_{\boldsymbol{X}_{\!+}-\{\boldsymbol{x}_{+}\}}^{(\gamma_{+})}$
is the only factor of $p_{\boldsymbol{X}_{\!+}-\{\boldsymbol{x}_{+}\}}^{(\xi,\gamma_{+})}$,
which depends on $\boldsymbol{X}_{\!+}\!-\!\{\boldsymbol{x}_{+}\}$
(see \eqref{deltaGLMB-1-1}). Further inspection of \eqref{e:Psi}
and \eqref{e:MS_Lkhd_Abbrev_6-1} reveals that the detection probability
functions $P_{D,+}^{(c)}(\cdot;\boldsymbol{X}_{\!+}\!-\!\{\boldsymbol{x}_{+}\})$,
$c\!\!\in\!\!\{1{\textstyle :}C\}$ are the only constituent terms
that depend on $\boldsymbol{X}_{\!+}\!-\!\{\boldsymbol{x}_{+}\}$.
Moreover, it follows from \eqref{eq:proposed_detection_model} that
$P_{D,+}^{(c)}(\boldsymbol{x}_{+};\boldsymbol{X}_{\!+}\!-\!\{\boldsymbol{x}_{+}\})$
only takes on two values, depending on whether $\boldsymbol{x}_{+}$
falls in the shadow region of $\boldsymbol{X}_{\!+}\!-\!\{\boldsymbol{x}_{+}\}$
w.r.t. camera $c$. Assuming the positions of the elements of $\boldsymbol{X}_{\!+}\!-\!\{\boldsymbol{x}_{+}\}$
are concentrated around their predicted values according to the prediction
densities $p_{+}^{(\xi)}\left(\cdot,\ell\right),$ $\ell\in\mathcal{L}(\boldsymbol{X}_{\!+}\!-\!\{\boldsymbol{x}_{+}\})$,
we can approximate $P_{D,+}^{(c)}(\cdot;\boldsymbol{X}_{\!+}\!-\!\{\boldsymbol{x}_{+}\})$
by replacing the set $\boldsymbol{X}_{\!+}\!-\{\boldsymbol{x}_{+}\}$
with its predicted value. Noting that the term $\delta_{I_{+}}[\mathcal{L}\left(\boldsymbol{X}_{\!+}\right)]$
in \eqref{deltaGLMB-next_time} implies $\mathcal{L}(\boldsymbol{X}_{\!+})=I_{+}$,
the prediction of $\boldsymbol{X}_{\!+}\!-\!\{\boldsymbol{x}_{+}\}$
is 
\begin{equation}
\boldsymbol{X}_{\!+}^{(\xi,I_{+})}=\{(x_{+}^{(\xi,\ell)},\ell):\ell\in I_{+}-\mathcal{L}(\boldsymbol{x}_{+})\},\label{pred_big_X_for_pd}
\end{equation}
where $x_{+}^{(\xi,\ell)}$ denotes an estimate (e.g. mean, mode)
from the density $p_{+}^{(\xi)}\left(\cdot,\ell\right)$, which is
either the birth density $f_{B,+}(\cdot,\ell)$ if $\ell\in\mathbb{B}_{+}$
or $\int f_{S,+}(\cdot|x,\ell)p^{(\xi)}\left(x,\ell\right)dx$ if
$\ell\notin\mathbb{B}_{+}$ \cite{vo2013labeled}.

The above approximation translates to
\begin{equation}
p_{\boldsymbol{X}_{\!+}-\{\boldsymbol{x}_{+}\}}^{(\xi,\gamma_{+})}\approx p_{\boldsymbol{X}_{\!+}^{(\xi,I_{+})}}^{(\xi,\gamma_{+})},\label{big_X_approx}
\end{equation}
which is independent of $\boldsymbol{X}_{\!+}\!-\{\boldsymbol{x}_{+}\}$,
thereby turning \eqref{deltaGLMB-next_time} into a GLMB. Moreover,
the computation of this GLMB approximation to \eqref{deltaGLMB-next_time}
only differs from the MS-GLMB recursion \eqref{e:GLMB_operator} in
the detection probabilities 

\begin{equation}
P_{D,+}^{(\xi,I_{+}\!)}\!(\ell)\!\triangleq\!\left(\!P_{D,+}^{(1)}((\hat{x}_{+}\!,\!\ell);\!\boldsymbol{X}_{\!+}^{(\xi,I_{+}\!)}),\!...,\!P_{D,+}^{(C)}((\hat{x}_{+}\!,\!\ell);\!\boldsymbol{X}_{\!+}^{(\xi,I_{+}\!)})\!\right)\!,\label{eq:pd_for_section4}
\end{equation}
where $\ell=\mathcal{L}(\boldsymbol{x}_{+})$, and $\hat{x}_{+}$
denotes an estimate (e.g. mean, mode) from the density $p_{+}^{(\xi)}\left(\cdot,\ell\right)$.
Specifically, the GLMB approximation of the multi-object filtering
density can be propagated by the MS-GLMB recursion
\begin{equation}
\boldsymbol{\pi}_{+}\!=\!\Omega\!\left(\!\boldsymbol{\pi};\!\{P_{D,+}^{(\xi,I_{+})}(\ell):\!\ell\!\in I_{+},\left(\xi,I_{+}\right)\!\in\!\Xi\!\times\!\mathcal{F}(\mathbb{L_{+}})\}\!\right)\!.\label{deltaGLMB-approx_next_time}
\end{equation}
The integration of the proposed occlusion model (via the detection
probabilities) into the MS-GLMB filter is shown in Fig. \ref{MVGLMB_OC_proc_chain}.
The implementation of this so-called MV-GLMB-OC filter is discussed
in the next section. 

\section{Implementation \label{sec:Implementation}}

This section describe the implementation of the proposed filter for
ellipsoidal objects. Section \ref{subsec:Object-Representation,-Motion}
provides mathematical representations for the objects and the multi-object
model parameters. Propagation of the MV-GLMB-OC filtering density
is then described in Section \ref{subsec:Multi-view-GLMB-Filter_with_truncation}.%
\begin{comment}
Having described how GLMB filtering can be applied to resolve occlusions
in multi-view settings with the suitable detection model, this section
describes how the filter is implemented for 3D people tracking.
\end{comment}

\subsection{Object Representation and Model Parameters\label{subsec:Object-Representation,-Motion}}

Each object is represented by an axis-aligned ellipsoid. For an object
with labeled state $\boldsymbol{x}=(x,\ell)$, the position $x^{(p)}$
is the centroid, and the shape parameter $x^{(s)}$ is a vector containing
the logarithms of the half-lengths of the ellipsoid's principal axes.
Further, the time-evolution of the state vector $x$ is modeled by
a linear Gaussian transition density:
\begin{equation}
f_{S,+}(x_{+}|x,\ell)=\mathcal{N}\left(x_{+};\mathrm{F}x+\left[\!\!\begin{array}{c}
0_{6\times1}\\
-\upsilon^{(s)}/2
\end{array}\!\!\right],\mathrm{Q}\right),\label{eq:transition_density}
\end{equation}

\noindent where 

\noindent \vspace*{-0.7cm}

\begin{align}
\mathrm{F} = & \!\left[\begin{array}{c;{2pt/2pt}c} \mathrm{I}_{3}\otimes\left[\begin{array}{cc} 1 & T\\ 0 & 1 \end{array}\right] & 0_{6\times3}\\
\hdashline[2pt/2pt]
0_{3\times6} & \mathrm{I}_{3} \end{array}\right],\\
Q = & \!\left[\begin{array}{c;{2pt/2pt}c} \!\!\!\mathrm{diag}(\upsilon^{(p)})\!\otimes\!\left[\!\!\begin{array}{c} \frac{T^{2}}{2}\\ T \end{array}\!\!\right]\!\!\left[\!\!\begin{array}{cc} \frac{T^{2}}{2} & T\end{array}\!\!\right] & 0_{6\times3}\\
\hdashline[2pt/2pt]
0_{3\times6} & \!\mathrm{diag}(\upsilon^{(s)}) \end{array}\!\!\right],
\end{align}

\noindent $T$ is the sampling period, $\upsilon^{(p)}$ and $\upsilon^{(s)}$
are, respectively, 3D vectors of noise variances for the components
of the centroid and shape parameter (in logarithm) of the ellipsoid.
This transition density describes a nearly constant velocity model
for the centroid and a Gaussian random-walk for the shape parameter.
Gaussianity of the logarithms of the half-lengths is equivalent to
modeling the half-lengths as log-normals, which ensure that they are
non-negative. Note that these log-normals have mean 1, and variances
$e^{\upsilon_{i}^{(s)}}-1$, $i=1,2,3$, where $\upsilon_{i}^{(s)}$
is the $i\mathrm{^{th}}$ components of $\upsilon^{(s)}$. Hence,
the observed half-lengths are randomly scaled versions of their nominal
values, with an expected scaling factor of 1. 

\noindent 
\begin{figure}
\subfloat[]{\includegraphics[width=0.4\columnwidth]{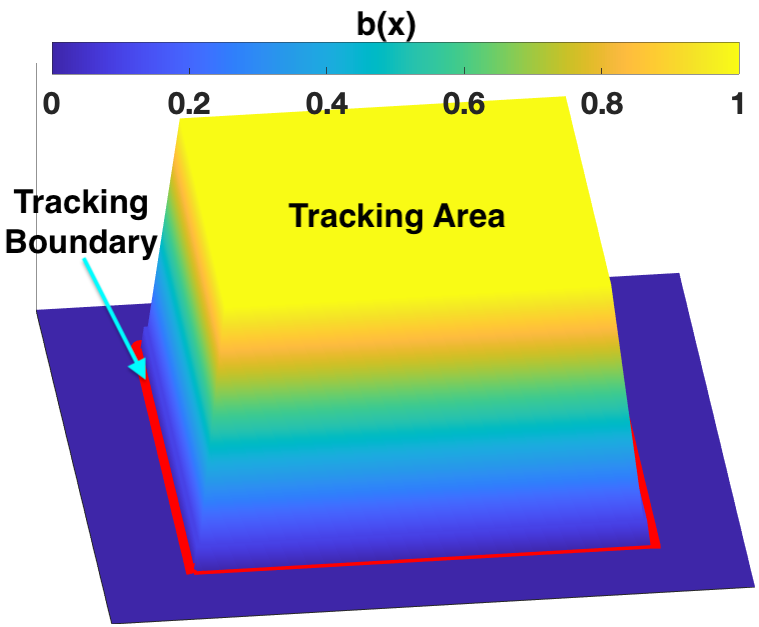}

}\subfloat[]{\includegraphics[width=0.6\columnwidth]{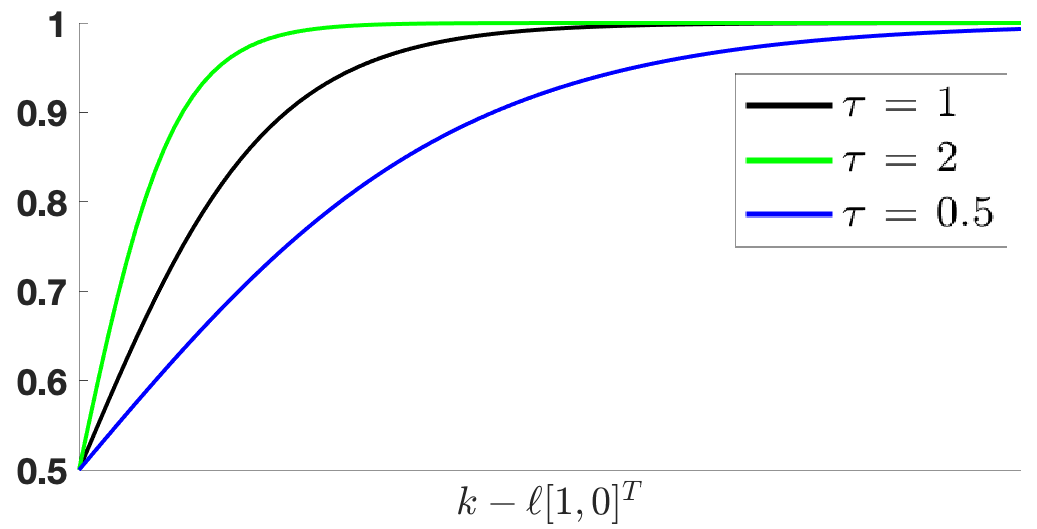}

}

\caption{Illustration of the survival probability model: (a) The scene mask
$b(x)$; (b) The control parameter $\tau$ of the sigmoid function. }
\label{scene_masks}
\end{figure}

\noindent \vspace*{-0.8cm}

\noindent 
\begin{figure*}[t]
\includegraphics[width=2\columnwidth]{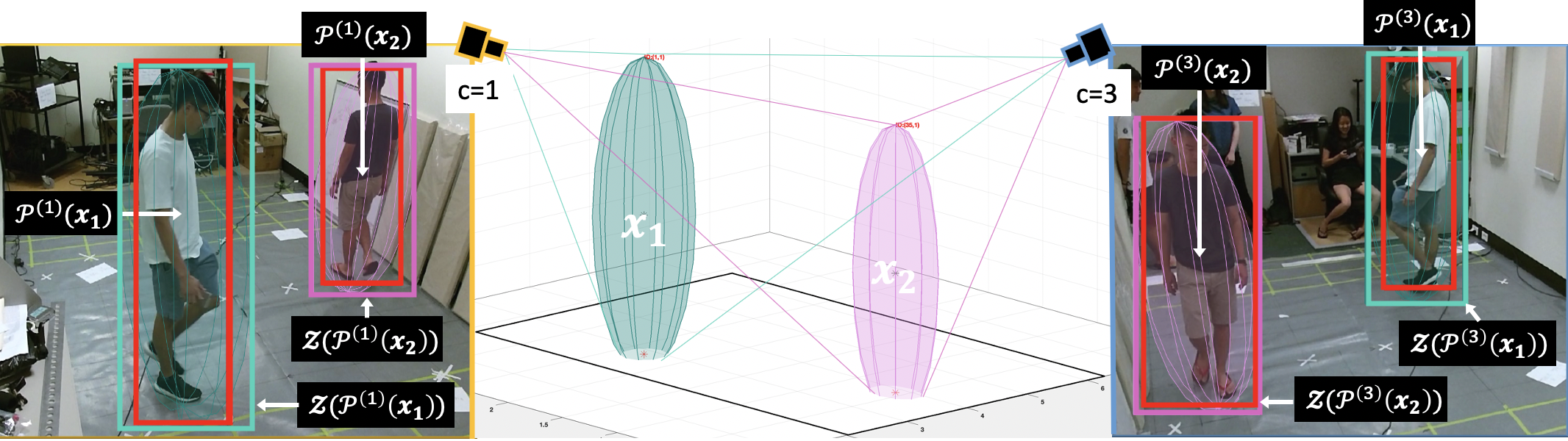}

\caption{The projections $\mathcal{P}^{(c)}$ of two quadrics (in cyan and
pink) onto two image views $(c=1,3)$ result in 2D conics. The transformation
$\mathcal{Z}$ yields the corresponding estimated bounding boxes (in
cyan and pink). The estimated bounding box and the measured bounding
box (in red) from monocular detector formulate the measurement likelihood
\eqref{likelihood_eq}.}
\label{meas_state_eq_illstra}
\end{figure*}

\noindent \vspace*{-0.8cm}

Empirically, objects that are in the scene for a long time, are more
likely to remain in the scene, unless they are close to the borders
(exit regions). This can be modeled via the following object survival
probability \cite{kim2019labeled}:

\begin{equation}
P_{S}(x,\ell)=\frac{b(x)}{1+\mathrm{exp}(-\tau(k-\ell[1,0]^{T}))},\label{death_model-1-1}
\end{equation}

\noindent where $b(x)$ is the the scene mask (chosen to be close
to one in the middle of the scene, and close to zero in the designated
exit regions and beyond) as depicted in Fig. \ref{scene_masks} (a),
and $\tau$ is the control parameter of the sigmoid function that
is dependent on the duration (age) of the track $k-\ell[1,0]^{T}$
as depicted in Fig. \ref{scene_masks} (b). 

The detection probability \eqref{eq:proposed_detection_model}-\eqref{eq:proposed_detection_model_2}
can be computed in closed form when the objects extents are ellipsoids.
As alluded to in Section \ref{subsec:Detection-Model}, the shadow
region indicator function $1_{S^{(c)}(\boldsymbol{y})}(\cdot)$ used
for checking whether an object is in the shadow region of the object
$\boldsymbol{y}$, can be determined analytically. Suppose that $R(\boldsymbol{y})$
in \eqref{eq:shadow_region} is a quadric, then it intersects the
line $\overline{(u^{(c)},x^{(p)})}$ (between $u^{(c)}$ and $x^{(p)}$)
if the roots of a certain quadratic equation are real \cite{schneider2002geometric}.
Consequently, for an axis-aligned ellipsoidal object representation,
the shadow region indicator function is given by 
\begin{equation}
1_{S^{(c)}(\boldsymbol{y})}(\boldsymbol{x})=\begin{cases}
1, & \text{ \ensuremath{\left(\mathcal{B}_{\boldsymbol{x},\boldsymbol{y}}^{(c)}\right){}^{2}-4\mathcal{A}_{\boldsymbol{x},\boldsymbol{y}}^{(c)}\mathcal{C}_{\boldsymbol{y}}^{(c)}\geq0} }\\
0, & \text{otherwise}
\end{cases},\label{eq:indicator_pd-1-1}
\end{equation}

\noindent where
\begin{align}
\mathcal{A}_{\boldsymbol{x},\boldsymbol{y}}^{(c)}= & (x^{(p)}-u^{(c)})^{T}\left(\mathrm{diag}(y^{(s)})\right)^{-2}(x^{(p)}-u^{(c)}),\\
\mathcal{B}_{\boldsymbol{x},\boldsymbol{y}}^{(c)}= & (x^{(p)}-u^{(c)})^{T}\left[2\left(\mathrm{diag}(y^{(s)})\right)^{-2}u^{(c)}+d_{\boldsymbol{y}}\right],\\
\mathcal{C}_{\boldsymbol{y}}^{(c)}= & (u^{(c)})^{T}\left[\left(\mathrm{diag}(y^{(s)})\right)^{-2}u^{(c)}+d_{\boldsymbol{y}}\right]+\mathcal{E}_{\boldsymbol{y}},\\
d_{\boldsymbol{y}}= & -2\frac{y^{(p)}}{(y^{(s)}\cdot y^{(s)})},\ \mathcal{E}_{\boldsymbol{y}}=\left\Vert y^{(p)}/y^{(s)}\right\Vert _{2}^{2}-1,
\end{align}
and $u^{(c)}$ is the position of camera $c$, with multiplication/division
of two vectors of the same dimension to be understood as point-wise
multiplication/division. %

In addition, using quadric projection \cite[pp. 201]{hartley2003multiple},
the relationship between the estimated bounding box $\Phi^{(c)}(\boldsymbol{x})$
and measured bounding box $z^{(c)}$ captured in the measurement likelihood
(\ref{likelihood_eq}), has the following closed form

\begin{equation}
\Phi^{(c)}(\boldsymbol{x})\triangleq\mathcal{Z}(\mathcal{P}^{(c)}(\boldsymbol{x})),\label{eq:meas_state_transformation}
\end{equation}

\noindent where

\noindent \begin{equation} 
\mathcal{P}^{(c)}(\boldsymbol{x})\!=\!\left(\!\mathrm{P}_{3\times4}^{(c)}\left[\!\begin{array}{c;{2pt/2pt}c} \!\!\!(\mathrm{diag}(x^{(s)}))^{-2}\!\! & d_{\boldsymbol{x}}/2\\
\hdashline[2pt/2pt]
d^{T}_{\boldsymbol{x}}/2 & \mathcal{E}_{\boldsymbol{x}} \end{array}\!\right]^{-1} \!\!\!\!\!\!\!(\mathrm{P}_{3\times4}^{(c)})^{T}\!\right)^{-1}\!\!\!\!\!\!\!\!,\!\!\!
\end{equation}

\noindent \vspace*{-0.4cm}
\begin{align}
\mathcal{Z}\left(\left[\begin{array}{c;{2pt/2pt}c} \mathrm{A} & r\\
\hdashline[2pt/2pt]
r^{T} & q \end{array}\right]\right) = & \left[\begin{array}{c} -\mathcal{Q}\mathrm{D}^{-1}\mathcal{Q}^{T}r\\ 2\nu\left\Vert [1,0]\mathcal{Q}\mathrm{D}^{-0.5}\right\Vert _{2}\\ 2\nu\left\Vert [0,1]\mathcal{Q}\mathrm{D}^{-0.5}\right\Vert _{2} \end{array}\right],\\
\nu = &(r^{T}\mathcal{Q}\mathrm{D}^{-1}\mathcal{Q}^{T}r-q)^{0.5}, 
\end{align}

\noindent $\mathcal{Q}$ is a matrix containing the eigenvectors of
A, and $\mathrm{D}$ is a diagonal matrix of the eigenvalues of A.
Given the camera matrices $\mathrm{P_{3\times4}^{(1)}},...,P_{3\times4}^{(C)}$,
$\mathcal{P}^{(c)}(\cdot)$ is a matrix-to-matrix projection that
transforms the quadric into a conic on each image of camera $c$ \cite[pp. 201]{hartley2003multiple}.
$\mathcal{Z}(\cdot)$ is a matrix-to-vector transformation that transforms
the conic into a 4D bounding box (in the same format as $z^{(c)}$).
The illustration of the overall transformation \eqref{eq:meas_state_transformation}
is depicted in Fig. \ref{meas_state_eq_illstra}.

The Poisson false alarms intensity for camera $c$ is $\kappa^{(c)}\triangleq\lambda_{c}\mathcal{U}(\cdot)$,
where $\lambda_{c}$ is the false-positive (clutter) rate, and $\mathcal{U}(\cdot)$
is a uniform distribution on the measurement space $\mathbb{Z}^{(c)}$.
In many visual tracking cases, this value can either be estimated
offline or manually tuned. The false alarm intensity can be estimated
by the Cardinalized Probability Hypothesis Density (CPHD) clutter
estimator \cite{mahler2011cphd}. In this work, we bootstrap the CPHD
clutter intensity estimator output to the tracker \cite{do2019multiple}.

\subsection{MV-GLMB-OC Filter Implementation \label{subsec:Multi-view-GLMB-Filter_with_truncation}}

The number of components of the GLMB filtering density grows super-exponentially
over time. To maintain tractability in GLMB filter implementations,
truncating insignificant components has been proven to minimize the
$L_{1}$ approximation error \cite{vo2019multi}. This truncation
strategy can be formulated as an NP-hard multi-dimensional assignment
problem \cite{vo2019multi}. Nonetheless, it can be solved by exploiting
certain structural properties, and suitable adaptation of 2D assignment
solutions such as Murty's or Auction \cite{vo2019multi}. 

The MV-GLMB-OC recursion described in Section \ref{subsec:Multi-view-GLMB-Update},
can be directly implemented with separate prediction and update, i.e.
by computing a truncated version of the prediction \eqref{deltaGLMB}
and the corresponding detection probabilities $\{P_{D,+}^{(\xi,I_{+})}(\ell):\!\ell\!\in I_{+},\left(\xi,I_{+}\right)\!\in\!\Xi\!\times\!\mathcal{F}(\mathbb{L_{+}})\}$,
then using this to compute a truncated version of the update \eqref{deltaGLMB-approx_next_time}.
This strategy requires keeping a significant portion of the predicted
components that would end up as updated components with negligible
weights, thereby wasting computations in solving a large number of
2D assignment problems. Thus, this approach is inefficient and becomes
infeasible for systems with many sensors \cite{vo2019multi}. 

In this work, we exploit an efficient GLMB truncation strategy that
has a linear complexity in the sum of the measurements across all
sensors \cite{vo2019multi}. This approach bypasses the prediction
truncation, and returns the significant components of the next GLMB
filtering density \eqref{deltaGLMB-approx_next_time} by sampling
from a discrete probability distribution proportional to the weights
of the components \cite{vo2019multi}. This means GLMB components
with higher weights are more likely to be selected than those with
lower weights. For the MV-GLMB-OC recursion, this discrete probability
distribution $s(\cdot;P_{D,+})$ of the GLMB components, is determined
by the detection probabilities $P_{D,+}\!\triangleq\!\{P_{D,+}^{(\xi,I_{+})}(\ell)\!:\!\ell\negmedspace\in\negmedspace I_{+},\left(\xi,I_{+}\right)\negmedspace\in\negmedspace\Xi\!\times\!\mathcal{F}(\mathbb{L_{+}})\}$
(and other multi-object system parameters, which are suppressed for
clarity) \cite{vo2019multi}. However, since truncation of the prediction
\eqref{deltaGLMB} has been bypassed, the predicted components $\{(\xi,I_{+})\negmedspace\in\negmedspace\Xi\!\times\!\mathcal{F}(\mathbb{L_{+}})\}$
and their corresponding detection probabilities are not available.
Nonetheless, importance sampling can be used to generate weighted
samples of $s(\cdot;P_{D,+})$ by sampling from $s(\cdot;\widehat{P}_{D,+})$,
where $\widehat{P}_{D,+}\negmedspace\triangleq\negmedspace\{P_{D,+}^{(\xi,I\uplus\mathbb{B}_{+})}(\ell)\!:\!\ell\negmedspace\in\negmedspace I\uplus\mathbb{B}_{+},\left(\xi,I\right)\negmedspace\in\negmedspace\Xi\!\times\!\mathcal{F}(\mathbb{L})\}$,
and then re-weight the resulting samples accordingly \cite{ristic2003beyond}.
Note that the detection probabilities $\widehat{P}_{D,+}$ can be
readily computed from the components of the (truncated) current GLMB
filtering density $\{(w^{\left(I,\xi\right)},p^{\left(\xi\right)})\!:\!(I,\xi)\negmedspace\in\negmedspace\mathcal{F}(\mathbb{L})\!\times\!\Xi\}$.
Moreover, $P_{D,+}^{(\xi,I\uplus\mathbb{B}_{+})}\!\preceq\!P_{D,+}^{(\xi,I_{+})}$,
for any $I_{+}\!\subseteq\!I\uplus\mathbb{B}_{+}$, it follows from
\cite{vo2017efficient} that $s(\cdot;\widehat{P}_{D,+})$ is more
diffused than $s(\cdot;P_{D,+})$, i.e. the support of $s(\cdot;\widehat{P}_{D,+})$
contains the support of $s(\cdot;P_{D,+})$. 

The MS-GLMB and MV-GLMB-OC recursions are presented in Algorithm \ref{High_level_MSGLMB_pseudocode}
and \ref{High_level_MV-GLMB-OC_pseudocode} respectively. Observe
that the main difference is the additional computation of the detection
probabilities prior to and re-weighting after the Gibbs sampling step
in the MV-GLMB-OC filter. %

In this work, the object's birth density $f_{B,+}(\cdot,\ell)$, single-object
transition \eqref{eq:transition_density} and likelihood \eqref{likelihood_eq}
are all Gaussians. Standard Kalman prediction and Unscented Kalman
update are used to evaluate the single-object filtering density $p_{+}^{(\xi_{+})}$,
which results in a Gaussian.%

\medskip{}

\noindent 
\begin{algorithm}[H]
\noindent {\small{}\rule[0.5ex]{1\columnwidth}{0.5pt}}{\small\par}

\textsf{\textbf{\footnotesize{}Global Input:}}\textsf{\footnotesize{}
}{\footnotesize{}$\left\{ \left(P_{B,+}(\ell),f_{B,+}(\cdot,\ell)\right)\right\} _{\ell\in\mathbb{B}_{+}},\boldsymbol{f}_{S,+}\left(\cdot|\cdot\right),P_{S}(\cdot)$}{\footnotesize\par}

\textsf{\textbf{\footnotesize{}Global Input:}}{\footnotesize{} $\kappa,P_{D},g$}{\footnotesize\par}

\textsf{\textbf{\footnotesize{}Input:}}\textsf{\footnotesize{} $\boldsymbol{\pi}\triangleq\left\{ \left(w^{\left(I,\xi\right)},p^{\left(\xi\right)}\right):\left(I,\xi\right)\in\mathcal{F}(\mathbb{L})\times\Xi\right\} $}{\footnotesize\par}

\textsf{\textbf{\footnotesize{}Output: $\boldsymbol{\pi}_{+}\triangleq\left\{ \left(w_{+}^{\left(I_{+},\xi_{+}\right)},p_{+}^{\left(\xi_{+}\right)}\right)\!:\left(I_{+},\xi_{+}\right)\in\mathcal{F}(\mathbb{L_{+}})\times\Xi_{+}\right\} $}}{\footnotesize\par}

{\footnotesize{}\smallskip{}
}{\footnotesize\par}

\noindent {\footnotesize{}\rule[0.5ex]{1\columnwidth}{0.5pt}}{\footnotesize\par}

{\footnotesize{}\smallskip{}
}{\footnotesize\par}

\noindent {\footnotesize{}}%
\begin{tabular*}{0.8\columnwidth}{@{\extracolsep{\fill}}l}
\textsf{\textbf{\footnotesize{}for}}\textsf{\footnotesize{} $\left(I,\xi\right)\in\mathcal{F}(\mathbb{L})\times\Xi$}\tabularnewline
\textsf{\footnotesize{}$\quad$Construct stationary distribution from
inputs}\tabularnewline
\textsf{\footnotesize{}$\quad$Run Gibbs sampler to obtain samples
$\gamma_{+}$ }{\footnotesize{}\cite[Algorithm 3]{vo2019multi}}\tabularnewline
\textsf{\footnotesize{}$\quad$Use samples $\gamma_{+}$ to compute
$\boldsymbol{\pi}_{+}$ }\tabularnewline
\textsf{\textbf{\footnotesize{}end for }}\tabularnewline
\textsf{\footnotesize{}Extract labeled state estimates}\tabularnewline
\end{tabular*}{\footnotesize\par}

{\small{}\caption{MS-GLMB Filter \cite{vo2019multi}}
\label{High_level_MSGLMB_pseudocode}}{\small\par}
\end{algorithm}

{\footnotesize{}\medskip{}
}{\footnotesize\par}

\noindent 
\begin{algorithm}[H]
\noindent {\small{}\rule[0.5ex]{1\columnwidth}{0.5pt}}{\small\par}

\textsf{\textbf{\footnotesize{}Global Input:}}\textsf{\footnotesize{}
}{\footnotesize{}$\left\{ \left(P_{B,+}(\ell),f_{B,+}(\cdot,\ell)\right)\right\} _{\ell\in\mathbb{B}_{+}},\boldsymbol{f}_{S,+}\left(\cdot|\cdot\right),P_{S}(\cdot)$}{\footnotesize\par}

\textsf{\textbf{\footnotesize{}Global Input:}}{\footnotesize{} $\kappa,P_{D},g$}{\footnotesize\par}

\textsf{\textbf{\footnotesize{}Input:}}\textsf{\footnotesize{} $\boldsymbol{\pi}\triangleq\left\{ \left(w^{\left(I,\xi\right)},p^{\left(\xi\right)}\right):\left(I,\xi\right)\in\mathcal{F}(\mathbb{L})\times\Xi\right\} $}{\footnotesize\par}

\textsf{\textbf{\footnotesize{}Output: $\boldsymbol{\pi}_{+}\triangleq\left\{ \left(w_{+}^{\left(I_{+},\xi_{+}\right)},p_{+}^{\left(\xi_{+}\right)}\right)\!:\left(I_{+},\xi_{+}\right)\in\mathcal{F}(\mathbb{L_{+}})\times\Xi_{+}\right\} $}}{\footnotesize\par}

{\footnotesize{}\smallskip{}
}{\footnotesize\par}

\noindent {\footnotesize{}\rule[0.5ex]{1\columnwidth}{0.5pt}}{\footnotesize\par}

{\footnotesize{}\smallskip{}
}{\footnotesize\par}

\noindent {\footnotesize{}}%
\begin{tabular*}{0.8\columnwidth}{@{\extracolsep{\fill}}l}
\textsf{\textbf{\footnotesize{}for}}\textsf{\footnotesize{} $\left(I,\xi\right)\in\mathcal{F}(\mathbb{L})\times\Xi$}\tabularnewline
\textsf{\footnotesize{}$\quad$Compute occlusion-based probability
of detection }\tabularnewline
\textsf{\footnotesize{}$\qquad$$\{P_{D,+}^{(\xi,I\uplus\mathbb{B}_{+})}(\ell):\ell\in I\uplus\mathbb{B}_{+}\}$
via \eqref{eq:pd_for_section4}}\tabularnewline
\textsf{\footnotesize{}$\quad$Construct stationary distribution from
inputs and }\tabularnewline
\textsf{\footnotesize{}$\qquad$$\{P_{D,+}^{(\xi,I\uplus\mathbb{B}_{+})}(\ell):\ell\in I\uplus\mathbb{B}_{+}\}$}\tabularnewline
\textsf{\footnotesize{}$\quad$Run Gibbs sampler to obtain samples
$\gamma_{+}$}{\footnotesize{} \cite[Algorithm 3]{vo2019multi}}\tabularnewline
\textsf{\footnotesize{}$\quad$Update occlusion-based probability
of detection }\tabularnewline
\textsf{\footnotesize{}$\qquad$$\{P_{D,+}^{(\xi,\mathcal{L}(\gamma_{+}))}(\ell):\ell\in\mathcal{L}(\gamma_{+})\}$,
via \eqref{eq:pd_for_section4}}\tabularnewline
\textsf{\footnotesize{}$\quad$Use samples $\gamma_{+}$, $\{P_{D,+}^{(\xi,\mathcal{L}(\gamma_{+}))}(\ell):\ell\in\mathcal{L}(\gamma_{+})\}$
to compute $\boldsymbol{\pi}_{+}$}\tabularnewline
\textsf{\textbf{\footnotesize{}end for}}\tabularnewline
\textsf{\footnotesize{}Extract labeled state estimates}\tabularnewline
\end{tabular*}{\footnotesize\par}

{\small{}\caption{MV-GLMB-OC Filter }
\label{High_level_MV-GLMB-OC_pseudocode}}{\small\par}
\end{algorithm}

\section{Experiments \label{sec:Experimental-Results}}

This section demonstrates the three main advantages of the proposed
MV-GLMB-OC approach. The first is the capability to produce 3D object
trajectories using independent monocular detections from multiple
views, where each object is represented as a 3D ellipsoid of unknown
location and extent (Section \ref{subsec:WILDTRACKS-Dataset}). The
second is the amenability for uninterrupted/seamless operation in
the event that cameras are added, removed or repositioned on the fly
(Section \ref{subsec:CMC-Dataset-and}). The third is the flexibility
of not confining objects to the ground plane, which is demonstrated
by tracking people jumping and falling (Section \ref{subsec:3D-Multi-Modal-Tracking}).
The effectiveness of the proposed occlusion model is also studied,
by comparing the tracking performance of the MV-GLMB-OC against that
of the standard MS-GLMB filter.

We first focus our demonstrations on the latest WILDTRACKS dataset\footnote{https://www.epfl.ch/labs/cvlab/data/data-wildtrack/},
which involves seven-cameras at 1920\texttimes 1080 resolution with
overlapping views. The WILDTRACKS dataset is also supplied with calibrated
intrinsic and extrinsic camera parameters, along with 3D ground plane
annotations although these are restricted to the ground plane. WILDTRACKS
was initially introduced to address various perceived shortcomings
in older multi-view datasets, the arguments for which were originally
presented in \cite{chavdarova2018wildtrack} and are summarized as
follows. The DukeMTMC dataset \cite{ristani2016performance} is essentially
non-overlapping in views and is now no longer available. The PETS
2009 S2.L1 dataset \cite{ferryman2009pets2009} has supposed inconsistencies
when projecting 3D points across the views. The EPFL, SALSA and Campus
datasets \cite{fleuret2008multicamera,alameda2015salsa,xu2016multi}
involve a relatively small number of people, and are relatively sparse
in terms of person density, but do not provide 3D annotations. In
addition, the EPFL-RLC dataset \cite{chavdarova2017deep} only provides
annotations for a small subset of the last 300 of 8000 frames. For
the same reasons that the authors of WILDTRACKS were motivated to
introduce their new dataset, the older multi-view datasets superseded
by WILDTRACKS are not suitable for evaluating the MV-GLMB-OC filter
in the 3D world frame.

In the context of demonstrating the MV-GLMB-OC approach however, the
WILDTRACKS dataset is not suitable for evaluating tracking performance
in full 3D, i.e. changes in all 3 x, y, z-coordinates. While WILDTRACKS
provides 3D annotations, these are restricted to the ground plane.
Moreover the annotations are for centroids only, and do not capture
the extent (in terms of length, width and height) of objects in the
world coordinates. In our performance comparisons, the outputs of
the proposed MV-GLMB-OC filter on WILDTRACKS are limited to the estimated
centroids projected onto the ground plane%
\begin{comment}
, and uses our best reconstruction of the true camera locations
\end{comment}
. To demonstrate the full capabilities of MV-GLMB-OC, it is critical
to have annotations of the 3D centroids and their 3D extent, along
with the ground truths for each of the camera locations. Consequently
we introduce a new Curtin Multi-Camera (CMC) dataset which meets these
requirements. 

\begin{figure}[H]
\centering%
\begin{comment}
\includegraphics[width=7cm,height=3.2cm]{For_README.png}
\end{comment}

\includegraphics[width=1\columnwidth]{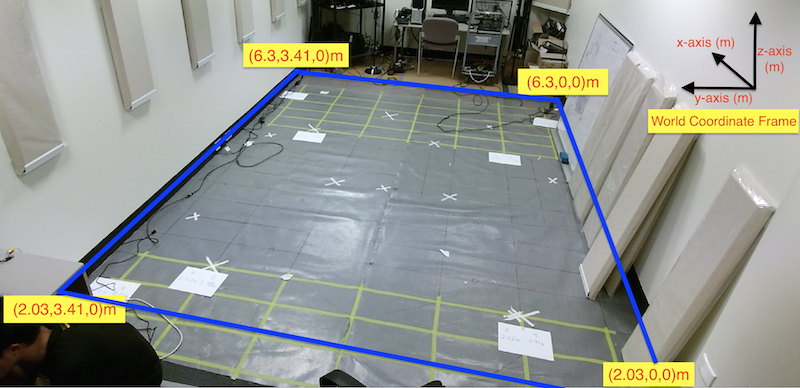}

\smallskip{}

\caption{Layout for CMC dataset: The blue line denotes the boundary of the
tracking area. The yellow boxes denote the coordinates of the boundary
in (x,y,z) axes. The 4 cameras are positioned (in sequence) at the
top 4 corners of the room. }
\label{Tracking area overview}

\medskip{}
\end{figure}

The new CMC dataset is a four-camera 1920x1024 resolution dataset
recorded at 4fps in a room with dimensions 7.67m x 3.41m x 2.7m. The
CMC dataset has 5 different sequences with varying levels of person
density and occlusion: CMC1 has a maximum of 3 people and virtually
no occlusion; CMC2 has a maximum of 10 people with some occlusion;
CMC3 has a maximum of 15 people with significant occlusion; while
CMC4 and CMC5 involve people jumping and falling with a maximum of
3 and 7 people respectively. CMC1 and CMC4 have low person density
and are intended for basic testing, while CMC2, CMC3 and CMC5 have
higher person density and significant visual occlusions across multiple
overlapping cameras, and are intended to highlight performance differences.
The convention for the world coordinate frame is illustrated in Fig.
\ref{Tracking area overview}. The origin is at the lower corner and
the ground plane corresponds to the x-y plane i.e.\textit{ $z$} =
$0$. In every sequence, each person enters the tracking area at $(2.03\mathrm{m},0.71\mathrm{m})$
with an average height of $1.7\mathrm{m}$. The dataset is also supplied
with camera locations and parameters, along with annotations for 3D
centroid and extent. The 2D monocular annotation for bounding boxes
is carried out with the MATLAB Image Labeler Tool, and the world coordinates
are obtained by averaging the homographic projection of the feet coordinates
from each view. The actual height and width of each person is used
for the annotation. 

\noindent 

A common setting for object survival and detection model parameters
is used in both evaluations on the WILDTRACKS and CMC datasets. Specifically:
the survival probability $P_{S}(\boldsymbol{x})$ given by (\ref{death_model-1-1}),
is parameterized by the control parameter $\tau$ = 0.5 and the scene
mask $b(\cdot)$ with a margin of 0.3m inside the border of the tracking
area; the detection probability, given in Section \ref{subsec:Detection-Model}
is parameterized by $P_{D}^{(c)}(\boldsymbol{x})=0.9$ and $\beta=0.1$.
For all cameras, the observed bounding box model is described in (\ref{likelihood_eq}),
with position noise parameterized by $\upsilon_{p}^{(c)}=[400,400]^{T}$,
and the extent noise parameterized by $\upsilon_{e}^{(c)}=[0.01,0.0025]{}^{T}$(on
the logarithms of the half-lengths of the principal axes). 

\noindent 
\begin{table*}
\captionsetup{justification=centering} 

\caption{\small{}WILDTRACKS Performance Benchmarks for 3D Position Estimates (restricted to the ground plane)}

\centering 

{\scriptsize{}}%
\begin{tabular}{|c|c|c|c|c|c|c|c|c|c|c|c|c||c|}
\hline 
{\scriptsize{}Detector and Tracker $\!\!\!\!$} & {\scriptsize{}IDF1 $\uparrow$} & {\scriptsize{}IDP $\uparrow$} & {\scriptsize{}IDR $\uparrow$} & {\scriptsize{}$\!\!\!$MT $\uparrow$$\!\!\!$} & {\scriptsize{}$\!\!\!$PT $\downarrow$$\!\!\!$} & {\scriptsize{}$\!\!\!$ML $\downarrow$$\!\!\!$} & {\scriptsize{}FP $\downarrow$} & {\scriptsize{}FN $\downarrow$} & {\scriptsize{}IDs $\downarrow$} & {\scriptsize{}FM $\downarrow$} & {\scriptsize{}$\!\!\!$MOTA $\uparrow$$\!\!\!$} & {\scriptsize{}$\!\!\!$MOTP $\uparrow$$\!\!\!$} & {\scriptsize{}$\!\!$$\mathrm{OSPA^{(2)}}$}\textcolor{black}{\scriptsize{}$\downarrow$}{\scriptsize{}$\!\!$}\tabularnewline
\hline 
\multirow{1}{*}{{\scriptsize{}YOLOv3+MV-GLMB-OC $\!\!\!\!$}} & {\scriptsize{}74.3\%} & \textbf{\scriptsize{}85.0\%} & \textbf{\scriptsize{}75.9\%} & {\scriptsize{}$\!\!\!$}\textbf{\textcolor{black}{\scriptsize{}136}}{\scriptsize{}$\!\!\!$} & {\scriptsize{}$\!\!\!$}\textcolor{black}{\scriptsize{}111}{\scriptsize{}$\!\!\!$} & {\scriptsize{}$\!\!\!$}\textcolor{black}{\scriptsize{}37}{\scriptsize{}$\!\!\!$} & \textbf{\textcolor{black}{\scriptsize{}424}} & \textcolor{black}{\scriptsize{}1333} & \textcolor{black}{\scriptsize{}104} & \textcolor{black}{\scriptsize{}86} & {\scriptsize{}$\!\!\!$69.7\%$\!\!\!$} & {\scriptsize{}$\!\!\!$}\textbf{\scriptsize{}73.2\%}{\scriptsize{}$\!\!\!$} & {\scriptsize{}$\!\!$}\textbf{\scriptsize{}0.69m}{\scriptsize{}$\!\!$}\tabularnewline
\hline 
\multirow{1}{*}{{\scriptsize{}YOLOv3+MS-GLMB $\!\!\!\!$}} & {\scriptsize{}74.2\%} & {\scriptsize{}79.0\%} & {\scriptsize{}69.9\%} & {\scriptsize{}$\!\!\!$}\textcolor{black}{\scriptsize{}116}{\scriptsize{}$\!\!\!$} & {\scriptsize{}$\!\!\!$}\textcolor{black}{\scriptsize{}85}{\scriptsize{}$\!\!\!$} & {\scriptsize{}$\!\!\!$}\textcolor{black}{\scriptsize{}83}{\scriptsize{}$\!\!\!$} & \textcolor{black}{\scriptsize{}841} & \textcolor{black}{\scriptsize{}1951} & \textcolor{black}{\scriptsize{}139} & \textcolor{black}{\scriptsize{}105} & {\scriptsize{}$\!\!\!$61.9\%$\!\!\!$} & {\scriptsize{}$\!\!\!$68.3\%$\!\!\!$} & {\scriptsize{}$\!\!$0.81m$\!\!$}\tabularnewline
\hline 
\multirow{1}{*}{{\scriptsize{}Faster-RCNN(VGG16)+MV-GLMB-OC $\!\!\!\!$}} & {\scriptsize{}76.5\%} & {\scriptsize{}84.5\%} & {\scriptsize{}70.0\%} & {\scriptsize{}$\!\!\!$}\textcolor{black}{\scriptsize{}119}{\scriptsize{}$\!\!\!$} & {\scriptsize{}$\!\!\!$}\textcolor{black}{\scriptsize{}118}{\scriptsize{}$\!\!\!$} & {\scriptsize{}$\!\!\!$}\textcolor{black}{\scriptsize{}47}{\scriptsize{}$\!\!\!$} & \textcolor{black}{\scriptsize{}545} & \textcolor{black}{\scriptsize{}1621} & \textcolor{black}{\scriptsize{}104} & \textcolor{black}{\scriptsize{}81} & {\scriptsize{}$\!\!\!$65.3\%$\!\!\!$} & {\scriptsize{}$\!\!\!$71.9\%$\!\!\!$} & {\scriptsize{}$\!\!$0.72m$\!\!$}\tabularnewline
\hline 
\multirow{1}{*}{{\scriptsize{}Faster-RCNN(VGG16)+MS-GLMB $\!\!\!\!$}} & {\scriptsize{}75.5\%} & {\scriptsize{}76.8\%} & {\scriptsize{}74.3\%} & {\scriptsize{}$\!\!\!$}\textcolor{black}{\scriptsize{}98}{\scriptsize{}$\!\!\!$} & {\scriptsize{}$\!\!\!$}\textcolor{black}{\scriptsize{}104}{\scriptsize{}$\!\!\!$} & {\scriptsize{}$\!\!\!$}\textcolor{black}{\scriptsize{}82}{\scriptsize{}$\!\!\!$} & \textcolor{black}{\scriptsize{}1114} & \textcolor{black}{\scriptsize{}1716} & \textcolor{black}{\scriptsize{}179} & \textcolor{black}{\scriptsize{}116} & {\scriptsize{}$\!\!\!$61.5\%$\!\!\!$} & {\scriptsize{}$\!\!\!$65.8\%$\!\!\!$} & {\scriptsize{}$\!\!$0.88m$\!\!$}\tabularnewline
\hline 
\multirow{1}{*}{{\scriptsize{}Deep-Occlusion+GLMB $\!\!\!\!$}} & {\scriptsize{}72.5\%} & {\scriptsize{}82.7\%} & {\scriptsize{}72.2\%} & {\scriptsize{}$\!\!\!$}\textcolor{black}{\scriptsize{}160}{\scriptsize{}$\!\!\!$} & {\scriptsize{}$\!\!\!$}\textcolor{black}{\scriptsize{}86}{\scriptsize{}$\!\!\!$} & {\scriptsize{}$\!\!\!$}\textcolor{black}{\scriptsize{}39}{\scriptsize{}$\!\!\!$} & \textcolor{black}{\scriptsize{}960} & \textbf{\textcolor{black}{\scriptsize{}990}} & \textcolor{black}{\scriptsize{}107} & \textbf{\textcolor{black}{\scriptsize{}64}} & {\scriptsize{}$\!\!\!$70.1\%$\!\!\!$} & {\scriptsize{}$\!\!\!$63.1\%$\!\!\!$} & {\scriptsize{}$\!\!$0.73m$\!\!$}\tabularnewline
\hline 
\multirow{1}{*}{{\scriptsize{}Deep-Occlusion+KSP+ptrack $\!\!\!\!$}} & \textbf{\scriptsize{}78.4\%} & {\scriptsize{}84.4\%} & {\scriptsize{}73.1\%} & {\scriptsize{}$\!\!\!$}\textcolor{black}{\scriptsize{}72}{\scriptsize{}$\!\!\!$} & {\scriptsize{}$\!\!\!$}\textbf{\textcolor{black}{\scriptsize{}74}}{\scriptsize{}$\!\!\!$} & {\scriptsize{}$\!\!\!$}\textbf{\textcolor{black}{\scriptsize{}25}}{\scriptsize{}$\!\!\!$} & \textcolor{black}{\scriptsize{}2007} & \textcolor{black}{\scriptsize{}5830} & \textbf{\textcolor{black}{\scriptsize{}103}} & \textcolor{black}{\scriptsize{}95} & {\scriptsize{}$\!\!\!$}\textbf{\scriptsize{}72.2\%}{\scriptsize{}$\!\!\!$} & {\scriptsize{}$\!\!\!$60.3\%$\!\!\!$} & {\scriptsize{}$\!\!$0.75m$\!\!$}\tabularnewline
\hline 
\end{tabular}{\scriptsize\par}

\smallskip{}

\raggedright

\emph{\footnotesize{}CLEAR MOT scores and OSPA}\textsuperscript{\emph{\footnotesize{}(2)}}\emph{\footnotesize{}
distance are calculated on standard position estimates ($\uparrow$
means higher is better while $\downarrow$ means lower is better).
Three different detectors are considered -Deep-Occlusion (multiocular),
Faster-RCNN}{\footnotesize{}(VGG16)}\emph{\footnotesize{} (monocular)
and YOLOv3 (monocular). Three types of trackers are considered -KSP+ptrack
or GLMB (single-sensor), MV-GLMB-OC (multi-view with occlusion model)
and MS-GLMB (multi-sensor without occlusion model).}{\footnotesize\par}

\noindent \vspace*{-0.6cm}

\label{WILDTRACKS_Tracking_Results}
\end{table*}

\noindent \vspace*{-.901cm}

\subsection{Performance Evaluation Criteria}

\subsubsection{Standard Evaluation on 3D Position Estimates}

The performance of various combinations of detectors and trackers
are evaluated using the CLEAR MOT devkit provided in \cite{leal2015motchallenge}.
For computing CLEAR MOT, we adhere to the convention of using the
Euclidean distance ($L_{2}$-norm) on the estimated 3D centroid with
a threshold of 1m. 

For MOT, the following performance indicators are reported: Multiple
Object Tracking Accuracy (MOTA) which penalizes normalized false negatives
(FNs), false positives (FPs) and identity switches (IDs) between consecutive
frames; Multiple Object Tracking Precision (MOTP) which accounts for
the overall dissimilarity between all true positives and the corresponding
ground truth objects \cite{kasturi2008framework}; Mostly Tracked
(MT), Partially Tracked (PT), Mostly Lost (MT) which indicate how
much of the trajectory is retained or lost by the tracker; Fragmentations
(FM) which account for interrupted tracks based on ground truth trajectories;
Identity Precision (IDP), Identity Recall (IDR) and $F_{1}$ score
(IDF1) which are analogous to the standard \emph{precision}, standard
\emph{recall} and $F_{1}$ score with identifications (tracks) \cite{ristani2016performance}.
For reference, we also provide performance indicators on the bounding
box detections, where we set the threshold at 0.5 and report: Multiple
Object Detection Accuracy (MODA) which accounts for misdetections
and false alarms; Multiple Object Detection Precision (MODP) which
accounts for the spatial overlap information between the bounding
boxes; \emph{precision} which is the measure of exactness; and \emph{recall}
which is the measure of quality. 

We note that CLEAR MOT is traditionally calculated over the entire
scenario window, and thus the tracking performance is reported after
the entire data stream has been processed. To evaluate the live or
online tracking performance over time, we employ the Optimal Sub-Pattern
Assignment (OSPA\textsuperscript{(2)}) distance between two sets
of tracks \cite{beard2020solution}. This distance is based on the
OSPA metric that captures both localization and cardinality errors
between two finite sets of a metric space with a suitable base-distance
between objects (e.g. the Euclidean distance) \cite{schuhmacher2008consistent}.
The OSPA\textsuperscript{(2)} metric is defined as the OSPA distance
between two sets of tracks over a time window. Details for OSPA and
OSPA\textsuperscript{(2)} metrics are given in Appendix \ref{subsec:OSPA2}.
By design, OSPA\textsuperscript{(2)} captures both localization and
cardinality errors between the set of true and estimated tracks, and
penalizes switched tracks or label changes \cite{beard2020solution}.
The resultant metric carries the interpretation of a time-averaged
per-track error. In our evaluation of the position estimate in real
world coordinates, we use a 3D Euclidean base-distance for OSPA\textsuperscript{(2)}
with order parameter $1$ and cutoff parameter 1m. Performance evaluation
for live or online tracking is given by plotting the error over a
sliding window of length $L_{w}=10$ frames, while overall performance
is captured in a single number by calculating the error over the entire
scenario window. 

\subsubsection{GIoU Based Evaluation on 3D Position with Extent}

As the proposed MV-GLMB-OC filter outputs 3D estimates of the object
centroid and extent, we extend the performance evaluations to capture
the joint error in the centroid and extent. This is achieved by employing
an alternative base-distance between two objects, in this case a 3D
generalized intersection over union (GIoU), which extends the commonly
used IoU to non-overlapping bounding boxes \cite{Rezatofighi_2018_CVPR}.
The details for the IoU and GIoU metrics are given in Appendix \ref{subsec:Intersection-over-Union-(IoU)-an}.
It is important to note that if there is no overlap between the ground
truth and estimated shape, the IoU distance is zero regardless of
their separation, whereas the GIoU distance captures the extent of
the error while retaining the metric property \cite{Rezatofighi_2018_CVPR}.
We present evaluations of the estimated centroid with extent for CLEAR
MOT (using a GIoU base-distance with a threshold of 0.5) and OSPA\textsuperscript{(2)}
metric with GIoU base-distance (and with unit order and cut-off parameters).
We refer the reader to \cite{Dat2020trustworthy} for the rationale
and discussions on the use of OSPA\textsuperscript{(2)}-GIoU for
performance evaluation. 

\subsection{WILDTRACKS Dataset \label{subsec:WILDTRACKS-Dataset}}

We test MV-GLMB-OC against the latest multi-camera detector (Deep-Occlusion)
\cite{baque2017deep} coupled with the \emph{k}-shortest-path (KSP)
algorithm \cite{berclaz2011multiple} and \emph{ptrack} as shown in
\cite{chavdarova2018wildtrack} (Deep-Occlusion+KSP+ptrack). KSP is
an optimization algorithm that finds the most likely sequence of ground
plane occupancies (trajectories) given by the multi-camera detector,
and \emph{ptrack} described in \cite{maksai2017non} improves and
smooths over tracks by learning motion patterns. As a baseline comparison,
we employ the Deep-Occlusion multi-camera detector combined with single-view
GLMB (Deep-Occlusion+GLMB). Since WILDTRACKS provides annotations
in real-world coordinates but restricted to the ground plane, tracking
is performed in real-world coordinates but also restricted to the
ground plane. To further explore the performance of MV-GLMB-OC, we
also run experiments using monocular detections from each of the cameras.
For the detectors, we use the monocular backbone of the Deep-Occlusion
detector i.e. VGG16-net trained using Faster-RCNN \cite{ren2015faster},
and separately with the newer YOLOv3 \cite{redmon2018yolov3}, to
produce separate monocular detections for input to MV-GLMB-OC. Since
WILDTRACKS does not supply the camera positions required for our proposed
occlusion model, we reconstruct the camera positions from the given
camera parameters. We note that KSP and/or ptrack is an offline or
batch method, while GLMB is online or recursive, and provides estimates
on the fly. 

\subsubsection{Model Parameters}

The birth density is adaptive/measurement-driven (see Section F in
\cite{reuter2014labeled}) with $P_{B,+}(\ell)=0.001$ and $f_{B,+}(x,\ell)=\mathcal{N}(x;\mu_{B,+}^{(\ell)},0.1^{2}\mathrm{I}_{9})$
where $\mu_{B,+}^{(\ell)}$ is obtained via clustering (e.g. \emph{k}-means).
The single-object transition is as described in (\ref{eq:transition_density})
with position noise and extent (in logarithm) noise parameterized
by:
\begin{align*}
\upsilon^{(p)}=[0.0016,0.0016,0.0016]{}^{T},\\
\upsilon^{(s)}=[0.0036,0.0036,0.0004]{}^{T}.
\end{align*}

\subsubsection{Discussion\emph{ \label{subsec:WildtrackS_Discussion}}}

Table \ref{WILDTRACKS_Tracking_Results} shows the CLEAR MOT and OSPA\textsuperscript{(2)}
benchmarks for MV-GLMB-OC (with occlusion modeling) and MS-GLMB (without
occlusion modeling) with two different detectors YOLOv3 and Faster-RCNN(VGG16).
Results for Deep-Occlusion+KSP+ptrack being the reference, are reproduced
directly from the original paper \cite{chavdarova2018wildtrack}.
The results indicate that the two trackers based on multi-camera detections,
i.e. Deep-Occlusion+KSP+ptrack and Deep-Occlusion+GLMB, have very
similar tracking performance in terms of MOTA/MOTP and OSPA\textsuperscript{(2)}.
Importantly, closer examination of the tracking results based on multiple
monocular detections indicates that performance is significantly improved
with the addition of the occlusion model. This can be seen from the
relative changes in the MOTA/MOTP and OSPA\textsuperscript{(2)}.
Several observations can also be drawn from comparing the multi-camera
detector with batch processing method (Deep-Occlusion+KSP+ptrack),
and the related monocular detector with online processing (Faster-RCNN(VGG16)+MV-GLMB-OC).
While the MOTP improves due to the use of multiple monocular detectors,
the MOTA degrades due to the use of an online method which is unable
to correct past estimates. This is corroborated by the overall OSPA\textsuperscript{(2)}
value which improves slightly from Deep-Occlusion+KSP+ptrack to Faster-RCNN(VGG16)+MV-GLMB-OC.
Surprisingly, the results based on YOLOv3 are better across the board
than that for Faster-RCNN(VGG16), even though YOLOv3 is more efficient
than Faster-RCNN(VGG16).  For reference, the CLEAR evaluations for
the detectors used in the experiment are presented in Appendix \ref{subsec:Monocular-Detector-Results},
from which it is noted that the monocular detections are generally
much poorer than the multi-camera detections due to severe occlusions. 

\noindent 
\begin{table*}
\captionsetup{justification=centering} 

\caption{\small{}CMC1,2,3 Performance Benchmarks for 3D Position Estimates }

\centering

\textcolor{black}{\scriptsize{}}%
\begin{tabular}{|c|c|c|c|c|c|c|c|c|c|c|c|c||c|}
\hline 
\multicolumn{14}{|c|}{\textcolor{black}{\scriptsize{}CMC1 (Maximum/Average 3 people)}}\tabularnewline
\hline 
\textcolor{black}{\scriptsize{}Detector and Tracker }{\scriptsize{}$\!\!\!\!$} & \textcolor{black}{\scriptsize{}IDF1 $\uparrow$} & \textcolor{black}{\scriptsize{}IDP $\uparrow$} & \textcolor{black}{\scriptsize{}IDR $\uparrow$} & {\scriptsize{}$\!\!\!$}\textcolor{black}{\scriptsize{}MT $\uparrow$}{\scriptsize{}$\!\!\!$} & {\scriptsize{}$\!\!\!$}\textcolor{black}{\scriptsize{}PT $\downarrow$}{\scriptsize{}$\!\!\!$} & {\scriptsize{}$\!\!\!$}\textcolor{black}{\scriptsize{}ML $\downarrow$}{\scriptsize{}$\!\!\!$} & \textcolor{black}{\scriptsize{}FP $\downarrow$} & \textcolor{black}{\scriptsize{}FN $\downarrow$} & \textcolor{black}{\scriptsize{}IDs $\downarrow$} & \textcolor{black}{\scriptsize{}FM $\downarrow$} & {\scriptsize{}$\!\!\!$}\textcolor{black}{\scriptsize{}MOTA $\uparrow$}{\scriptsize{}$\!\!\!$} & {\scriptsize{}$\!\!\!$}\textcolor{black}{\scriptsize{}MOTP $\uparrow$}{\scriptsize{}$\!\!\!$} & {\scriptsize{}$\!\!$$\mathrm{OSPA^{(2)}}$}\textcolor{black}{\scriptsize{}$\downarrow$}{\scriptsize{}$\!\!$}\tabularnewline
\hline 
\multirow{1}{*}{\textcolor{black}{\scriptsize{}YOLOv3+MV-GLMB-OC }{\scriptsize{}$\!\!\!\!$}} & \textbf{\textcolor{black}{\scriptsize{}99.7\%}} & \textbf{\textcolor{black}{\scriptsize{}99.4\%}} & \textbf{\textcolor{black}{\scriptsize{}100\%}} & \textbf{\scriptsize{}$\!\!\!$}\textbf{\textcolor{black}{\scriptsize{}3}}\textbf{\scriptsize{}$\!\!\!$} & \textbf{\scriptsize{}$\!\!\!$}\textbf{\textcolor{black}{\scriptsize{}0}}\textbf{\scriptsize{}$\!\!\!$} & \textbf{\scriptsize{}$\!\!\!$}\textbf{\textcolor{black}{\scriptsize{}0}}\textbf{\scriptsize{}$\!\!\!$} & \textbf{\textcolor{black}{\scriptsize{}4}} & \textbf{\textcolor{black}{\scriptsize{}0}} & \textbf{\textcolor{black}{\scriptsize{}0}} & \textbf{\textcolor{black}{\scriptsize{}0}} & {\scriptsize{}$\!\!\!$}\textbf{\textcolor{black}{\scriptsize{}99.4\%}}{\scriptsize{}$\!\!\!$} & {\scriptsize{}$\!\!\!$}\textbf{\textcolor{black}{\scriptsize{}91.8\%}}{\scriptsize{}$\!\!\!$} & {\scriptsize{}$\!\!$}\textbf{\scriptsize{}0.13m}{\scriptsize{}$\!\!$}\tabularnewline
\hline 
\multirow{1}{*}{\textcolor{black}{\scriptsize{}YOLOv3+MV-GLMB-OC{*} }{\scriptsize{}$\!\!\!\!$}} & \textcolor{black}{\scriptsize{}98.9\%} & \textcolor{black}{\scriptsize{}97.9\%} & \textcolor{black}{\scriptsize{}99.8\%} & \textbf{\scriptsize{}$\!\!\!$}\textbf{\textcolor{black}{\scriptsize{}3}}\textbf{\scriptsize{}$\!\!\!$} & \textbf{\scriptsize{}$\!\!\!$}\textbf{\textcolor{black}{\scriptsize{}0}}\textbf{\scriptsize{}$\!\!\!$} & \textbf{\scriptsize{}$\!\!\!$}\textbf{\textcolor{black}{\scriptsize{}0}}\textbf{\scriptsize{}$\!\!\!$} & \textcolor{black}{\scriptsize{}14} & \textcolor{black}{\scriptsize{}1} & \textbf{\textcolor{black}{\scriptsize{}0}} & \textbf{\textcolor{black}{\scriptsize{}0}} & {\scriptsize{}$\!\!\!$}\textcolor{black}{\scriptsize{}97.7\%}{\scriptsize{}$\!\!\!$} & {\scriptsize{}$\!\!\!$}\textcolor{black}{\scriptsize{}90.5\%}{\scriptsize{}$\!\!\!$} & {\scriptsize{}$\!\!$0.16m$\!\!$}\tabularnewline
\hline 
\multirow{1}{*}{\textcolor{black}{\scriptsize{}YOLOv3+MS-GLMB }{\scriptsize{}$\!\!\!\!$}} & \textcolor{black}{\scriptsize{}95.9\%} & \textcolor{black}{\scriptsize{}92.3\%} & \textcolor{black}{\scriptsize{}99.8\%} & \textbf{\scriptsize{}$\!\!\!$}\textbf{\textcolor{black}{\scriptsize{}3}}\textbf{\scriptsize{}$\!\!\!$} & \textbf{\scriptsize{}$\!\!\!$}\textbf{\textcolor{black}{\scriptsize{}0}}\textbf{\scriptsize{}$\!\!\!$} & \textbf{\scriptsize{}$\!\!\!$}\textbf{\textcolor{black}{\scriptsize{}0}}\textbf{\scriptsize{}$\!\!\!$} & \textcolor{black}{\scriptsize{}55} & \textcolor{black}{\scriptsize{}1} & \textcolor{black}{\scriptsize{}1} & \textbf{\textcolor{black}{\scriptsize{}0}} & {\scriptsize{}$\!\!\!$}\textcolor{black}{\scriptsize{}91.3\%}{\scriptsize{}$\!\!\!$} & {\scriptsize{}$\!\!\!$}\textcolor{black}{\scriptsize{}91.4\%}{\scriptsize{}$\!\!\!$} & {\scriptsize{}$\!\!$0.34m$\!\!$}\tabularnewline
\hline 
\multirow{1}{*}{\textcolor{black}{\scriptsize{}Faster-RCNN(VGG16)+MV-GLMB-OC }{\scriptsize{}$\!\!\!\!$}} & \textcolor{black}{\scriptsize{}99.5\%} & \textcolor{black}{\scriptsize{}99.1\%} & \textbf{\textcolor{black}{\scriptsize{}100\%}} & \textbf{\scriptsize{}$\!\!\!$}\textbf{\textcolor{black}{\scriptsize{}3}}\textbf{\scriptsize{}$\!\!\!$} & \textbf{\scriptsize{}$\!\!\!$}\textbf{\textcolor{black}{\scriptsize{}0}}\textbf{\scriptsize{}$\!\!\!$} & \textbf{\scriptsize{}$\!\!\!$}\textbf{\textcolor{black}{\scriptsize{}0}}\textbf{\scriptsize{}$\!\!\!$} & \textcolor{black}{\scriptsize{}6} & \textbf{\textcolor{black}{\scriptsize{}0}} & \textbf{\textcolor{black}{\scriptsize{}0}} & \textbf{\textcolor{black}{\scriptsize{}0}} & {\scriptsize{}$\!\!\!$}\textcolor{black}{\scriptsize{}99.1\%}{\scriptsize{}$\!\!\!$} & {\scriptsize{}$\!\!\!$}\textbf{\textcolor{black}{\scriptsize{}91.8\%}}{\scriptsize{}$\!\!\!$} & {\scriptsize{}$\!\!$}\textbf{\scriptsize{}0.13m}{\scriptsize{}$\!\!$}\tabularnewline
\hline 
\multirow{1}{*}{\textcolor{black}{\scriptsize{}Faster-RCNN(VGG16)+MV-GLMB-OC{*} }{\scriptsize{}$\!\!\!\!$}} & \textcolor{black}{\scriptsize{}95.5\%} & \textcolor{black}{\scriptsize{}91.4\%} & \textbf{\textcolor{black}{\scriptsize{}100\%}} & \textbf{\scriptsize{}$\!\!\!$}\textbf{\textcolor{black}{\scriptsize{}3}}\textbf{\scriptsize{}$\!\!\!$} & \textbf{\scriptsize{}$\!\!\!$}\textbf{\textcolor{black}{\scriptsize{}0}}\textbf{\scriptsize{}$\!\!\!$} & \textbf{\scriptsize{}$\!\!\!$}\textbf{\textcolor{black}{\scriptsize{}0}}\textbf{\scriptsize{}$\!\!\!$} & \textcolor{black}{\scriptsize{}62} & \textbf{\textcolor{black}{\scriptsize{}0}} & \textcolor{black}{\scriptsize{}1} & \textbf{\textcolor{black}{\scriptsize{}0}} & {\scriptsize{}$\!\!\!$}\textcolor{black}{\scriptsize{}90.4\%}{\scriptsize{}$\!\!\!$} & {\scriptsize{}$\!\!\!$}\textcolor{black}{\scriptsize{}90.5\%}{\scriptsize{}$\!\!\!$} & {\scriptsize{}$\!\!$0.14m$\!\!$}\tabularnewline
\hline 
\multirow{1}{*}{\textcolor{black}{\scriptsize{}Faster-RCNN(VGG16)+MS-GLMB}{\scriptsize{}
$\!\!\!\!$}} & \textcolor{black}{\scriptsize{}99.6\%} & \textcolor{black}{\scriptsize{}99.2\%} & \textbf{\textcolor{black}{\scriptsize{}100\%}} & \textbf{\scriptsize{}$\!\!\!$}\textbf{\textcolor{black}{\scriptsize{}3}}\textbf{\scriptsize{}$\!\!\!$} & \textbf{\scriptsize{}$\!\!\!$}\textbf{\textcolor{black}{\scriptsize{}0}}\textbf{\scriptsize{}$\!\!\!$} & \textbf{\scriptsize{}$\!\!\!$}\textbf{\textcolor{black}{\scriptsize{}0}}\textbf{\scriptsize{}$\!\!\!$} & \textcolor{black}{\scriptsize{}5} & \textbf{\textcolor{black}{\scriptsize{}0}} & \textbf{\textcolor{black}{\scriptsize{}0}} & \textbf{\textcolor{black}{\scriptsize{}0}} & {\scriptsize{}$\!\!\!$}\textcolor{black}{\scriptsize{}99.2\%}{\scriptsize{}$\!\!\!$} & {\scriptsize{}$\!\!\!$}\textcolor{black}{\scriptsize{}91.4\%}{\scriptsize{}$\!\!\!$} & {\scriptsize{}$\!\!$0.36m$\!\!$}\tabularnewline
\hline 
\end{tabular}{\scriptsize\par}

\textcolor{black}{\scriptsize{}}%
\begin{tabular}{|c|c|c|c|c|c|c|c|c|c|c|c|c||c|}
\hline 
\multicolumn{14}{|c|}{\textcolor{black}{\scriptsize{}CMC2 (Maximum/Average 10 people)}}\tabularnewline
\hline 
\textcolor{black}{\scriptsize{}Detector and Tracker}{\scriptsize{}$\!\!\!\!$} & \textcolor{black}{\scriptsize{}IDF1 $\uparrow$} & \textcolor{black}{\scriptsize{}IDP $\uparrow$} & \textcolor{black}{\scriptsize{}IDR $\uparrow$} & {\scriptsize{}$\!\!\!$}\textcolor{black}{\scriptsize{}MT $\uparrow$}{\scriptsize{}$\!\!\!$} & {\scriptsize{}$\!\!\!$}\textcolor{black}{\scriptsize{}PT $\downarrow$}{\scriptsize{}$\!\!\!$} & {\scriptsize{}$\!\!\!$}\textcolor{black}{\scriptsize{}ML $\downarrow$}{\scriptsize{}$\!\!\!$} & \textcolor{black}{\scriptsize{}FP $\downarrow$} & \textcolor{black}{\scriptsize{}FN $\downarrow$} & \textcolor{black}{\scriptsize{}IDs $\downarrow$} & \textcolor{black}{\scriptsize{}FM $\downarrow$} & {\scriptsize{}$\!\!\!$}\textcolor{black}{\scriptsize{}MOTA $\uparrow$}{\scriptsize{}$\!\!\!$} & {\scriptsize{}$\!\!\!$}\textcolor{black}{\scriptsize{}MOTP $\uparrow$}{\scriptsize{}$\!\!\!$} & {\scriptsize{}$\!\!$$\mathrm{OSPA^{(2)}}$}\textcolor{black}{\scriptsize{}$\downarrow$}{\scriptsize{}$\!\!$}\tabularnewline
\hline 
\multirow{1}{*}{\textcolor{black}{\scriptsize{}YOLOv3+MV-GLMB-OC }{\scriptsize{}$\!\!\!\!$}} & \textbf{\textcolor{black}{\scriptsize{}91.0\%}} & \textbf{\textcolor{black}{\scriptsize{}91.1\%}} & \textbf{\textcolor{black}{\scriptsize{}91.3\%}} & {\scriptsize{}$\!\!\!$}\textbf{\textcolor{black}{\scriptsize{}10}}{\scriptsize{}$\!\!\!$} & {\scriptsize{}$\!\!\!$}\textbf{\textcolor{black}{\scriptsize{}0}}{\scriptsize{}$\!\!\!$} & \textbf{\scriptsize{}$\!\!\!$}\textbf{\textcolor{black}{\scriptsize{}0}}\textbf{\scriptsize{}$\!\!\!$} & \textcolor{black}{\scriptsize{}16} & \textbf{\textcolor{black}{\scriptsize{}11}} & \textbf{\textcolor{black}{\scriptsize{}9}} & \textbf{\textcolor{black}{\scriptsize{}2}} & \textbf{\textcolor{black}{\scriptsize{}98.3\%}} & \textcolor{black}{\scriptsize{}81.7\%} & \textbf{\scriptsize{}0.30m}\tabularnewline
\hline 
\multirow{1}{*}{\textcolor{black}{\scriptsize{}YOLOv3+MV-GLMB-OC{*}}{\scriptsize{}$\!\!\!\!$}} & \textcolor{black}{\scriptsize{}90.1\%} & \textcolor{black}{\scriptsize{}90.2\%} & \textcolor{black}{\scriptsize{}90.0\%} & {\scriptsize{}$\!\!\!$}\textbf{\textcolor{black}{\scriptsize{}10}}{\scriptsize{}$\!\!\!$} & {\scriptsize{}$\!\!\!$}\textbf{\textcolor{black}{\scriptsize{}0}}{\scriptsize{}$\!\!\!$} & \textbf{\scriptsize{}$\!\!\!$}\textbf{\textcolor{black}{\scriptsize{}0}}\textbf{\scriptsize{}$\!\!\!$} & \textcolor{black}{\scriptsize{}38} & \textcolor{black}{\scriptsize{}29} & \textcolor{black}{\scriptsize{}11} & \textcolor{black}{\scriptsize{}7} & \textcolor{black}{\scriptsize{}96.2\%} & \textcolor{black}{\scriptsize{}78.9\%} & {\scriptsize{}0.34m}\tabularnewline
\hline 
\multirow{1}{*}{\textcolor{black}{\scriptsize{}YOLOv3+MS-GLMB}{\scriptsize{}$\!\!\!\!$}} & \textcolor{black}{\scriptsize{}67.7\%} & \textcolor{black}{\scriptsize{}79.9\%} & \textcolor{black}{\scriptsize{}58.9\%} & {\scriptsize{}$\!\!\!$}\textcolor{black}{\scriptsize{}4}{\scriptsize{}$\!\!\!$} & {\scriptsize{}$\!\!\!$}\textcolor{black}{\scriptsize{}6}{\scriptsize{}$\!\!\!$} & \textbf{\scriptsize{}$\!\!\!$}\textbf{\textcolor{black}{\scriptsize{}0}}\textbf{\scriptsize{}$\!\!\!$} & \textcolor{black}{\scriptsize{}8} & \textcolor{black}{\scriptsize{}550} & \textcolor{black}{\scriptsize{}34} & \textcolor{black}{\scriptsize{}30} & \textcolor{black}{\scriptsize{}71.5\%} & \textcolor{black}{\scriptsize{}74.4\%} & {\scriptsize{}0.70m}\tabularnewline
\hline 
\multirow{1}{*}{\textcolor{black}{\scriptsize{}Faster-RCNN(VGG16)+MV-GLMB-OC}{\scriptsize{}$\!\!\!\!$}} & \textcolor{black}{\scriptsize{}90.6\%} & \textcolor{black}{\scriptsize{}90.5\%} & \textcolor{black}{\scriptsize{}90.9\%} & {\scriptsize{}$\!\!\!$}\textbf{\textcolor{black}{\scriptsize{}10}}{\scriptsize{}$\!\!\!$} & \textbf{\scriptsize{}$\!\!\!$}\textbf{\textcolor{black}{\scriptsize{}0}}\textbf{\scriptsize{}$\!\!\!$} & \textbf{\scriptsize{}$\!\!\!$}\textbf{\textcolor{black}{\scriptsize{}0}}\textbf{\scriptsize{}$\!\!\!$} & \textcolor{black}{\scriptsize{}50} & \textcolor{black}{\scriptsize{}37} & \textbf{\textcolor{black}{\scriptsize{}9}} & \textcolor{black}{\scriptsize{}5} & \textcolor{black}{\scriptsize{}95.4\%} & \textbf{\textcolor{black}{\scriptsize{}83.7\%}} & {\scriptsize{}0.35m}\tabularnewline
\hline 
\multirow{1}{*}{\textcolor{black}{\scriptsize{}Faster-RCNN(VGG16)+MV-GLMB-OC{*}}{\scriptsize{}$\!\!\!\!$}} & \textcolor{black}{\scriptsize{}86.2\%} & \textcolor{black}{\scriptsize{}85.5\%} & \textcolor{black}{\scriptsize{}87.5\%} & {\scriptsize{}$\!\!\!$}\textbf{\textcolor{black}{\scriptsize{}10}}{\scriptsize{}$\!\!\!$} & {\scriptsize{}$\!\!\!$}\textbf{\textcolor{black}{\scriptsize{}0}}{\scriptsize{}$\!\!\!$} & \textbf{\scriptsize{}$\!\!\!$}\textbf{\textcolor{black}{\scriptsize{}0}}\textbf{\scriptsize{}$\!\!\!$} & \textcolor{black}{\scriptsize{}120} & \textcolor{black}{\scriptsize{}60} & \textcolor{black}{\scriptsize{}25} & \textcolor{black}{\scriptsize{}13} & \textcolor{black}{\scriptsize{}90.1\%} & \textcolor{black}{\scriptsize{}79.8\%} & {\scriptsize{}0.48m}\tabularnewline
\hline 
\multirow{1}{*}{\textcolor{black}{\scriptsize{}Faster-RCNN(VGG16)+MS-GLMB}{\scriptsize{}$\!\!\!\!$}} & \textcolor{black}{\scriptsize{}75.3\%} & \textcolor{black}{\scriptsize{}81.9\%} & \textcolor{black}{\scriptsize{}69.7\%} & {\scriptsize{}$\!\!\!$}\textcolor{black}{\scriptsize{}7}{\scriptsize{}$\!\!\!$} & {\scriptsize{}$\!\!\!$}\textcolor{black}{\scriptsize{}3}{\scriptsize{}$\!\!\!$} & \textbf{\scriptsize{}$\!\!\!$}\textbf{\textcolor{black}{\scriptsize{}0}}\textbf{\scriptsize{}$\!\!\!$} & \textbf{\textcolor{black}{\scriptsize{}7}} & \textcolor{black}{\scriptsize{}316} & \textcolor{black}{\scriptsize{}23} & \textcolor{black}{\scriptsize{}19} & \textcolor{black}{\scriptsize{}83.3\%} & \textcolor{black}{\scriptsize{}80.4\%} & {\scriptsize{}0.58m}\tabularnewline
\hline 
\end{tabular}{\scriptsize\par}

\textcolor{black}{\scriptsize{}}%
\begin{tabular}{|c|c|c|c|c|c|c|c|c|c|c|c|c||c|}
\hline 
\multicolumn{14}{|c|}{\textcolor{black}{\scriptsize{}CMC3 (Maximum/Average 15 people)}}\tabularnewline
\hline 
\textcolor{black}{\scriptsize{}Detector and Tracker}{\scriptsize{}$\!\!\!\!$} & \textcolor{black}{\scriptsize{}IDF1 $\uparrow$} & \textcolor{black}{\scriptsize{}IDP $\uparrow$} & \textcolor{black}{\scriptsize{}IDR $\uparrow$} & {\scriptsize{}$\!\!\!$}\textcolor{black}{\scriptsize{}MT $\uparrow$}{\scriptsize{}$\!\!\!$} & {\scriptsize{}$\!\!\!$}\textcolor{black}{\scriptsize{}PT $\downarrow$}{\scriptsize{}$\!\!\!$} & {\scriptsize{}$\!\!\!$}\textcolor{black}{\scriptsize{}ML $\downarrow$}{\scriptsize{}$\!\!\!$} & \textcolor{black}{\scriptsize{}FP $\downarrow$} & \textcolor{black}{\scriptsize{}FN $\downarrow$} & \textcolor{black}{\scriptsize{}IDs $\downarrow$} & \textcolor{black}{\scriptsize{}FM $\downarrow$} & {\scriptsize{}$\!\!\!$}\textcolor{black}{\scriptsize{}MOTA $\uparrow$}{\scriptsize{}$\!\!\!$} & {\scriptsize{}$\!\!\!$}\textcolor{black}{\scriptsize{}MOTP $\uparrow$}{\scriptsize{}$\!\!\!$} & {\scriptsize{}$\!\!$$\mathrm{OSPA^{(2)}}$}\textcolor{black}{\scriptsize{}$\downarrow$}{\scriptsize{}$\!\!$}\tabularnewline
\hline 
\multirow{1}{*}{\textcolor{black}{\scriptsize{}YOLOv3+MV-GLMB-OC}{\scriptsize{}$\!\!\!\!$}} & \textbf{\textcolor{black}{\scriptsize{}77.9\%}} & \textbf{\textcolor{black}{\scriptsize{}79.7\%}} & \textbf{\textcolor{black}{\scriptsize{}76.1\%}} & {\scriptsize{}$\!\!\!$}\textbf{\textcolor{black}{\scriptsize{}13}}{\scriptsize{}$\!\!\!$} & {\scriptsize{}$\!\!\!$}\textbf{\textcolor{black}{\scriptsize{}2}}{\scriptsize{}$\!\!\!$} & \textbf{\scriptsize{}$\!\!\!$}\textbf{\textcolor{black}{\scriptsize{}0}}\textbf{\scriptsize{}$\!\!\!$} & \textcolor{black}{\scriptsize{}63} & \textbf{\textcolor{black}{\scriptsize{}191}} & \textbf{\textcolor{black}{\scriptsize{}44}} & \textcolor{black}{\scriptsize{}33} & \textbf{\textcolor{black}{\scriptsize{}89.5\%}} & \textbf{\textcolor{black}{\scriptsize{}76.4\%}} & \textbf{\scriptsize{}0.51m}\tabularnewline
\hline 
\multirow{1}{*}{\textcolor{black}{\scriptsize{}YOLOv3+MV-GLMB-OC{*}}{\scriptsize{}$\!\!\!\!$}} & \textcolor{black}{\scriptsize{}72.1\%} & \textcolor{black}{\scriptsize{}77.9\%} & \textcolor{black}{\scriptsize{}67.2\%} & {\scriptsize{}$\!\!\!$}\textcolor{black}{\scriptsize{}11}{\scriptsize{}$\!\!\!$} & {\scriptsize{}$\!\!\!$}\textcolor{black}{\scriptsize{}4}{\scriptsize{}$\!\!\!$} & \textbf{\scriptsize{}$\!\!\!$}\textbf{\textcolor{black}{\scriptsize{}0}}\textbf{\scriptsize{}$\!\!\!$} & \textcolor{black}{\scriptsize{}47} & \textcolor{black}{\scriptsize{}437} & \textcolor{black}{\scriptsize{}51} & \textcolor{black}{\scriptsize{}37} & \textcolor{black}{\scriptsize{}81.1\%} & \textcolor{black}{\scriptsize{}72.3\%} & {\scriptsize{}0.61m}\tabularnewline
\hline 
\multirow{1}{*}{\textcolor{black}{\scriptsize{}YOLOv3+MS-GLMB}{\scriptsize{}$\!\!\!\!$}} & \textcolor{black}{\scriptsize{}50.5\%} & \textcolor{black}{\scriptsize{}69.9\%} & \textcolor{black}{\scriptsize{}39.5\%} & {\scriptsize{}$\!\!\!$}\textcolor{black}{\scriptsize{}0}{\scriptsize{}$\!\!\!$} & {\scriptsize{}$\!\!\!$}\textcolor{black}{\scriptsize{}15}{\scriptsize{}$\!\!\!$} & \textbf{\scriptsize{}$\!\!\!$}\textbf{\textcolor{black}{\scriptsize{}0}}\textbf{\scriptsize{}$\!\!\!$} & \textcolor{black}{\scriptsize{}5} & \textcolor{black}{\scriptsize{}1234} & \textcolor{black}{\scriptsize{}54} & \textcolor{black}{\scriptsize{}51} & \textcolor{black}{\scriptsize{}54.2\%} & \textcolor{black}{\scriptsize{}67.8\%} & {\scriptsize{}0.83m}\tabularnewline
\hline 
\multirow{1}{*}{\textcolor{black}{\scriptsize{}Faster-RCNN(VGG16)+MV-GLMB-OC}{\scriptsize{}$\!\!\!\!$}} & \textcolor{black}{\scriptsize{}71.7\%} & \textcolor{black}{\scriptsize{}74.9\%} & \textcolor{black}{\scriptsize{}68.8\%} & {\scriptsize{}$\!\!\!$}\textcolor{black}{\scriptsize{}12}{\scriptsize{}$\!\!\!$} & {\scriptsize{}$\!\!\!$}\textcolor{black}{\scriptsize{}3}{\scriptsize{}$\!\!\!$} & \textbf{\scriptsize{}$\!\!\!$}\textbf{\textcolor{black}{\scriptsize{}0}}\textbf{\scriptsize{}$\!\!\!$} & \textcolor{black}{\scriptsize{}71} & \textcolor{black}{\scriptsize{}303} & \textbf{\textcolor{black}{\scriptsize{}44}} & \textbf{\textcolor{black}{\scriptsize{}32}} & \textcolor{black}{\scriptsize{}85.2\%} & \textcolor{black}{\scriptsize{}73.5\%} & {\scriptsize{}0.61m}\tabularnewline
\hline 
\multirow{1}{*}{\textcolor{black}{\scriptsize{}Faster-RCNN(VGG16)+MV-GLMB-OC{*}}{\scriptsize{}$\!\!\!\!$}} & \textcolor{black}{\scriptsize{}67.7\%} & \textcolor{black}{\scriptsize{}72.1\%} & \textcolor{black}{\scriptsize{}63.8\%} & {\scriptsize{}$\!\!\!$}\textcolor{black}{\scriptsize{}10}{\scriptsize{}$\!\!\!$} & {\scriptsize{}$\!\!\!$}\textcolor{black}{\scriptsize{}5}{\scriptsize{}$\!\!\!$} & \textbf{\scriptsize{}$\!\!\!$}\textbf{\textcolor{black}{\scriptsize{}0}}\textbf{\scriptsize{}$\!\!\!$} & \textcolor{black}{\scriptsize{}92} & \textcolor{black}{\scriptsize{}419} & \textcolor{black}{\scriptsize{}59} & \textcolor{black}{\scriptsize{}44} & \textcolor{black}{\scriptsize{}79.8\%} & \textcolor{black}{\scriptsize{}68.0\%} & {\scriptsize{}0.70m}\tabularnewline
\hline 
\multirow{1}{*}{\textcolor{black}{\scriptsize{}Faster-RCNN(VGG16)+MS-GLMB}{\scriptsize{}$\!\!\!\!$}} & \textcolor{black}{\scriptsize{}54.3\%} & \textcolor{black}{\scriptsize{}73.2\%} & \textcolor{black}{\scriptsize{}43.1\%} & {\scriptsize{}$\!\!\!$}\textcolor{black}{\scriptsize{}0}{\scriptsize{}$\!\!\!$} & {\scriptsize{}$\!\!\!$}\textcolor{black}{\scriptsize{}15}{\scriptsize{}$\!\!\!$} & \textbf{\scriptsize{}$\!\!\!$}\textbf{\textcolor{black}{\scriptsize{}0}}\textbf{\scriptsize{}$\!\!\!$} & \textbf{\textcolor{black}{\scriptsize{}3}} & \textcolor{black}{\scriptsize{}1165} & \textcolor{black}{\scriptsize{}53} & \textcolor{black}{\scriptsize{}55} & \textcolor{black}{\scriptsize{}56.8\%} & \textcolor{black}{\scriptsize{}65.9\%} & {\scriptsize{}0.81m}\tabularnewline
\hline 
\end{tabular}{\scriptsize\par}

\textcolor{black}{\vspace{0.01cm}
}

\raggedright

\emph{\footnotesize{}CLEAR MOT scores and OSPA}\textsuperscript{\emph{\footnotesize{}(2)}}\emph{\footnotesize{}
distance are calculated on standard position estimates ($\uparrow$
means higher is better while $\downarrow$ means lower is better).
Two different detectors are considered - Faster-RCNN(VGG16) (monocular)
and YOLOv3 (monocular). Two types of trackers are considered - MV-GLMB-OC
(multi-view with occlusion model) and MS-GLMB (multi-sensor without
occlusion model). The asterisk ({*}) indicates the multi-camera reconfiguration
experiment.}%
\begin{comment}
\begin{tabular}{|c|c|c|c|c|c|c|c|c|c|c|c|c|}
\hline 
VKS1{*} \tablefootnote{youtube link for VKS1{*}} & 100.0\% & 100.0\% & 100.0\% & 3 & 0 & 0 & 0 & 0 & 0 & 0 & 100.0\% & 100.0\%\tabularnewline
\hline 
\hline 
VKS3{*} \tablefootnote{youtube link for VKS3{*}} & 96.3\% & 96.4\% & 96.1\% & 11 & 0 & 0 & 9 & 16 & 6 & 2 & 98.5\% & 89.4\%\tabularnewline
\hline 
\end{tabular}
\end{comment}

\vspace*{0.08cm}

\label{MOT_Euclidean_CMC1_CMC2_CMC3}
\end{table*}

\noindent 
\begin{table*}
\captionsetup{justification=centering} 

\caption{\small{}CMC1,2,3 Performance Benchmarks for 3D Centroid with Extent Estimates}

\centering

\textcolor{black}{\scriptsize{}}%
\begin{tabular}{|c|c|c|c|c|c|c|c|c|c|c|c|c||c|}
\hline 
\multicolumn{14}{|c|}{\textcolor{black}{\scriptsize{}CMC1 (Maximum/Average 3 people)}}\tabularnewline
\hline 
\textcolor{black}{\scriptsize{}Detector and Tracker}{\scriptsize{}$\!\!\!\!$} & \textcolor{black}{\scriptsize{}IDF1 $\uparrow$} & \textcolor{black}{\scriptsize{}IDP $\uparrow$} & \textcolor{black}{\scriptsize{}IDR $\uparrow$} & {\scriptsize{}$\!\!\!$}\textcolor{black}{\scriptsize{}MT $\uparrow$}{\scriptsize{}$\!\!\!$} & {\scriptsize{}$\!\!\!$}\textcolor{black}{\scriptsize{}PT $\downarrow$}{\scriptsize{}$\!\!\!$} & {\scriptsize{}$\!\!\!$}\textcolor{black}{\scriptsize{}ML $\downarrow$}{\scriptsize{}$\!\!\!$} & \textcolor{black}{\scriptsize{}FP $\downarrow$} & \textcolor{black}{\scriptsize{}FN $\downarrow$} & \textcolor{black}{\scriptsize{}IDs $\downarrow$} & \textcolor{black}{\scriptsize{}FM $\downarrow$} & {\scriptsize{}$\!\!\!$}\textcolor{black}{\scriptsize{}MOTA $\uparrow$}{\scriptsize{}$\!\!\!$} & {\scriptsize{}$\!\!\!$}\textcolor{black}{\scriptsize{}MOTP $\uparrow$}{\scriptsize{}$\!\!\!$} & {\scriptsize{}$\!\!$$\mathrm{OSPA^{(2)}}$}\textcolor{black}{\scriptsize{}$\downarrow$}{\scriptsize{}$\!\!$}\tabularnewline
\hline 
\multirow{1}{*}{\textcolor{black}{\scriptsize{}YOLOv3+MV-GLMB-OC}{\scriptsize{}$\!\!\!\!$}} & \textbf{\textcolor{black}{\scriptsize{}99.7\%}} & \textbf{\textcolor{black}{\scriptsize{}99.4\%}} & \textbf{\textcolor{black}{\scriptsize{}100\%}} & \textbf{\scriptsize{}$\!\!\!$}\textbf{\textcolor{black}{\scriptsize{}3}}\textbf{\scriptsize{}$\!\!\!$} & \textbf{\textcolor{black}{\scriptsize{}0}} & \textbf{\textcolor{black}{\scriptsize{}0}} & \textbf{\textcolor{black}{\scriptsize{}4}} & \textbf{\textcolor{black}{\scriptsize{}0}} & \textbf{\textcolor{black}{\scriptsize{}0}} & \textbf{\textcolor{black}{\scriptsize{}0}} & \textbf{\textcolor{black}{\scriptsize{}99.4\%}} & \textbf{\textcolor{black}{\scriptsize{}67.8\%}} & \textbf{\scriptsize{}0.20}\tabularnewline
\hline 
\multirow{1}{*}{\textcolor{black}{\scriptsize{}YOLOv3+MV-GLMB-OC{*}}{\scriptsize{}$\!\!\!\!$}} & \textcolor{black}{\scriptsize{}98.9\%} & \textcolor{black}{\scriptsize{}97.9\%} & \textcolor{black}{\scriptsize{}99.8\%} & \textbf{\scriptsize{}$\!\!\!$}\textbf{\textcolor{black}{\scriptsize{}3}}\textbf{\scriptsize{}$\!\!\!$} & \textbf{\textcolor{black}{\scriptsize{}0}} & \textbf{\textcolor{black}{\scriptsize{}0}} & \textcolor{black}{\scriptsize{}14} & \textcolor{black}{\scriptsize{}1} & \textbf{\textcolor{black}{\scriptsize{}0}} & \textbf{\textcolor{black}{\scriptsize{}0}} & \textcolor{black}{\scriptsize{}97.7\%} & \textcolor{black}{\scriptsize{}66.7\%} & \textbf{\scriptsize{}0.20}\tabularnewline
\hline 
\multirow{1}{*}{\textcolor{black}{\scriptsize{}YOLOv3+MS-GLMB }{\scriptsize{}$\!\!\!\!$}} & \textcolor{black}{\scriptsize{}95.9\%} & \textcolor{black}{\scriptsize{}92.3\%} & \textcolor{black}{\scriptsize{}99.8\%} & \textbf{\scriptsize{}$\!\!\!$}\textbf{\textcolor{black}{\scriptsize{}3}}\textbf{\scriptsize{}$\!\!\!$} & \textbf{\textcolor{black}{\scriptsize{}0}} & \textbf{\textcolor{black}{\scriptsize{}0}} & \textcolor{black}{\scriptsize{}55} & \textcolor{black}{\scriptsize{}1} & \textcolor{black}{\scriptsize{}1} & \textbf{\textcolor{black}{\scriptsize{}0}} & \textcolor{black}{\scriptsize{}91.3\%} & \textcolor{black}{\scriptsize{}67.5\%} & {\scriptsize{}0.40}\tabularnewline
\hline 
\multirow{1}{*}{\textcolor{black}{\scriptsize{}Faster-RCNN(VGG16)+MV-GLMB-OC}{\scriptsize{}$\!\!\!\!$}} & \textcolor{black}{\scriptsize{}99.5\%} & \textcolor{black}{\scriptsize{}99.1\%} & \textbf{\textcolor{black}{\scriptsize{}100\%}} & \textbf{\scriptsize{}$\!\!\!$}\textbf{\textcolor{black}{\scriptsize{}3}}\textbf{\scriptsize{}$\!\!\!$} & \textbf{\textcolor{black}{\scriptsize{}0}} & \textbf{\textcolor{black}{\scriptsize{}0}} & \textcolor{black}{\scriptsize{}6} & \textbf{\textcolor{black}{\scriptsize{}0}} & \textbf{\textcolor{black}{\scriptsize{}0}} & \textbf{\textcolor{black}{\scriptsize{}0}} & \textcolor{black}{\scriptsize{}99.1\%} & \textcolor{black}{\scriptsize{}67.5\%} & \textbf{\scriptsize{}0.20}\tabularnewline
\hline 
\multirow{1}{*}{\textcolor{black}{\scriptsize{}Faster-RCNN(VGG16)+MV-GLMB-OC{*}}{\scriptsize{}$\!\!\!\!$}} & \textcolor{black}{\scriptsize{}95.5\%} & \textcolor{black}{\scriptsize{}91.4\%} & \textbf{\textcolor{black}{\scriptsize{}100\%}} & \textbf{\scriptsize{}$\!\!\!$}\textbf{\textcolor{black}{\scriptsize{}3}}\textbf{\scriptsize{}$\!\!\!$} & \textbf{\textcolor{black}{\scriptsize{}0}} & \textbf{\textcolor{black}{\scriptsize{}0}} & \textcolor{black}{\scriptsize{}62} & \textbf{\textcolor{black}{\scriptsize{}0}} & \textcolor{black}{\scriptsize{}1} & \textbf{\textcolor{black}{\scriptsize{}0}} & \textcolor{black}{\scriptsize{}90.4\%} & \textcolor{black}{\scriptsize{}67.2\%} & \textbf{\scriptsize{}0.20}\tabularnewline
\hline 
\multirow{1}{*}{\textcolor{black}{\scriptsize{}Faster-RCNN(VGG16)+MS-GLMB}{\scriptsize{}$\!\!\!\!$}} & \textcolor{black}{\scriptsize{}99.6\%} & \textcolor{black}{\scriptsize{}99.2\%} & \textbf{\textcolor{black}{\scriptsize{}100\%}} & \textbf{\scriptsize{}$\!\!\!$}\textbf{\textcolor{black}{\scriptsize{}3}}\textbf{\scriptsize{}$\!\!\!$} & \textbf{\textcolor{black}{\scriptsize{}0}} & \textbf{\textcolor{black}{\scriptsize{}0}} & \textcolor{black}{\scriptsize{}5} & \textbf{\textcolor{black}{\scriptsize{}0}} & \textbf{\textcolor{black}{\scriptsize{}0}} & \textbf{\textcolor{black}{\scriptsize{}0}} & \textcolor{black}{\scriptsize{}99.2\%} & \textcolor{black}{\scriptsize{}66.9\%} & {\scriptsize{}0.40}\tabularnewline
\hline 
\end{tabular}{\scriptsize\par}

\textcolor{black}{\scriptsize{}}%
\begin{tabular}{|c|c|c|c|c|c|c|c|c|c|c|c|c||c|}
\hline 
\multicolumn{14}{|c|}{\textcolor{black}{\scriptsize{}CMC2 (Maximum/Average 10 people)}}\tabularnewline
\hline 
\textcolor{black}{\scriptsize{}Detector and Tracker}{\scriptsize{}$\!\!\!\!$} & \textcolor{black}{\scriptsize{}IDF1 $\uparrow$} & \textcolor{black}{\scriptsize{}IDP $\uparrow$} & \textcolor{black}{\scriptsize{}IDR $\uparrow$} & {\scriptsize{}$\!\!\!$}\textcolor{black}{\scriptsize{}MT $\uparrow$}{\scriptsize{}$\!\!\!$} & {\scriptsize{}$\!\!\!$}\textcolor{black}{\scriptsize{}PT $\downarrow$}{\scriptsize{}$\!\!\!$} & {\scriptsize{}$\!\!\!$}\textcolor{black}{\scriptsize{}ML $\downarrow$}{\scriptsize{}$\!\!\!$} & \textcolor{black}{\scriptsize{}FP $\downarrow$} & \textcolor{black}{\scriptsize{}FN $\downarrow$} & \textcolor{black}{\scriptsize{}IDs $\downarrow$} & \textcolor{black}{\scriptsize{}FM $\downarrow$} & {\scriptsize{}$\!\!\!$}\textcolor{black}{\scriptsize{}MOTA $\uparrow$}{\scriptsize{}$\!\!\!$} & {\scriptsize{}$\!\!\!$}\textcolor{black}{\scriptsize{}MOTP $\uparrow$}{\scriptsize{}$\!\!\!$} & {\scriptsize{}$\!\!$$\mathrm{OSPA^{(2)}}$}\textcolor{black}{\scriptsize{}$\downarrow$}{\scriptsize{}$\!\!$}\tabularnewline
\hline 
\multirow{1}{*}{\textcolor{black}{\scriptsize{}YOLOv3+MV-GLMB-OC}{\scriptsize{}$\!\!\!\!$}} & \textbf{\textcolor{black}{\scriptsize{}87.9\%}} & \textbf{\textcolor{black}{\scriptsize{}87.5\%}} & \textbf{\textcolor{black}{\scriptsize{}87.7\%}} & \textbf{\textcolor{black}{\scriptsize{}10}} & \textbf{\textcolor{black}{\scriptsize{}0}} & \textbf{\textcolor{black}{\scriptsize{}0}} & \textbf{\textcolor{black}{\scriptsize{}19}} & \textbf{\textcolor{black}{\scriptsize{}14}} & \textbf{\textcolor{black}{\scriptsize{}8}} & \textbf{\textcolor{black}{\scriptsize{}2}} & \textbf{\textcolor{black}{\scriptsize{}98.0\%}} & \textbf{\textcolor{black}{\scriptsize{}62.3\%}} & \textbf{\scriptsize{}0.32}\tabularnewline
\hline 
\multirow{1}{*}{\textcolor{black}{\scriptsize{}YOLOv3+MV-GLMB-OC{*}}{\scriptsize{}$\!\!\!\!$}} & \textcolor{black}{\scriptsize{}87.3\%} & \textcolor{black}{\scriptsize{}87.1\%} & \textcolor{black}{\scriptsize{}87.5\%} & \textbf{\textcolor{black}{\scriptsize{}10}} & \textbf{\textcolor{black}{\scriptsize{}0}} & \textbf{\textcolor{black}{\scriptsize{}0}} & \textcolor{black}{\scriptsize{}53} & \textcolor{black}{\scriptsize{}44} & \textcolor{black}{\scriptsize{}14} & \textcolor{black}{\scriptsize{}12} & \textcolor{black}{\scriptsize{}94.7\%} & \textcolor{black}{\scriptsize{}57.0\%} & {\scriptsize{}0.38}\tabularnewline
\hline 
\multirow{1}{*}{\textcolor{black}{\scriptsize{}YOLOv3+MS-GLMB}{\scriptsize{}$\!\!\!\!$}} & \textcolor{black}{\scriptsize{}59.4\%} & \textcolor{black}{\scriptsize{}69.9\%} & \textcolor{black}{\scriptsize{}51.7\%} & \textcolor{black}{\scriptsize{}4} & \textcolor{black}{\scriptsize{}6} & \textbf{\textcolor{black}{\scriptsize{}0}} & \textcolor{black}{\scriptsize{}21} & \textcolor{black}{\scriptsize{}563} & \textcolor{black}{\scriptsize{}30} & \textcolor{black}{\scriptsize{}31} & \textcolor{black}{\scriptsize{}70.4\%} & \textcolor{black}{\scriptsize{}55.7\%} & {\scriptsize{}0.62}\tabularnewline
\hline 
\multirow{1}{*}{\textcolor{black}{\scriptsize{}Faster-RCNN(VGG16)+MV-GLMB-OC}{\scriptsize{}$\!\!\!\!$}} & \textcolor{black}{\scriptsize{}86.7\%} & \textcolor{black}{\scriptsize{}86.5\%} & \textcolor{black}{\scriptsize{}87.0\%} & \textbf{\textcolor{black}{\scriptsize{}10}} & \textbf{\textcolor{black}{\scriptsize{}0}} & \textbf{\textcolor{black}{\scriptsize{}0}} & \textcolor{black}{\scriptsize{}68} & \textcolor{black}{\scriptsize{}55} & \textcolor{black}{\scriptsize{}10} & \textcolor{black}{\scriptsize{}8} & \textcolor{black}{\scriptsize{}93.6\%} & \textcolor{black}{\scriptsize{}60.9\%} & {\scriptsize{}0.34}\tabularnewline
\hline 
\multirow{1}{*}{\textcolor{black}{\scriptsize{}Faster-RCNN(VGG16)+MV-GLMB-OC{*}}{\scriptsize{}$\!\!\!\!$}} & \textcolor{black}{\scriptsize{}81.3\%} & \textcolor{black}{\scriptsize{}80.2\%} & \textcolor{black}{\scriptsize{}82.5\%} & \textbf{\textcolor{black}{\scriptsize{}10}} & \textbf{\textcolor{black}{\scriptsize{}0}} & \textbf{\textcolor{black}{\scriptsize{}0}} & \textcolor{black}{\scriptsize{}127} & \textcolor{black}{\scriptsize{}67} & \textcolor{black}{\scriptsize{}33} & \textcolor{black}{\scriptsize{}15} & \textcolor{black}{\scriptsize{}89.1\%} & \textcolor{black}{\scriptsize{}55.0\%} & {\scriptsize{}0.45}\tabularnewline
\hline 
\multirow{1}{*}{\textcolor{black}{\scriptsize{}Faster-RCNN(VGG16)+MS-GLMB}{\scriptsize{}$\!\!\!\!$}} & \textcolor{black}{\scriptsize{}68.6\%} & \textcolor{black}{\scriptsize{}74.6\%} & \textcolor{black}{\scriptsize{}63.5\%} & \textcolor{black}{\scriptsize{}7} & \textcolor{black}{\scriptsize{}3} & \textbf{\textcolor{black}{\scriptsize{}0}} & \textcolor{black}{\scriptsize{}23} & \textcolor{black}{\scriptsize{}332} & \textcolor{black}{\scriptsize{}23} & \textcolor{black}{\scriptsize{}21} & \textcolor{black}{\scriptsize{}81.8\%} & \textcolor{black}{\scriptsize{}57.1\%} & {\scriptsize{}0.52}\tabularnewline
\hline 
\end{tabular}{\scriptsize\par}

\textcolor{black}{\scriptsize{}}%
\begin{tabular}{|c|c|c|c|c|c|c|c|c|c|c|c|c||c|}
\hline 
\multicolumn{14}{|c|}{\textcolor{black}{\scriptsize{}CMC3 (Maximum/Average 15 people)}}\tabularnewline
\hline 
\textcolor{black}{\scriptsize{}Detector and Tracker}{\scriptsize{}$\!\!\!\!$} & \textcolor{black}{\scriptsize{}IDF1 $\uparrow$} & \textcolor{black}{\scriptsize{}IDP $\uparrow$} & \textcolor{black}{\scriptsize{}IDR $\uparrow$} & {\scriptsize{}$\!\!\!$}\textcolor{black}{\scriptsize{}MT $\uparrow$}{\scriptsize{}$\!\!\!$} & {\scriptsize{}$\!\!\!$}\textcolor{black}{\scriptsize{}PT $\downarrow$}{\scriptsize{}$\!\!\!$} & {\scriptsize{}$\!\!\!$}\textcolor{black}{\scriptsize{}ML $\downarrow$}{\scriptsize{}$\!\!\!$} & \textcolor{black}{\scriptsize{}FP $\downarrow$} & \textcolor{black}{\scriptsize{}FN $\downarrow$} & \textcolor{black}{\scriptsize{}IDs $\downarrow$} & \textcolor{black}{\scriptsize{}FM $\downarrow$} & {\scriptsize{}$\!\!\!$}\textcolor{black}{\scriptsize{}MOTA $\uparrow$}{\scriptsize{}$\!\!\!$} & {\scriptsize{}$\!\!\!$}\textcolor{black}{\scriptsize{}MOTP $\uparrow$}{\scriptsize{}$\!\!\!$} & {\scriptsize{}$\!\!$$\mathrm{OSPA^{(2)}}$}\textcolor{black}{\scriptsize{}$\downarrow$}{\scriptsize{}$\!\!$}\tabularnewline
\hline 
\multirow{1}{*}{\textcolor{black}{\scriptsize{}YOLOv3+MV-GLMB-OC}{\scriptsize{}$\!\!\!\!$}} & \textbf{\textcolor{black}{\scriptsize{}70.7\%}} & \textbf{\textcolor{black}{\scriptsize{}72.3\%}} & \textbf{\textcolor{black}{\scriptsize{}69.1\%}} & \textbf{\textcolor{black}{\scriptsize{}14}} & \textbf{\textcolor{black}{\scriptsize{}1}} & \textbf{\textcolor{black}{\scriptsize{}0}} & \textcolor{black}{\scriptsize{}94} & \textbf{\textcolor{black}{\scriptsize{}222}} & \textbf{\textcolor{black}{\scriptsize{}45}} & \textbf{\textcolor{black}{\scriptsize{}37}} & \textbf{\textcolor{black}{\scriptsize{}87.2\%}} & \textbf{\textcolor{black}{\scriptsize{}52.8\%}} & \textbf{\scriptsize{}0.53}\tabularnewline
\hline 
\multirow{1}{*}{\textcolor{black}{\scriptsize{}YOLOv3+MV-GLMB-OC{*}}{\scriptsize{}$\!\!\!\!$}} & \textcolor{black}{\scriptsize{}60.8\%} & \textcolor{black}{\scriptsize{}65.7\%} & \textcolor{black}{\scriptsize{}56.6\%} & \textcolor{black}{\scriptsize{}9} & \textcolor{black}{\scriptsize{}6} & \textbf{\textcolor{black}{\scriptsize{}0}} & \textcolor{black}{\scriptsize{}91} & \textcolor{black}{\scriptsize{}481} & \textcolor{black}{\scriptsize{}66} & \textcolor{black}{\scriptsize{}56} & \textcolor{black}{\scriptsize{}77.4\%} & \textcolor{black}{\scriptsize{}46.4\%} & {\scriptsize{}0.60}\tabularnewline
\hline 
\multirow{1}{*}{\textcolor{black}{\scriptsize{}YOLOv3+MS-GLMB}{\scriptsize{}$\!\!\!\!$}} & \textcolor{black}{\scriptsize{}41.4\%} & \textcolor{black}{\scriptsize{}57.3\%} & \textcolor{black}{\scriptsize{}32.4\%} & \textcolor{black}{\scriptsize{}0} & \textcolor{black}{\scriptsize{}15} & \textbf{\textcolor{black}{\scriptsize{}0}} & \textbf{\textcolor{black}{\scriptsize{}10}} & \textcolor{black}{\scriptsize{}1239} & \textcolor{black}{\scriptsize{}64} & \textcolor{black}{\scriptsize{}60} & \textcolor{black}{\scriptsize{}53.5\%} & \textcolor{black}{\scriptsize{}46.7\%} & {\scriptsize{}0.76}\tabularnewline
\hline 
\multirow{1}{*}{\textcolor{black}{\scriptsize{}Faster-RCNN(VGG16)+MV-GLMB-OC}{\scriptsize{}$\!\!\!\!$}} & \textcolor{black}{\scriptsize{}63.7\%} & \textcolor{black}{\scriptsize{}66.6\%} & \textcolor{black}{\scriptsize{}61.1\%} & \textcolor{black}{\scriptsize{}12} & \textcolor{black}{\scriptsize{}3} & \textbf{\textcolor{black}{\scriptsize{}0}} & \textcolor{black}{\scriptsize{}97} & \textcolor{black}{\scriptsize{}329} & \textcolor{black}{\scriptsize{}63} & \textcolor{black}{\scriptsize{}41} & \textcolor{black}{\scriptsize{}82.7\%} & \textbf{\textcolor{black}{\scriptsize{}52.8\%}} & {\scriptsize{}0.58}\tabularnewline
\hline 
\multirow{1}{*}{\textcolor{black}{\scriptsize{}Faster-RCNN(VGG16)+MV-GLMB-OC{*}}{\scriptsize{}$\!\!\!\!$}} & \textcolor{black}{\scriptsize{}57.3\%} & \textcolor{black}{\scriptsize{}61.0\%} & \textcolor{black}{\scriptsize{}54.0\%} & \textcolor{black}{\scriptsize{}10} & \textcolor{black}{\scriptsize{}5} & \textbf{\textcolor{black}{\scriptsize{}0}} & \textcolor{black}{\scriptsize{}133} & \textcolor{black}{\scriptsize{}460} & \textcolor{black}{\scriptsize{}78} & \textcolor{black}{\scriptsize{}60} & \textcolor{black}{\scriptsize{}76.3\%} & \textcolor{black}{\scriptsize{}47.9\%} & {\scriptsize{}0.66}\tabularnewline
\hline 
\multirow{1}{*}{\textcolor{black}{\scriptsize{}Faster-RCNN(VGG16)+MS-GLMB}{\scriptsize{}$\!\!\!\!$}} & \textcolor{black}{\scriptsize{}45.7\%} & \textcolor{black}{\scriptsize{}61.7\%} & \textcolor{black}{\scriptsize{}36.3\%} & \textcolor{black}{\scriptsize{}0} & \textcolor{black}{\scriptsize{}15} & \textbf{\textcolor{black}{\scriptsize{}0}} & \textcolor{black}{\scriptsize{}13} & \textcolor{black}{\scriptsize{}1175} & \textcolor{black}{\scriptsize{}61} & \textcolor{black}{\scriptsize{}67} & \textcolor{black}{\scriptsize{}55.8\%} & \textcolor{black}{\scriptsize{}46.6\%} & {\scriptsize{}0.75}\tabularnewline
\hline 
\end{tabular}{\scriptsize\par}

\label{MOT_GIoU_CMC1_CMC2_CMC3-1}\smallskip{}

\raggedright

\emph{\footnotesize{}CLEAR MOT scores and OSPA}\textsuperscript{\emph{\footnotesize{}(2)}}\emph{\footnotesize{}
distance are calculated with a 3D GIoU base-distance for estimates
of 3D centroid with extent ($\uparrow$ means higher is better while
$\downarrow$ means lower is better). Two different detectors are
considered - Faster-RCNN(VGG16) (monocular) and YOLOv3 (monocular).
Two types of trackers are considered - MV-GLMB-OC (multi-view with
occlusion model) and MS-GLMB (multi-sensor without occlusion model).
The asterisk ({*}) indicates the multi-camera reconfiguration experiment}\emph{\small{}.}%
\begin{comment}
\begin{tabular}{|c|c|c|c|c|c|c|c|c|c|c|c|c|}
\hline 
VKS1{*} \tablefootnote{youtube link for VKS1{*}} & 100.0\% & 100.0\% & 100.0\% & 3 & 0 & 0 & 0 & 0 & 0 & 0 & 100.0\% & 100.0\%\tabularnewline
\hline 
\hline 
VKS3{*} \tablefootnote{youtube link for VKS3{*}} & 96.3\% & 96.4\% & 96.1\% & 11 & 0 & 0 & 9 & 16 & 6 & 2 & 98.5\% & 89.4\%\tabularnewline
\hline 
\end{tabular}
\end{comment}

\vspace*{-0.4cm}
\end{table*}

\noindent \vspace*{-1.2cm}

\subsection{CMC1, CMC2 and CMC3 \label{subsec:CMC-Dataset-and}}

\noindent 
\begin{figure*}[t]
\includegraphics[width=0.5\columnwidth]{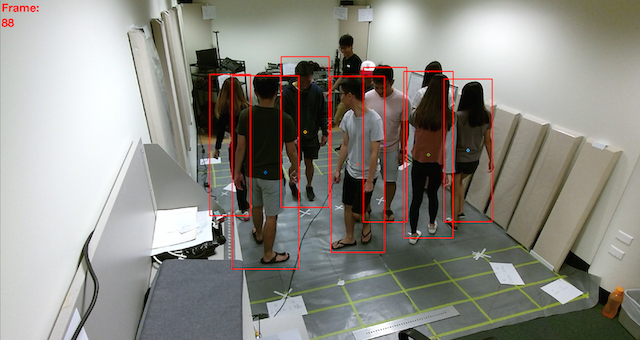}\includegraphics[width=0.5\columnwidth]{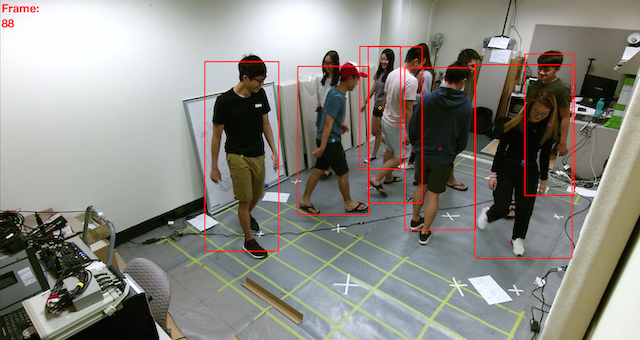}\includegraphics[width=0.5\columnwidth]{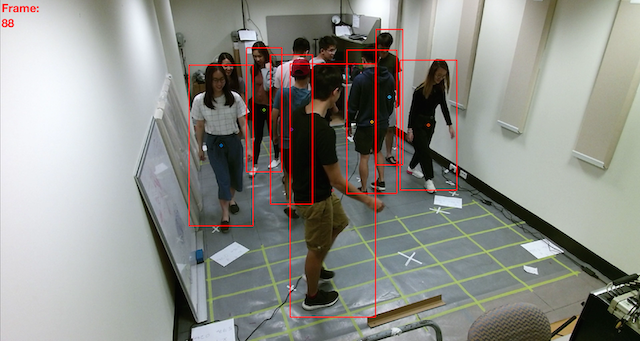}\includegraphics[width=0.5\columnwidth]{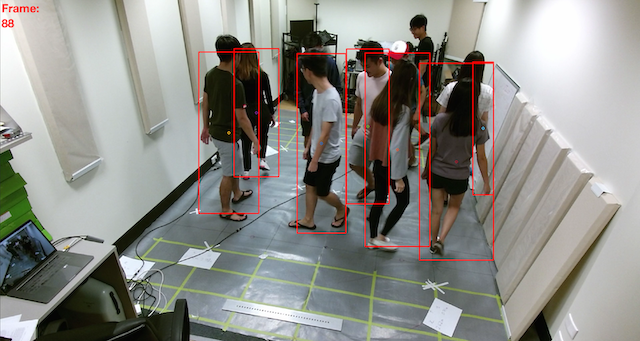}

\includegraphics[width=0.5\columnwidth]{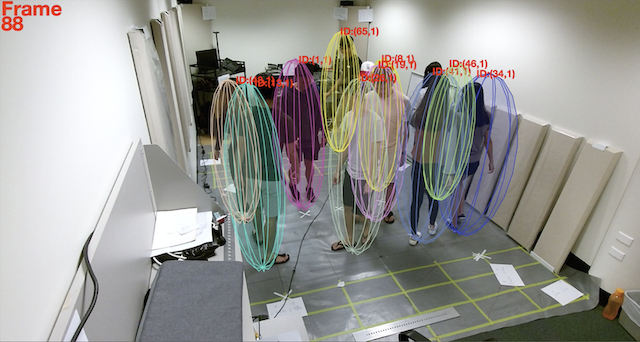}\includegraphics[width=0.5\columnwidth]{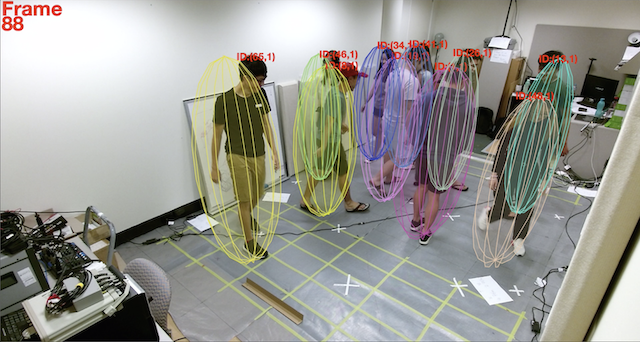}\includegraphics[width=0.5\columnwidth]{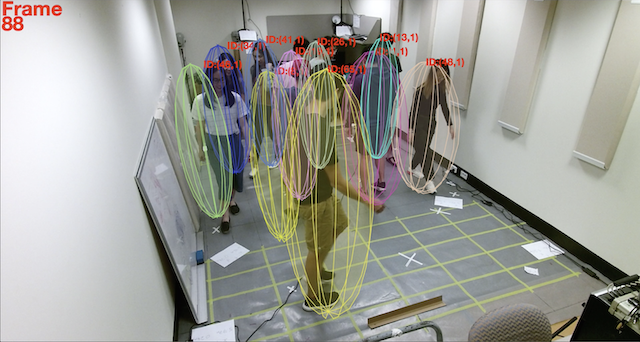}\includegraphics[width=0.5\columnwidth]{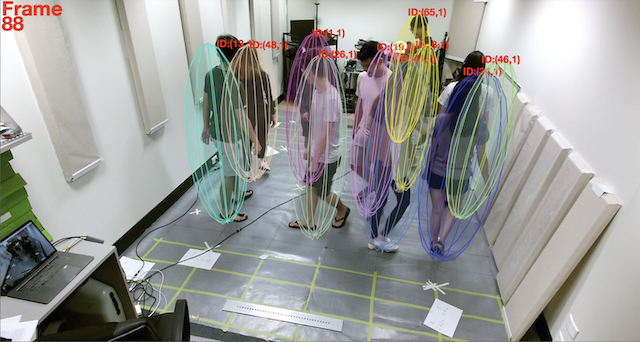}

\smallskip{}

\caption{CMC2 Camera 1 to 4 (left to right): YOLOv3 detections (top row) and
MV-GLMB-OC estimates (bottom row).}
\label{CMC_2_figures}
\end{figure*}

\noindent \vspace*{-0.8cm}

\noindent 
\begin{figure*}[t]
\centering

\includegraphics[width=0.5\columnwidth]{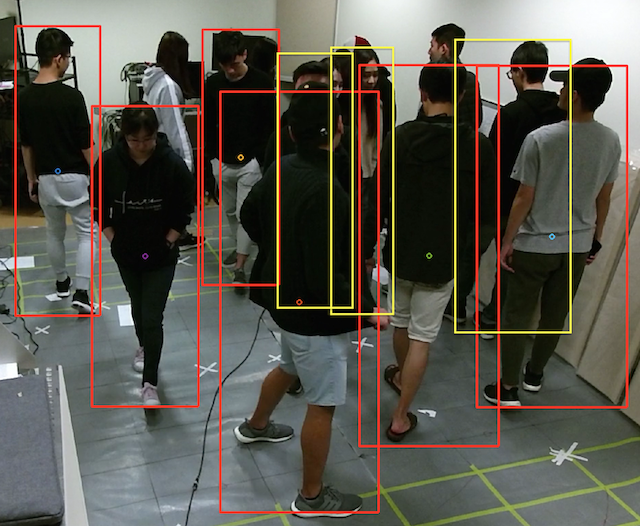}\includegraphics[width=0.45\columnwidth]{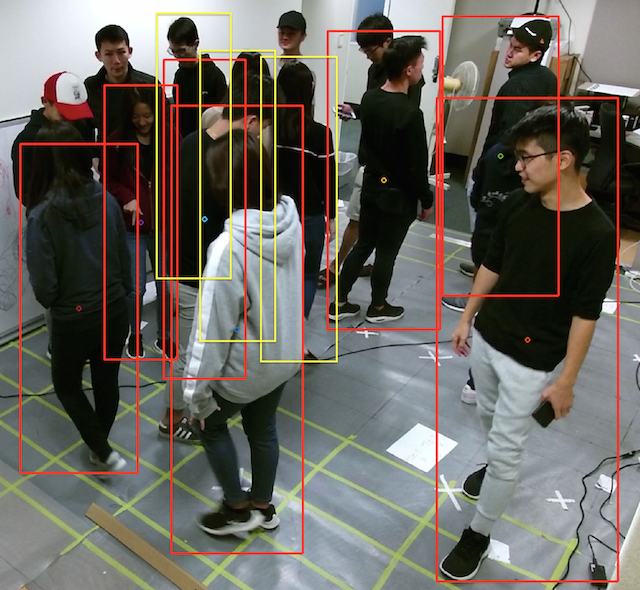}\includegraphics[width=0.5\columnwidth]{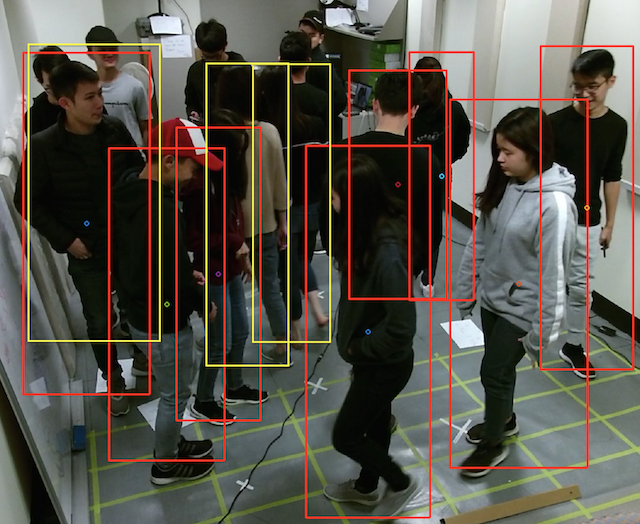}\includegraphics[width=0.47\columnwidth]{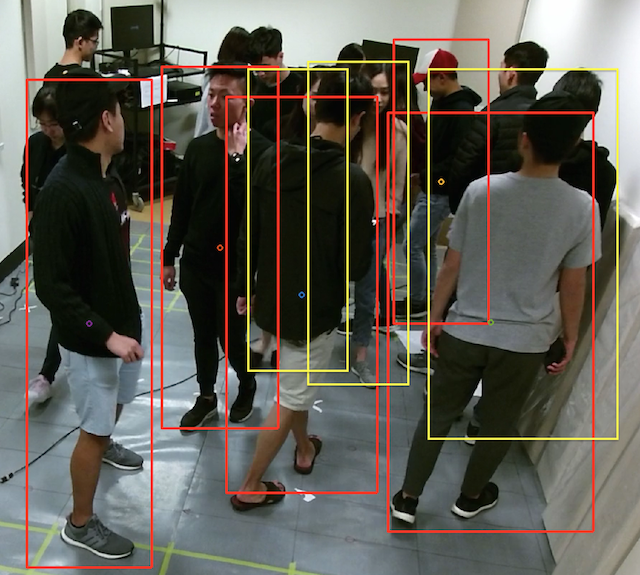}

\smallskip{}

\caption{CMC3 Camera 1 to 4 (left to right): YOLOv3 detections (red bounding
boxes) and people that are occluded in all four cameras (yellow bounding
boxes).}
\label{CMC3_figures}
\end{figure*}

\noindent \vspace*{-0.8cm}

This subsection focuses on scenarios with people walking in order
of increasing difficulty, i.e. CMC1-CMC3. Similar to the WILDTRACKS
evaluation, we evaluate our method based on 2 monocular detectors,
namely Faster-RCNN(VGG16) and YOLOv3. For each sequence, the effect
of the occlusion model is studied by comparing the proposed MV-GLMB-OC
with the standard MS-GLMB filter. 

\subsubsection{Model Parameters}

Unlike WILDTRACKS where objects enter the scene from anywhere at the
boundary, in CMC we know the location of objects entering the scene.
Hence, we specify the birth parameters as $P_{B,+}(\ell)=0.001$ and
$f_{B,+}(x,\ell)=\mathcal{N}(x;\mu_{B,+},0.1^{2}\mathrm{I}_{9})$
where
\begin{align*}
\mu_{B,+}= & [2.03\;0\;0.71\;0\;0.825\;0\;-\negmedspace1.2\;-\negmedspace1.2\;-\negmedspace0.18]^{T}.
\end{align*}
We use the single-object transition density (\ref{eq:transition_density})
with position noise and extent (in logarithm) noise parameterized
by:
\begin{align*}
\upsilon^{(p)}=[0.0012,0.0012,0.0012]^{T},\\
\upsilon^{(s)}=[0.0036,0.0036,0.0004]^{T}.
\end{align*}

\subsubsection{Effectiveness of Occlusion Model \label{subsec:Discussion-on-Ablation}}

Table \ref{MOT_Euclidean_CMC1_CMC2_CMC3} shows the CLEAR MOT and
OSPA\textsuperscript{(2)} benchmarks with a Euclidean base-distance,
for the estimated 3D centroids only. Table \ref{MOT_GIoU_CMC1_CMC2_CMC3-1}
shows the CLEAR MOT and OSPA\textsuperscript{(2)} benchmarks with
a 3D GIoU base-distance, for the estimated 3D centroids and extent.
Both tables compare the tracking performance with and without and
occlusion model, i.e. MV-GLMB-OC and MS-GLMB respectively. The asterisked
entry denotes the multi-camera reconfiguration case which is discussed
later on. All results are presented for two different detectors YOLOv3
and Faster-RCNN(VGG16).

We focus our initial examination on the non-asterisked entries in
Tables \ref{MOT_Euclidean_CMC1_CMC2_CMC3} and \ref{MOT_GIoU_CMC1_CMC2_CMC3-1}.
This corresponds to the case where all cameras are operational. For
the sparse scenario CMC1, both MV-GLMB-OC and MS-GLMB on either detectors
achieved a close to perfect CLEAR MOT scores in MOTA and MOTP. Some
of the flagged FPs are caused by track initiation/termination mismatches
with the ground truths (annotations). The OSPA\textsuperscript{(2)}
values are relatively low due to the sparsity of the scenario.

For the medium scenario CMC2, Fig. \ref{CMC_2_figures} shows a screenshot
of the detections and the MV-GLMB-OC estimates. In this case, MV-GLMB-OC
on both detectors managed to maintain consistent tracks and accurate
estimates overall. The CLEAR MOT benchmarks for CMC2 show high MOTA
and MOTP but with some FNs and FPs. We observe an improvement in performance
for MV-GLMB-OC over MS-GLMB, and on both detectors due to the inclusion
of occlusion modeling. The improvement in performance due to occlusion
modeling is also reflected in the OSPA\textsuperscript{(2)}.

For the dense scenario CMC3, MV-GLMB-OC on both detectors managed
to achieve acceptable MOTA/MOTP scores, but is penalized with high
FPs, FNs, IDs and FMs. This outcome occurs even with the proposed
occlusion model, as the algorithm fails when a person is totally occluded
in all views. An example of this occurrence is illustrated in Fig.
\ref{CMC3_figures}, where the red bounding boxes denote detections,
while the yellow bounding boxes indicate people who are undetected
in all views. Such an event could cause track termination/switching
and is reflected in the performance evaluation. It is evident from
Tables \ref{MOT_Euclidean_CMC1_CMC2_CMC3} and \ref{MOT_GIoU_CMC1_CMC2_CMC3-1}
that the tracking performance improves considerably with the occlusion
model. Examination of the OSPA\textsuperscript{(2)} error leads to
a similar conclusion.

Overall, YOLOv3+MV-GLMB-OC performs slightly better than Faster-RCNN(VGG16)+MV-GLMB-OC
due to better detections. The tracking performance of the proposed
MV-GLMB-OC filter generally degrades as the number of people in the
scene increases, since the visual occlusions become more frequent
and more difficult to resolve. The results of this study on the proposed
occlusion model suggest that without proper modeling of the probability
of detection, the algorithm fails to maintain tracks, resulting in
poorer tracking results. The CLEAR evaluation for the monocular detectors
used are given in Appendix \ref{subsec:Monocular-Detector-Results}.

\noindent 
\begin{figure*}
\includegraphics[width=2\columnwidth]{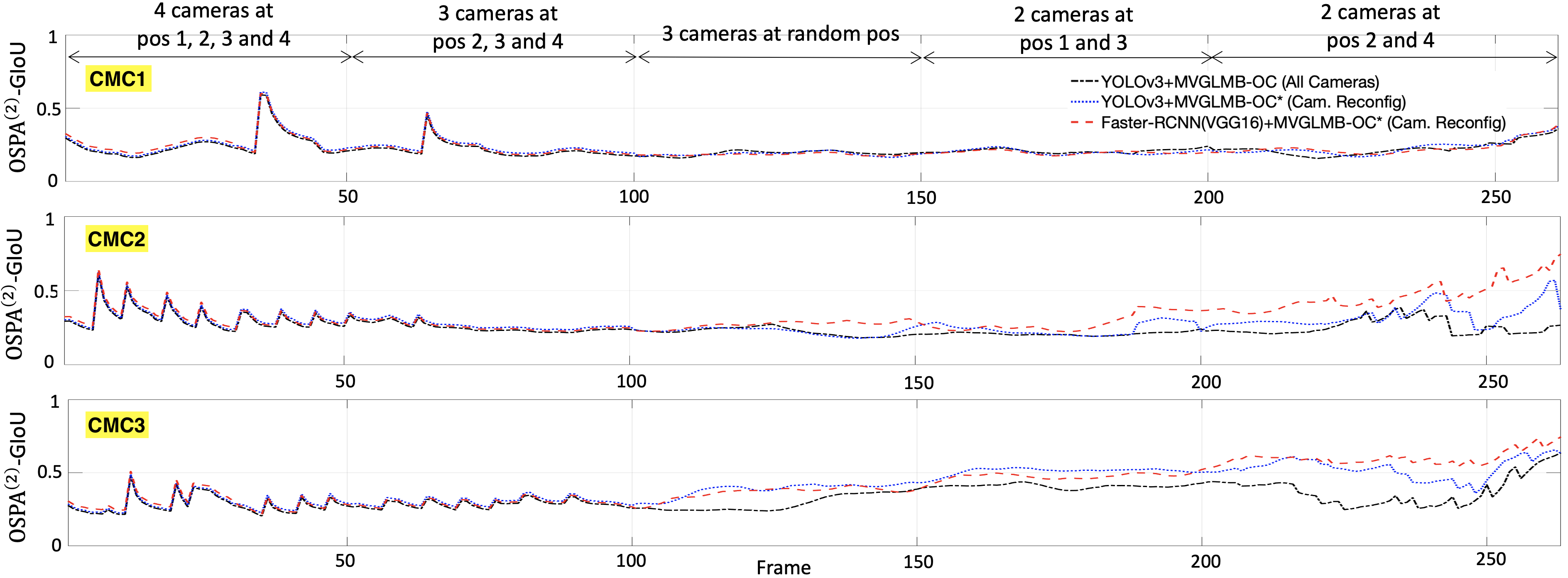}

\caption{Multi-Camera Reconfiguration Experiment: OSPA\protect\textsuperscript{(2)}
plots with 3D GIoU base-distance for estimates of 3D centroid with
extent. Three trackers are considered: YOLOv3+MV-GLMB-OC{*} (multi-camera
reconfiguration) and Faster-RCNN+MV-GLMB-OC{*} (multi-camera reconfiguration)
and with YOLOv3+MV-GLMB-OC (all cameras operational). }
\label{OSPA_GIoU_CMC1_CMC2_CMC3}
\end{figure*}

\noindent \vspace*{-1.0cm}

\subsubsection{Multi-Camera Reconfiguration \label{subsec:Multi-Camera-Reconfiguration}}

The MV-GLMB-OC approach requires only a one-off training on each monocular
detector, and hence can operate without retraining and without interruption,
in the event that cameras are added, removed or repositioned on the
fly. To demonstrate this capability, we design a multi-camera reconfiguration
experiment. At the start of the sequence, all four cameras are operational.
Later, one camera is taken offline to mimic a camera failure. Subsequently,
two cameras are taken offline to mimic a more severe camera failure.
After this, the two previously offline cameras are made operational,
while the previously operational cameras are taken offline, which
mimics the event that the two operational cameras are moved to different
locations. We benchmark the multi-camera reconfiguration experiment
against the ideal case when all cameras are operational. 

Results for the experiments on multi-camera reconfiguration are denoted
with an asterisk in Tables \ref{MOT_Euclidean_CMC1_CMC2_CMC3} and
\ref{MOT_GIoU_CMC1_CMC2_CMC3-1}. The reported CLEAR MOT scores and
OSPA\textsuperscript{(2)} errors show similar trends in respect of
inclusion of the occlusion model, increasing scenario density, and
relative performance on the two detectors. The tracking performance
in the multi-camera reconfiguration case is generally worse than the
case when all cameras are active. This relative observation is in
line with expectations, as there is less sensor data to resolve occlusions
and perform estimation. 

To facilitate an examination of the relative performance in further
detail, Fig. \ref{OSPA_GIoU_CMC1_CMC2_CMC3} plots the $\mathrm{OSPA}^{(2)}$
error with 3D GIoU base-distance over a sliding window with time.
As a reference point for the performance comparison, the YOLOv3+MV-GLMB-OC
with all cameras operational case is also shown. The spikes in the
error curve at the beginning and the end of the scenario are due to
mismatches in track initiation and termination with the ground truths.
For CMC1, we observe that the error curves are relatively close to
the reference case. This would be expected for a sparse scenario as
there are virtually no occlusions even when some cameras are offline.
For CMC2 and CMC3, the error curves for both YOLOv3+MV-GLMB-OC{*}
and Faster-RCNN(VGG16)+MV-GLMB-OC{*} begin to deviate midway into
sequence from the all cameras operational reference. The errors become
more pronounced entering the 2-camera only segment, as the more crowded
scenarios exacerbate the effect of occlusions and misdetections. Nonetheless,
the results show that the MV-GLMB-OC filter is able to accommodate
on-the-fly changes to the camera configurations.

\subsection{CMC4 and CMC5 (3D Multi-Modal Tracking) \label{subsec:3D-Multi-Modal-Tracking}}

Here we present the first multi-camera dataset with people jumping
and falling, which is more challenging for MOT than scenarios with
only normal walking. We demonstrate the versatility of the proposed
MOT framework by using a Jump Markov System (JMS), to cater for potential
switching between upright and fallen modes \cite{punchihewa2018multiple}. 

\noindent 
\begin{table*}
\captionsetup{justification=centering} 

\caption{\small{}CMC4,5 Performance Benchmarks for 3D Position Estimates }

\centering

\textcolor{black}{\scriptsize{}}%
\begin{tabular}{|c|c|c|c|c|c|c|c|c|c|c|c|c||c|}
\hline 
\multicolumn{14}{|c|}{\textcolor{black}{\scriptsize{}CMC4 (Jumping and Falling, Maximum/Average
3 people)}}\tabularnewline
\hline 
\textcolor{black}{\scriptsize{}Detector and Tracker}{\scriptsize{}$\!\!\!\!$} & \textcolor{black}{\scriptsize{}IDF1 $\uparrow$} & \textcolor{black}{\scriptsize{}IDP $\uparrow$} & \textcolor{black}{\scriptsize{}IDR $\uparrow$} & {\scriptsize{}$\!\!\!$}\textcolor{black}{\scriptsize{}MT $\uparrow$}{\scriptsize{}$\!\!\!$} & {\scriptsize{}$\!\!\!$}\textcolor{black}{\scriptsize{}PT $\downarrow$}{\scriptsize{}$\!\!\!$} & {\scriptsize{}$\!\!\!$}\textcolor{black}{\scriptsize{}ML $\downarrow$}{\scriptsize{}$\!\!\!$} & \textcolor{black}{\scriptsize{}FP $\downarrow$} & \textcolor{black}{\scriptsize{}FN $\downarrow$} & \textcolor{black}{\scriptsize{}IDs $\downarrow$} & \textcolor{black}{\scriptsize{}FM $\downarrow$} & {\scriptsize{}$\!\!\!$}\textcolor{black}{\scriptsize{}MOTA $\uparrow$}{\scriptsize{}$\!\!\!$} & {\scriptsize{}$\!\!\!$}\textcolor{black}{\scriptsize{}MOTP $\uparrow$}{\scriptsize{}$\!\!\!$} & {\scriptsize{}$\!\!$$\mathrm{OSPA^{(2)}}$}\textcolor{black}{\scriptsize{}$\downarrow$}{\scriptsize{}$\!\!$}\tabularnewline
\hline 
\multirow{1}{*}{\textcolor{black}{\scriptsize{}YOLOv3+MV-GLMB-OC}} & \textbf{\textcolor{black}{\scriptsize{}99.3\%}} & \textbf{\textcolor{black}{\scriptsize{}99.0\%}} & \textbf{\textcolor{black}{\scriptsize{}99.5\%}} & \textbf{\textcolor{black}{\scriptsize{}3}} & \textbf{\textcolor{black}{\scriptsize{}0}} & \textbf{\textcolor{black}{\scriptsize{}0}} & \textcolor{black}{\scriptsize{}4} & \textcolor{black}{\scriptsize{}2} & \textbf{\textcolor{black}{\scriptsize{}0}} & \textbf{\textcolor{black}{\scriptsize{}0}} & \textbf{\textcolor{black}{\scriptsize{}98.5\%}} & \textbf{\textcolor{black}{\scriptsize{}89.5\%}} & \textbf{\scriptsize{}0.16m}\tabularnewline
\hline 
\multirow{1}{*}{\textcolor{black}{\scriptsize{}YOLOv3+MV-GLMB-OC{*}}} & \textcolor{black}{\scriptsize{}95.0\%} & \textcolor{black}{\scriptsize{}93.5\%} & \textcolor{black}{\scriptsize{}96.5\%} & \textbf{\textcolor{black}{\scriptsize{}3}} & \textbf{\textcolor{black}{\scriptsize{}0}} & \textbf{\textcolor{black}{\scriptsize{}0}} & \textcolor{black}{\scriptsize{}17} & \textcolor{black}{\scriptsize{}4} & \textcolor{black}{\scriptsize{}5} & \textcolor{black}{\scriptsize{}1} & \textcolor{black}{\scriptsize{}93.6\%} & \textcolor{black}{\scriptsize{}87.7\%} & {\scriptsize{}0.18m}\tabularnewline
\hline 
\multirow{1}{*}{\textcolor{black}{\scriptsize{}YOLOv3+MS-GLMB}} & \textcolor{black}{\scriptsize{}95.9\%} & \textcolor{black}{\scriptsize{}94.0\%} & \textcolor{black}{\scriptsize{}97.8\%} & \textbf{\textcolor{black}{\scriptsize{}3}} & \textbf{\textcolor{black}{\scriptsize{}0}} & \textbf{\textcolor{black}{\scriptsize{}0}} & \textcolor{black}{\scriptsize{}21} & \textcolor{black}{\scriptsize{}5} & \textcolor{black}{\scriptsize{}4} & \textcolor{black}{\scriptsize{}1} & \textcolor{black}{\scriptsize{}92.6\%} & \textcolor{black}{\scriptsize{}86.4\%} & {\scriptsize{}0.21m}\tabularnewline
\hline 
\multirow{1}{*}{\textcolor{black}{\scriptsize{}Faster-RCNN(VGG16)+MV-GLMB-OC}} & \textcolor{black}{\scriptsize{}98.0\%} & \textcolor{black}{\scriptsize{}98.5\%} & \textcolor{black}{\scriptsize{}97.5\%} & \textbf{\textcolor{black}{\scriptsize{}3}} & \textbf{\textcolor{black}{\scriptsize{}0}} & \textbf{\textcolor{black}{\scriptsize{}0}} & \textbf{\textcolor{black}{\scriptsize{}3}} & \textcolor{black}{\scriptsize{}7} & \textcolor{black}{\scriptsize{}2} & \textcolor{black}{\scriptsize{}1} & \textcolor{black}{\scriptsize{}97.0\%} & \textcolor{black}{\scriptsize{}87.1\%} & {\scriptsize{}0.18m}\tabularnewline
\hline 
\multirow{1}{*}{\textcolor{black}{\scriptsize{}Faster-RCNN(VGG16)+MV-GLMB-OC{*}}} & \textcolor{black}{\scriptsize{}85.1\%} & \textcolor{black}{\scriptsize{}82.2\%} & \textcolor{black}{\scriptsize{}88.1\%} & \textbf{\textcolor{black}{\scriptsize{}3}} & \textbf{\textcolor{black}{\scriptsize{}0}} & \textbf{\textcolor{black}{\scriptsize{}0}} & \textcolor{black}{\scriptsize{}29} & \textbf{\textcolor{black}{\scriptsize{}0}} & \textcolor{black}{\scriptsize{}6} & \textbf{\textcolor{black}{\scriptsize{}0}} & \textcolor{black}{\scriptsize{}91.3\%} & \textcolor{black}{\scriptsize{}86.6\%} & {\scriptsize{}0.19m}\tabularnewline
\hline 
\multirow{1}{*}{\textcolor{black}{\scriptsize{}Faster-RCNN(VGG16)+MS-GLMB}} & \textcolor{black}{\scriptsize{}89.3\%} & \textcolor{black}{\scriptsize{}85.8\%} & \textcolor{black}{\scriptsize{}93.1\%} & \textbf{\textcolor{black}{\scriptsize{}3}} & \textbf{\textcolor{black}{\scriptsize{}0}} & \textbf{\textcolor{black}{\scriptsize{}0}} & \textcolor{black}{\scriptsize{}34} & \textbf{\textcolor{black}{\scriptsize{}0}} & \textcolor{black}{\scriptsize{}12} & \textbf{\textcolor{black}{\scriptsize{}0}} & \textcolor{black}{\scriptsize{}88.6\%} & \textcolor{black}{\scriptsize{}87.0\%} & {\scriptsize{}0.22m}\tabularnewline
\hline 
\end{tabular}{\scriptsize\par}

\textcolor{black}{\scriptsize{}}%
\begin{tabular}{|c|c|c|c|c|c|c|c|c|c|c|c|c||c|}
\hline 
\multicolumn{14}{|c|}{\textcolor{black}{\scriptsize{}CMC5 (Jumping and Falling, Maximum/Average
7 people)}}\tabularnewline
\hline 
\textcolor{black}{\scriptsize{}Detector and Tracker}{\scriptsize{}$\!\!\!\!$} & \textcolor{black}{\scriptsize{}IDF1 $\uparrow$} & \textcolor{black}{\scriptsize{}IDP $\uparrow$} & \textcolor{black}{\scriptsize{}IDR $\uparrow$} & {\scriptsize{}$\!\!\!$}\textcolor{black}{\scriptsize{}MT $\uparrow$}{\scriptsize{}$\!\!\!$} & {\scriptsize{}$\!\!\!$}\textcolor{black}{\scriptsize{}PT $\downarrow$}{\scriptsize{}$\!\!\!$} & {\scriptsize{}$\!\!\!$}\textcolor{black}{\scriptsize{}ML $\downarrow$}{\scriptsize{}$\!\!\!$} & \textcolor{black}{\scriptsize{}FP $\downarrow$} & \textcolor{black}{\scriptsize{}FN $\downarrow$} & \textcolor{black}{\scriptsize{}IDs $\downarrow$} & \textcolor{black}{\scriptsize{}FM $\downarrow$} & {\scriptsize{}$\!\!\!$}\textcolor{black}{\scriptsize{}MOTA $\uparrow$}{\scriptsize{}$\!\!\!$} & {\scriptsize{}$\!\!\!$}\textcolor{black}{\scriptsize{}MOTP $\uparrow$}{\scriptsize{}$\!\!\!$} & {\scriptsize{}$\!\!$$\mathrm{OSPA^{(2)}}$}\textcolor{black}{\scriptsize{}$\downarrow$}{\scriptsize{}$\!\!$}\tabularnewline
\hline 
\multirow{1}{*}{\textcolor{black}{\scriptsize{}YOLOv3+MV-GLMB-OC}} & \textbf{\textcolor{black}{\scriptsize{}60.5\%}} & \textbf{\textcolor{black}{\scriptsize{}63.5\%}} & \textbf{\textcolor{black}{\scriptsize{}61.3\%}} & \textbf{\textcolor{black}{\scriptsize{}3}} & \textcolor{black}{\scriptsize{}4} & \textbf{\textcolor{black}{\scriptsize{}0}} & \textbf{\scriptsize{}388} & \textbf{\textcolor{black}{\scriptsize{}933}} & \textbf{\textcolor{black}{\scriptsize{}55}} & \textbf{\scriptsize{}47} & \textbf{\textcolor{black}{\scriptsize{}61.1\%}} & \textbf{\textcolor{black}{\scriptsize{}69.3\%}} & \textbf{\scriptsize{}$\!\!$0.63m$\!\!$}\tabularnewline
\hline 
\multirow{1}{*}{\textcolor{black}{\scriptsize{}YOLOv3+MV-GLMB-OC{*}}} & \textcolor{black}{\scriptsize{}59.3\%} & \textcolor{black}{\scriptsize{}58.1\%} & \textcolor{black}{\scriptsize{}60.1\%} & \textbf{\textcolor{black}{\scriptsize{}3}} & \textcolor{black}{\scriptsize{}3} & \textcolor{black}{\scriptsize{}1} & \textcolor{black}{\scriptsize{}418} & \textcolor{black}{\scriptsize{}1172} & \textcolor{black}{\scriptsize{}69} & \textcolor{black}{\scriptsize{}60} & \textcolor{black}{\scriptsize{}56.7\%} & \textcolor{black}{\scriptsize{}63.9\%} & {\scriptsize{}$\!\!$0.69m$\!\!$}\tabularnewline
\hline 
\multirow{1}{*}{\textcolor{black}{\scriptsize{}YOLOv3+MS-GLMB }} & \textcolor{black}{\scriptsize{}50.9\%} & \textcolor{black}{\scriptsize{}51.1\%} & \textcolor{black}{\scriptsize{}47.6\%} & \textbf{\textcolor{black}{\scriptsize{}3}} & \textbf{\textcolor{black}{\scriptsize{}2}} & \textcolor{black}{\scriptsize{}2} & \textcolor{black}{\scriptsize{}735} & \textcolor{black}{\scriptsize{}1699} & \textcolor{black}{\scriptsize{}85} & \textcolor{black}{\scriptsize{}69} & \textcolor{black}{\scriptsize{}50.7\%} & \textcolor{black}{\scriptsize{}59.5\%} & {\scriptsize{}$\!\!$0.79m$\!\!$}\tabularnewline
\hline 
\multirow{1}{*}{\textcolor{black}{\scriptsize{}Faster-RCNN(VGG16)+MV-GLMB-OC}} & \textcolor{black}{\scriptsize{}60.1\%} & \textcolor{black}{\scriptsize{}62.5\%} & \textcolor{black}{\scriptsize{}60.1\%} & \textbf{\textcolor{black}{\scriptsize{}3}} & \textcolor{black}{\scriptsize{}4} & \textbf{\textcolor{black}{\scriptsize{}0}} & \textcolor{black}{\scriptsize{}410} & \textcolor{black}{\scriptsize{}1185} & \textcolor{black}{\scriptsize{}61} & \textcolor{black}{\scriptsize{}49} & \textcolor{black}{\scriptsize{}60.3\%} & \textcolor{black}{\scriptsize{}64.1\%} & {\scriptsize{}$\!\!$0.66m$\!\!$}\tabularnewline
\hline 
\multirow{1}{*}{\textcolor{black}{\scriptsize{}Faster-RCNN(VGG16)+MV-GLMB-OC{*}}} & \textcolor{black}{\scriptsize{}56.2\%} & \textcolor{black}{\scriptsize{}55.6\%} & \textcolor{black}{\scriptsize{}59.2\%} & \textbf{\textcolor{black}{\scriptsize{}3}} & \textcolor{black}{\scriptsize{}3} & \textcolor{black}{\scriptsize{}1} & {\scriptsize{}534} & \textcolor{black}{\scriptsize{}1493} & {\scriptsize{}63} & \textcolor{black}{\scriptsize{}61} & \textcolor{black}{\scriptsize{}55.7\%} & \textcolor{black}{\scriptsize{}63.6\%} & {\scriptsize{}$\!\!$0.70m$\!\!$}\tabularnewline
\hline 
\multirow{1}{*}{\textcolor{black}{\scriptsize{}Faster-RCNN(VGG16)+MS-GLMB}} & \textcolor{black}{\scriptsize{}49.1\%} & \textcolor{black}{\scriptsize{}49.7\%} & \textcolor{black}{\scriptsize{}46.1\%} & \textbf{\textcolor{black}{\scriptsize{}3}} & \textcolor{black}{\scriptsize{}3} & \textcolor{black}{\scriptsize{}1} & \textcolor{black}{\scriptsize{}781} & {\scriptsize{}1337} & {\scriptsize{}92} & \textcolor{black}{\scriptsize{}69} & \textcolor{black}{\scriptsize{}49.6\%} & \textcolor{black}{\scriptsize{}61.6\%} & {\scriptsize{}$\!\!$0.80m$\!\!$}\tabularnewline
\hline 
\end{tabular}{\scriptsize\par}

\textcolor{black}{\vspace{0.01cm}
}

\raggedright

\emph{\footnotesize{}CLEAR MOT scores and OSPA}\textsuperscript{\emph{\footnotesize{}(2)}}\emph{\footnotesize{}
distance are calculated on standard position estimates ($\uparrow$
means higher is better while $\downarrow$ means lower is better).
Two different detectors are considered - Faster-RCNN(VGG16) (monocular)
and YOLOv3 (monocular). Two types of trackers are considered - MV-GLMB-OC
(multi-view with occlusion model) and MS-GLMB (multi-sensor without
occlusion model). The asterisk ({*}) indicates the multi-camera reconfiguration
experiment}\emph{\small{}.}%
\begin{comment}
\begin{tabular}{|c|c|c|c|c|c|c|c|c|c|c|c|c|}
\hline 
VKS1{*} \tablefootnote{youtube link for VKS1{*}} & 100.0\% & 100.0\% & 100.0\% & 3 & 0 & 0 & 0 & 0 & 0 & 0 & 100.0\% & 100.0\%\tabularnewline
\hline 
\hline 
VKS3{*} \tablefootnote{youtube link for VKS3{*}} & 96.3\% & 96.4\% & 96.1\% & 11 & 0 & 0 & 9 & 16 & 6 & 2 & 98.5\% & 89.4\%\tabularnewline
\hline 
\end{tabular}
\end{comment}

\vspace*{0.08cm}

\label{MOT_Euclidean_CMC4_CMC5}
\end{table*}

\noindent 
\begin{table*}
\captionsetup{justification=centering} 

\caption{\small{}CMC4,5 Performance Benchmarks for 3D Centroid with Extent Estimates }

\centering

\textcolor{black}{\scriptsize{}}%
\begin{tabular}{|c|c|c|c|c|c|c|c|c|c|c|c|c||c|}
\hline 
\multicolumn{14}{|c|}{\textcolor{black}{\scriptsize{}CMC4 (Jumping and Falling, Maximum/Average
3 people)}}\tabularnewline
\hline 
\textcolor{black}{\scriptsize{}Detector and Tracker}{\scriptsize{}$\!\!\!\!$} & \textcolor{black}{\scriptsize{}IDF1 $\uparrow$} & \textcolor{black}{\scriptsize{}IDP $\uparrow$} & \textcolor{black}{\scriptsize{}IDR $\uparrow$} & {\scriptsize{}$\!\!\!$}\textcolor{black}{\scriptsize{}MT $\uparrow$}{\scriptsize{}$\!\!\!$} & {\scriptsize{}$\!\!\!$}\textcolor{black}{\scriptsize{}PT $\downarrow$}{\scriptsize{}$\!\!\!$} & {\scriptsize{}$\!\!\!$}\textcolor{black}{\scriptsize{}ML $\downarrow$}{\scriptsize{}$\!\!\!$} & \textcolor{black}{\scriptsize{}FP $\downarrow$} & \textcolor{black}{\scriptsize{}FN $\downarrow$} & \textcolor{black}{\scriptsize{}IDs $\downarrow$} & \textcolor{black}{\scriptsize{}FM $\downarrow$} & {\scriptsize{}$\!\!\!$}\textcolor{black}{\scriptsize{}MOTA $\uparrow$}{\scriptsize{}$\!\!\!$} & {\scriptsize{}$\!\!\!$}\textcolor{black}{\scriptsize{}MOTP $\uparrow$}{\scriptsize{}$\!\!\!$} & {\scriptsize{}$\!\!$$\mathrm{OSPA^{(2)}}$}\textcolor{black}{\scriptsize{}$\downarrow$}{\scriptsize{}$\!\!$}\tabularnewline
\hline 
\multirow{1}{*}{\textcolor{black}{\scriptsize{}YOLOv3+MV-GLMB-OC}} & \textbf{\textcolor{black}{\scriptsize{}99.3\%}} & \textbf{\textcolor{black}{\scriptsize{}99.0\%}} & \textbf{\textcolor{black}{\scriptsize{}99.5\%}} & \textbf{\textcolor{black}{\scriptsize{}3}} & \textbf{\textcolor{black}{\scriptsize{}0}} & \textbf{\textcolor{black}{\scriptsize{}0}} & \textcolor{black}{\scriptsize{}4} & \textcolor{black}{\scriptsize{}2} & \textbf{\textcolor{black}{\scriptsize{}0}} & \textbf{\textcolor{black}{\scriptsize{}0}} & \textbf{\textcolor{black}{\scriptsize{}98.5\%}} & \textbf{\textcolor{black}{\scriptsize{}60.1\%}} & \textbf{\scriptsize{}0.18}\tabularnewline
\hline 
\multirow{1}{*}{\textcolor{black}{\scriptsize{}YOLOv3+MV-GLMB-OC{*}}} & \textcolor{black}{\scriptsize{}95.0\%} & \textcolor{black}{\scriptsize{}93.5\%} & \textcolor{black}{\scriptsize{}96.5\%} & \textbf{\textcolor{black}{\scriptsize{}3}} & \textbf{\textcolor{black}{\scriptsize{}0}} & \textbf{\textcolor{black}{\scriptsize{}0}} & \textcolor{black}{\scriptsize{}17} & \textcolor{black}{\scriptsize{}4} & \textcolor{black}{\scriptsize{}5} & \textcolor{black}{\scriptsize{}1} & \textcolor{black}{\scriptsize{}93.6\%} & \textcolor{black}{\scriptsize{}58.9\%} & {\scriptsize{}0.20}\tabularnewline
\hline 
\multirow{1}{*}{\textcolor{black}{\scriptsize{}YOLOv3+MS-GLMB}} & \textcolor{black}{\scriptsize{}95.9\%} & \textcolor{black}{\scriptsize{}94.0\%} & \textcolor{black}{\scriptsize{}97.8\%} & \textbf{\textcolor{black}{\scriptsize{}3}} & \textbf{\textcolor{black}{\scriptsize{}0}} & \textbf{\textcolor{black}{\scriptsize{}0}} & \textcolor{black}{\scriptsize{}21} & \textcolor{black}{\scriptsize{}5} & \textcolor{black}{\scriptsize{}4} & \textcolor{black}{\scriptsize{}1} & \textcolor{black}{\scriptsize{}92.6\%} & \textcolor{black}{\scriptsize{}57.0\%} & {\scriptsize{}0.26}\tabularnewline
\hline 
\multirow{1}{*}{\textcolor{black}{\scriptsize{}Faster-RCNN(VGG16)+MV-GLMB-OC}} & \textcolor{black}{\scriptsize{}98.0\%} & \textcolor{black}{\scriptsize{}98.5\%} & \textcolor{black}{\scriptsize{}97.5\%} & \textbf{\textcolor{black}{\scriptsize{}3}} & \textbf{\textcolor{black}{\scriptsize{}0}} & \textbf{\textcolor{black}{\scriptsize{}0}} & \textbf{\textcolor{black}{\scriptsize{}3}} & \textcolor{black}{\scriptsize{}7} & \textcolor{black}{\scriptsize{}2} & \textcolor{black}{\scriptsize{}1} & \textcolor{black}{\scriptsize{}97.0\%} & \textcolor{black}{\scriptsize{}59.3\%} & {\scriptsize{}0.20}\tabularnewline
\hline 
\multirow{1}{*}{\textcolor{black}{\scriptsize{}Faster-RCNN(VGG16)+MV-GLMB-OC{*}}} & \textcolor{black}{\scriptsize{}85.1\%} & \textcolor{black}{\scriptsize{}82.2\%} & \textcolor{black}{\scriptsize{}88.1\%} & \textbf{\textcolor{black}{\scriptsize{}3}} & \textbf{\textcolor{black}{\scriptsize{}0}} & \textbf{\textcolor{black}{\scriptsize{}0}} & \textcolor{black}{\scriptsize{}29} & \textbf{\textcolor{black}{\scriptsize{}0}} & \textcolor{black}{\scriptsize{}6} & \textbf{\textcolor{black}{\scriptsize{}0}} & \textcolor{black}{\scriptsize{}91.3\%} & \textcolor{black}{\scriptsize{}56.2\%} & {\scriptsize{}0.24}\tabularnewline
\hline 
\multirow{1}{*}{\textcolor{black}{\scriptsize{}Faster-RCNN(VGG16)+MS-GLMB}} & \textcolor{black}{\scriptsize{}89.3\%} & \textcolor{black}{\scriptsize{}85.8\%} & \textcolor{black}{\scriptsize{}93.1\%} & \textbf{\textcolor{black}{\scriptsize{}3}} & \textbf{\textcolor{black}{\scriptsize{}0}} & \textbf{\textcolor{black}{\scriptsize{}0}} & \textcolor{black}{\scriptsize{}34} & \textbf{\textcolor{black}{\scriptsize{}0}} & \textcolor{black}{\scriptsize{}12} & \textbf{\textcolor{black}{\scriptsize{}0}} & \textcolor{black}{\scriptsize{}88.6\%} & \textcolor{black}{\scriptsize{}55.3\%} & {\scriptsize{}0.28}\tabularnewline
\hline 
\end{tabular}{\scriptsize\par}

\textcolor{black}{\scriptsize{}}%
\begin{tabular}{|c|c|c|c|c|c|c|c|c|c|c|c|c||c|}
\hline 
\multicolumn{14}{|c|}{\textcolor{black}{\scriptsize{}CMC5 (Jumping and Falling, Maximum/Average
7 people)}}\tabularnewline
\hline 
\textcolor{black}{\scriptsize{}Detector and Tracker}{\scriptsize{}$\!\!\!\!$} & \textcolor{black}{\scriptsize{}IDF1 $\uparrow$} & \textcolor{black}{\scriptsize{}IDP $\uparrow$} & \textcolor{black}{\scriptsize{}IDR $\uparrow$} & {\scriptsize{}$\!\!\!$}\textcolor{black}{\scriptsize{}MT $\uparrow$}{\scriptsize{}$\!\!\!$} & {\scriptsize{}$\!\!\!$}\textcolor{black}{\scriptsize{}PT $\downarrow$}{\scriptsize{}$\!\!\!$} & {\scriptsize{}$\!\!\!$}\textcolor{black}{\scriptsize{}ML $\downarrow$}{\scriptsize{}$\!\!\!$} & \textcolor{black}{\scriptsize{}FP $\downarrow$} & \textcolor{black}{\scriptsize{}FN $\downarrow$} & \textcolor{black}{\scriptsize{}IDs $\downarrow$} & \textcolor{black}{\scriptsize{}FM $\downarrow$} & {\scriptsize{}$\!\!\!$}\textcolor{black}{\scriptsize{}MOTA $\uparrow$}{\scriptsize{}$\!\!\!$} & {\scriptsize{}$\!\!\!$}\textcolor{black}{\scriptsize{}MOTP $\uparrow$}{\scriptsize{}$\!\!\!$} & {\scriptsize{}$\!\!$$\mathrm{OSPA^{(2)}}$}\textcolor{black}{\scriptsize{}$\downarrow$}{\scriptsize{}$\!\!$}\tabularnewline
\hline 
\multirow{1}{*}{\textcolor{black}{\scriptsize{}YOLOv3+MV-GLMB-OC}} & \textbf{\textcolor{black}{\scriptsize{}59.8\%}} & \textbf{\textcolor{black}{\scriptsize{}61.0\%}} & \textbf{\textcolor{black}{\scriptsize{}60.8\%}} & \textbf{\textcolor{black}{\scriptsize{}3}} & \textcolor{black}{\scriptsize{}4} & \textbf{\textcolor{black}{\scriptsize{}0}} & \textbf{\textcolor{black}{\scriptsize{}404}} & \textbf{\textcolor{black}{\scriptsize{}951}} & \textbf{\textcolor{black}{\scriptsize{}67}} & \textbf{\scriptsize{}54} & \textbf{\textcolor{black}{\scriptsize{}60.6\%}} & \textbf{\textcolor{black}{\scriptsize{}45.0\%}} & \textbf{\scriptsize{}$\!\!$0.65$\!\!$}\tabularnewline
\hline 
\multirow{1}{*}{\textcolor{black}{\scriptsize{}YOLOv3+MV-GLMB-OC{*}}} & \textcolor{black}{\scriptsize{}55.9\%} & \textcolor{black}{\scriptsize{}54.9\%} & \textcolor{black}{\scriptsize{}57.1\%} & \textbf{\textcolor{black}{\scriptsize{}3}} & \textcolor{black}{\scriptsize{}3} & \textcolor{black}{\scriptsize{}1} & \textcolor{black}{\scriptsize{}689} & \textcolor{black}{\scriptsize{}1125} & \textcolor{black}{\scriptsize{}80} & \textcolor{black}{\scriptsize{}85} & \textcolor{black}{\scriptsize{}55.3\%} & \textcolor{black}{\scriptsize{}43.4\%} & {\scriptsize{}$\!\!$0.71$\!\!$}\tabularnewline
\hline 
\multirow{1}{*}{\textcolor{black}{\scriptsize{}YOLOv3+MS-GLMB }} & \textcolor{black}{\scriptsize{}49.5\%} & \textcolor{black}{\scriptsize{}50.1\%} & \textcolor{black}{\scriptsize{}45.0\%} & \textbf{\textcolor{black}{\scriptsize{}3}} & \textbf{\textcolor{black}{\scriptsize{}2}} & \textcolor{black}{\scriptsize{}2} & \textcolor{black}{\scriptsize{}715} & \textcolor{black}{\scriptsize{}1750} & \textcolor{black}{\scriptsize{}94} & \textcolor{black}{\scriptsize{}91} & \textcolor{black}{\scriptsize{}49.3\%} & \textcolor{black}{\scriptsize{}42.6\%} & {\scriptsize{}$\!\!$0.78$\!\!$}\tabularnewline
\hline 
\multirow{1}{*}{\textcolor{black}{\scriptsize{}Faster-RCNN(VGG16)+MV-GLMB-OC}} & \textcolor{black}{\scriptsize{}58.1\%} & \textcolor{black}{\scriptsize{}60.8\%} & \textcolor{black}{\scriptsize{}59.4\%} & \textbf{\textcolor{black}{\scriptsize{}3}} & \textcolor{black}{\scriptsize{}4} & \textbf{\textcolor{black}{\scriptsize{}0}} & \textcolor{black}{\scriptsize{}451} & \textcolor{black}{\scriptsize{}1008} & \textcolor{black}{\scriptsize{}72} & {\scriptsize{}57} & \textcolor{black}{\scriptsize{}59.9\%} & \textcolor{black}{\scriptsize{}43.1\%} & {\scriptsize{}$\!\!$0.66$\!\!$}\tabularnewline
\hline 
\multirow{1}{*}{\textcolor{black}{\scriptsize{}Faster-RCNN(VGG16)+MV-GLMB-OC{*}}} & \textcolor{black}{\scriptsize{}55.9\%} & \textcolor{black}{\scriptsize{}53.6\%} & \textcolor{black}{\scriptsize{}51.6\%} & \textbf{\textcolor{black}{\scriptsize{}3}} & \textcolor{black}{\scriptsize{}3} & \textcolor{black}{\scriptsize{}1} & \textcolor{black}{\scriptsize{}569} & \textcolor{black}{\scriptsize{}1519} & \textcolor{black}{\scriptsize{}81} & \textcolor{black}{\scriptsize{}88} & \textcolor{black}{\scriptsize{}51.4\%} & \textcolor{black}{\scriptsize{}42.7\%} & {\scriptsize{}$\!\!$0.75$\!\!$}\tabularnewline
\hline 
\multirow{1}{*}{\textcolor{black}{\scriptsize{}Faster-RCNN(VGG16)+MS-GLMB}} & \textcolor{black}{\scriptsize{}48.8\%} & \textcolor{black}{\scriptsize{}45.3\%} & \textcolor{black}{\scriptsize{}41.7\%} & \textbf{\textcolor{black}{\scriptsize{}3}} & \textcolor{black}{\scriptsize{}3} & \textcolor{black}{\scriptsize{}1} & \textcolor{black}{\scriptsize{}734} & \textcolor{black}{\scriptsize{}1493} & \textcolor{black}{\scriptsize{}96} & {\scriptsize{}98} & \textcolor{black}{\scriptsize{}43.3\%} & \textcolor{black}{\scriptsize{}43.9\%} & {\scriptsize{}$\!\!$0.81$\!\!$}\tabularnewline
\hline 
\end{tabular}{\scriptsize\par}

\textcolor{black}{\smallskip{}
}

\raggedright

\emph{\footnotesize{}CLEAR MOT scores and OSPA}\textsuperscript{\emph{\footnotesize{}(2)}}\emph{\footnotesize{}
distance are calculated with a 3D GIoU base-distance for estimates
of 3D centroid with extent ($\uparrow$ means higher is better while
$\downarrow$ means lower is better). Two different detectors are
considered - Faster-RCNN(VGG16) (monocular) and YOLOv3 (monocular).
Two types of trackers are considered - MV-GLMB-OC (multi-view with
occlusion model) and MS-GLMB (multi-sensor without occlusion model).
The asterisk ({*}) indicates the multi-camera reconfiguration experiment.}{\footnotesize\par}

\vspace*{-0.4cm}

\label{MOT_GIoU_CMC4_CMC5-1}
\end{table*}

\noindent \vspace*{-1.2cm}

\subsubsection{Model Parameters}

Each state is augmented $\boldsymbol{x}$ with a discrete mode or
class $m\negmedspace\in\negmedspace\{0,1\}$, where $m\negmedspace=\negmedspace0$
corresponds to a standing state and $m\negmedspace=\negmedspace1$
corresponds to a fallen state. We consider the single-object state
as $(\boldsymbol{x},m)$, with single-object density $p^{(\xi)}(\boldsymbol{x},m)=p^{(\xi)}(\boldsymbol{x}|m)\mu{}^{(\xi)}(m)$.
The following single-object transition density and observation likelihood
are used
\[
\boldsymbol{f}\!_{S,+}(\boldsymbol{x}_{+}m_{+}\!|\boldsymbol{x},m)\!\!=\!\!f_{S,+}^{(m_{+})}\!(x_{+}\!|x,\ell,m)\delta_{\ell}[\ell_{+}]\mu_{+}\!(m_{+}\!|m),\!\!\!
\]
\begin{multline*}
\!\!\!g^{(c)}(z^{(c)}|\boldsymbol{x},m)\propto g_{e}^{(c)}\!(z_{e}^{(c)}\!|m)\times\\
\negmedspace\negmedspace\mathcal{N}\!\left(\!z^{(c)};\Phi^{(c)}(\boldsymbol{x})\!+\!\!\left[\!\!\!\!\begin{array}{c}
0_{2\times1}\\
-\upsilon_{e}^{(c,m)}\!\!/2
\end{array}\!\!\!\right]\!\!,\textrm{diag\!\!}\left(\!\left[\!\!\!\begin{array}{c}
\upsilon_{p}^{(c)}\\
\upsilon_{e}^{(c,m)}
\end{array}\!\!\!\!\right]\!\right)\!\!\right)\!\!.\negmedspace\negmedspace
\end{multline*}
The mode transition probabilities are $\mu_{+}(0|0)=0.6$, $\mu_{+}(1|0)=0.4$,
$\mu_{+}(0|1)=0.6$ and $\mu_{+}(1|1)=0.4$. 

For a standing object, i.e. $m\negmedspace=\negmedspace0$, we have
$\upsilon_{e}^{(c,0)}\negmedspace=\negmedspace\upsilon_{e}^{(c)}\negmedspace=\negmedspace[0.01,0.0025]^{T}$
in the above observation likelihood. Further, standing objects typically
have a bounding box size ratio (y-axis/x-axis) greater than one, thus
the mode dependent likelihood component is chosen as $g_{e}^{(c)}(z_{e}^{(c)}|0)=e^{\rho\left((\nicefrac{[0,1]z_{e}^{(c)}}{[1,0]z_{e}^{(c)})-1}\right)}$
for all cameras, where $\rho=2$ is a control parameter. The transition
density to another standing state $f_{S,+}^{(0)}(x_{+}|x,\ell,0)$,
is the same as per the previous subsection. 

For a fallen object, i.e. $m\negmedspace=\negmedspace1$, we have
$\upsilon_{e}^{(c,1)}\negmedspace=\negmedspace[0.0025,0.01]^{T}$
in the above observation likelihood, and the mode dependent likelihood
component is chosen as $g_{e}^{(c)}(z_{e}^{(c)}|1)=e^{-\rho\left((\nicefrac{[0,1]z_{e}^{(c)}}{[1,0]z_{e}^{(c)})-1}\right)}$
for all cameras because fallen objects typically have a bounding box
size ratio (y-axis/x-axis) less than one. The transition density to
another fallen state $f_{S,+}^{(1)}(x_{+}|x,\ell,1)$ is the same
as that for standing-to-standing except for the large variance $\upsilon^{(s)}=[0.15,0.15,0.04]{}^{T}$
to capture all possible orientations during the fall. 

For a state transition involving a mode switch i.e. standing-to-fallen
or fallen-to-standing, the transition density $f_{+}^{(1)}(x_{+}|x,\ell,0)$
or $f_{+}^{(0)}(x_{+}|x,\ell,1)$ takes the form (\ref{eq:transition_density}),
with position noise and extent (in logarithm) noise parameterized
by: 
\begin{align*}
\upsilon^{(p)}= & [0.0049,0.0049,0.0049]^{T},\\
\upsilon^{(s)}= & [0.01,0.01,0.01]^{T}.
\end{align*}
Notice that the position noise is increased in the case of a mode
switch compared to the case of no switching, in order to capture the
abrupt change in the size of the object during mode switching. 

The birth density is an LMB with parameters $P_{B,+}(\ell)=0.001$
and
\begin{align*}
f_{B,+}(x,\ell,0)= & 0.9\mathcal{N}(x;\mu_{B,+,0},\Sigma_{B,+,0}),\\
f_{B,+}^{(\ell)}(x,\ell,1)= & 0.1\mathcal{N}(x;\mu_{B,+,1},\Sigma_{B,+,1}),\\
\mu_{B,+,0}= & [2.03\;0\;0.71\;0\;0.825\;0\;-\negmedspace1.2\;-\negmedspace1.2\;-\negmedspace0.18]^{T},\\
\mu_{B,+,1}= & [2.03\;0\;0.71\;0\;0.413\;0\;-\negmedspace0.18\;-\negmedspace0.18\;-\negmedspace1.2]^{T},\\
\Sigma_{B,+,0}= & \Sigma_{B,+,1}=0.1^{2}\mathrm{I}_{9}.
\end{align*}

\subsubsection{Effectiveness of Occlusion Model \label{subsec:Effectiveness-of-Occlusion}}

Tables \ref{MOT_Euclidean_CMC4_CMC5} and \ref{MOT_GIoU_CMC4_CMC5-1}
show the CLEAR MOT and OSPA\textsuperscript{(2)} benchmarks for MV-GLMB-OC
and MS-GLMB on both detectors YOLOv3 and Faster-RCNN(VGG16). The CLEAR
evaluations for the monocular detections are given in Appendix \ref{subsec:Monocular-Detector-Results}. 

For CMC4 which has a maximum of 3 people, both MV-GLMB-OC and MS-GLMB
on either detectors achieved high CLEAR MOT scores in MOTA/MOTP, and
low OSPA\textsuperscript{(2)} errors. The incidence of FPs and FNs
is caused by track initiation/termination mismatches with the ground
truths. Nonetheless, we observe that on MOTA/MOTP and OSPA\textsuperscript{(2)},
MV-GLMB-OC outperforms MS-GLMB.

\begin{figure}
\includegraphics[width=4.4cm,height=4.5cm]{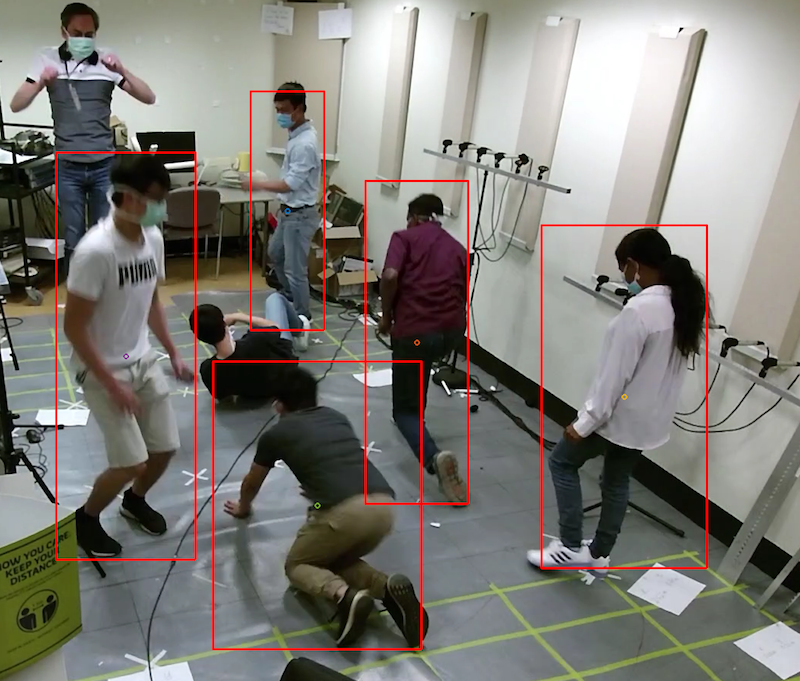}\includegraphics[width=4.4cm,height=4.5cm]{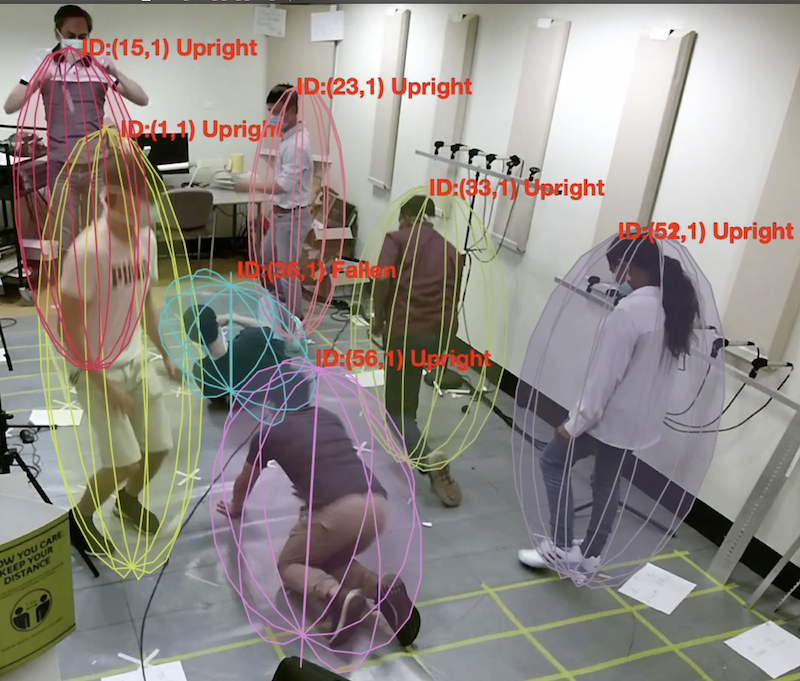}

\smallskip{}

\caption{CMC5 Camera 1: YOLOv3 detections (left) and MV-GLMB-OC estimates (right).}
\label{CMC5_figures}
\end{figure}

For CMC5 which has a maximum of 7 people, both MV-GLMB-OC and MS-GLMB
on either detectors were still capable of producing reasonable MOTA/MOTP
scores and OSPA\textsuperscript{(2)} errors. Fig. \ref{CMC5_figures}
shows a snapshot of detections and estimates on a single view. However,
due to poor detections and more occlusions in CMC5, we observe many
IDs and FNs. Again on MOTA/MOTP and OSPA\textsuperscript{(2)}, MV-GLMB-OC
outperforms MS-GLMB.

\noindent 

\noindent \vspace*{-0.7cm}

\noindent 
\begin{figure*}
\includegraphics[width=18cm,height=5cm]{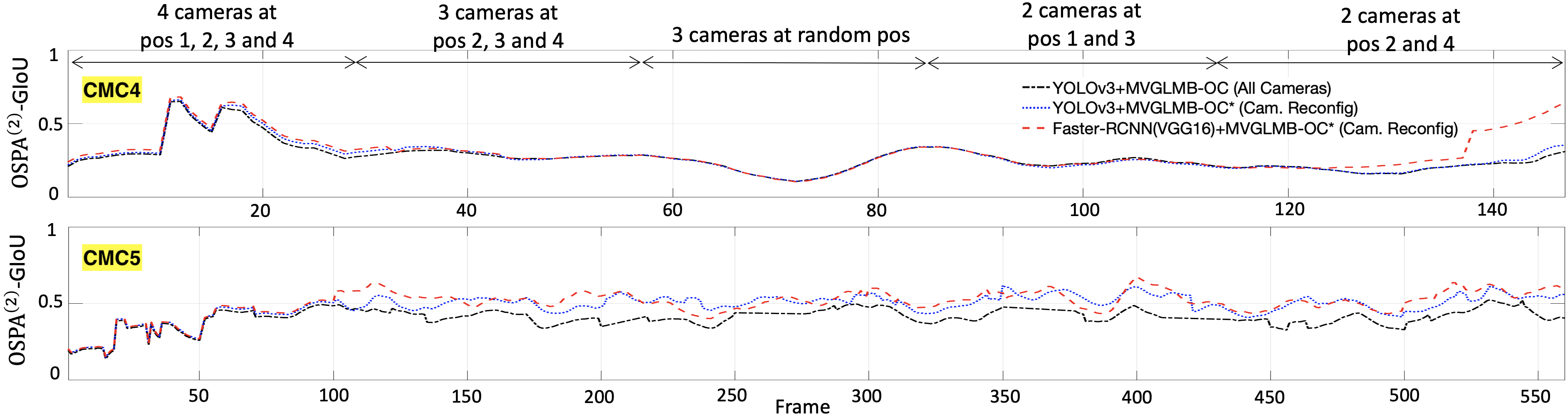}

\caption{Multi-Camera Reconfiguration Experiment: OSPA\protect\textsuperscript{(2)}
plots with 3D GIoU base-distance for estimates of 3D centroid with
extent. Three trackers are considered: YOLOv3+MV-GLMB-OC{*} (multi-camera
reconfiguration) and Faster-RCNN+MV-GLMB-OC{*} (multi-camera reconfiguration)
and with YOLOv3+MV-GLMB-OC (all cameras operational). }

\label{OSPA_GIoU_CMC4_CMC5}
\end{figure*}

\noindent \vspace*{-0.9cm}

\subsubsection{Multi-Camera Reconfiguration}

The multi-camera reconfiguration experiment described in Section \ref{subsec:Multi-Camera-Reconfiguration}
is repeated for the multi-modal datasets CMC4 and CMC5. The results
for the multi-camera reconfiguration are denoted with asterisks in
Tables \ref{MOT_Euclidean_CMC4_CMC5} and \ref{MOT_GIoU_CMC4_CMC5-1}.
The plot for $\mathrm{OSPA}^{(2)}$ with 3D GIoU base-distance over
a sliding window with time is given in Fig. \ref{OSPA_GIoU_CMC4_CMC5}.
While similar observations can be made from the experiments without
jumping and falling (CMC1-CMC3), the results for CMC4-CMC5 exhibit
different behavior for people in the fallen state. The estimated extent
is warped out of its ordinary shape when the person is on the ground,
and more data is required to infer the corresponding state of the
fallen person. In CMC4-CMC5, the effect of occlusions or misdetections
is exacerbated by having fewer cameras when the person is on the ground,
which would likely lead to track termination or switching. Nonetheless,
the results confirm that the JMS variant of the MV-GLMB-OC algorithm
can automatically accommodate multi-camera reconfiguration. 

\bigskip{}

\noindent 
\begin{table}[H]
\captionsetup{justification=centering} 

\caption{\small{}MV-GLMB-OC Runtime on WILDTRACKS and CMC}

\centering

\begin{tabular}{|c|c|c|c|}
\hline 
Dataset (Cams) & Frames & No. Obj (avg) & Exec. Time (s/frame)\tabularnewline
\hline 
W.T. (7) & 400 & 20 & 18.0\tabularnewline
\hline 
CMC1(4) & 261 & 3 & 0.1\tabularnewline
\hline 
CMC2 (4) & 263 & 10 & 3.2\tabularnewline
\hline 
CMC3 (4) & 263 & 15 & 7.9\tabularnewline
\hline 
CMC4 (4) & 147 & 3 & 0.4\tabularnewline
\hline 
CMC5 (4) & 560 & 7 & 5.5\tabularnewline
\hline 
\end{tabular}

\label{Experimental runtime}
\end{table}

\subsection{Runtimes}

The runtimes for the MV-GLMB-OC filter on the WILDTRACKS and CMC datasets
are summarized in Table \ref{Experimental runtime}. The current implementation
is via unoptimized MATLAB code. The reported runtimes appear to be
consistent with the computational complexity of the MV-GLMB-OC algorithm:
quadratic in the number of objects and linear in the sum of the number
of detections across all cameras. 

\section{Conclusions \label{sec:Conclusion}}

By developing a tractable 3D occlusion model, we have derived an online
Bayesian multi-view multi-object filtering algorithm that only requires
monocular detector training, independent of the multi-camera configurations.
This enables the multi-camera system to operate uninterrupted in the
event of extension/reconfiguration (including camera failures), obviating
the need for multi-view retraining. Moreover, it addresses the multi-camera
data association problem in a way that is scalable in the total number
of detections. Experiments on existing 3D multi-camera datasets have
demonstrate\textcolor{black}{d similar performance to the state-of-the-art
batch method. }We also demonstrated the ability of the proposed algorithm
to track in densely populated scenarios with high occlusions, and
with people jumping/falling in the 3D world frame. %
\begin{comment}
The propose framework can account for interactions between objects
by using the novel multi-object motion model proposed in \cite{gostar2019interactive}.
The framework can also accommodate more complex applications that
require re-establishing the identities of reappearing objects, instead
of simply terminating tracks once they exit the field of view. This
can be accomplished by incorporate appearance into the state vector,
which could be estimated on-line (e.g. via deep learning). When an
object is detected, the MOT filter hypothesizes whether it is a new
or reappearing object depending on how well its appearance feature
match the appearance features of the undetected current objects. Such
an extension incurs additional computational cost and further work
is needed for real-time applications. 
\end{comment}

\bibliographystyle{ieeetr}
\bibliography{References}
\vspace{-1.1cm}

\begin{IEEEbiography}[{{\includegraphics[clip,width=1in,height=1.25in,keepaspectratio]{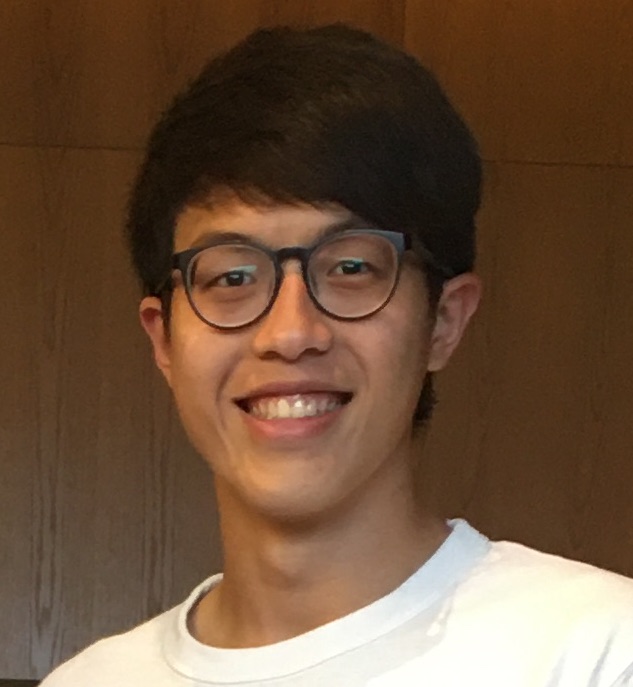}}}]{Jonah Ong}
 received the B.E. degree in electrical and power engineering with
first-class honors from Curtin University, Perth, Western Australia,
in 2018. He is currently working towards the Ph.D. degree in electrical
engineering at Curtin University. His research interests include statistical
signal processing, Bayesian filtering and estimation, random sets,
and multi-target tracking. 
\end{IEEEbiography}

\vspace{-1.1cm}
\begin{IEEEbiography}[{{\includegraphics[clip,width=1in,height=1.25in,keepaspectratio]{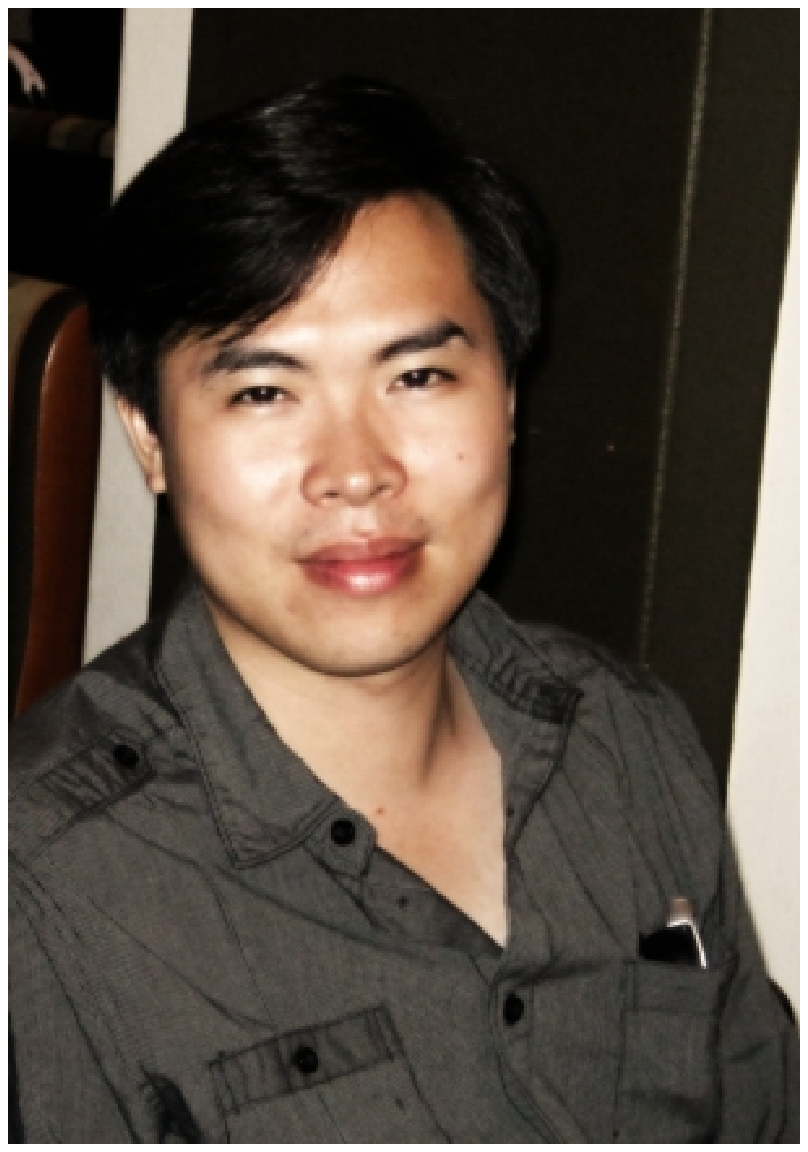}}}]{Ba-Tuong Vo}
 received the B.Sc. degree in applied mathematics and B.E. degree
in electrical and electronic engineering (with first-class honors)
in 2004 and the Ph.D. degree in engineering (with Distinction) in
2008, all from the University of Western Australia. He is currently
a Professor of Signal Processing at Curtin University and has primary
research interests in random sets, filtering and estimation, multiple
object systems. 
\end{IEEEbiography}

\vspace{-1.1cm}
\begin{IEEEbiography}[{{\includegraphics[clip,width=1in,height=1.25in,keepaspectratio]{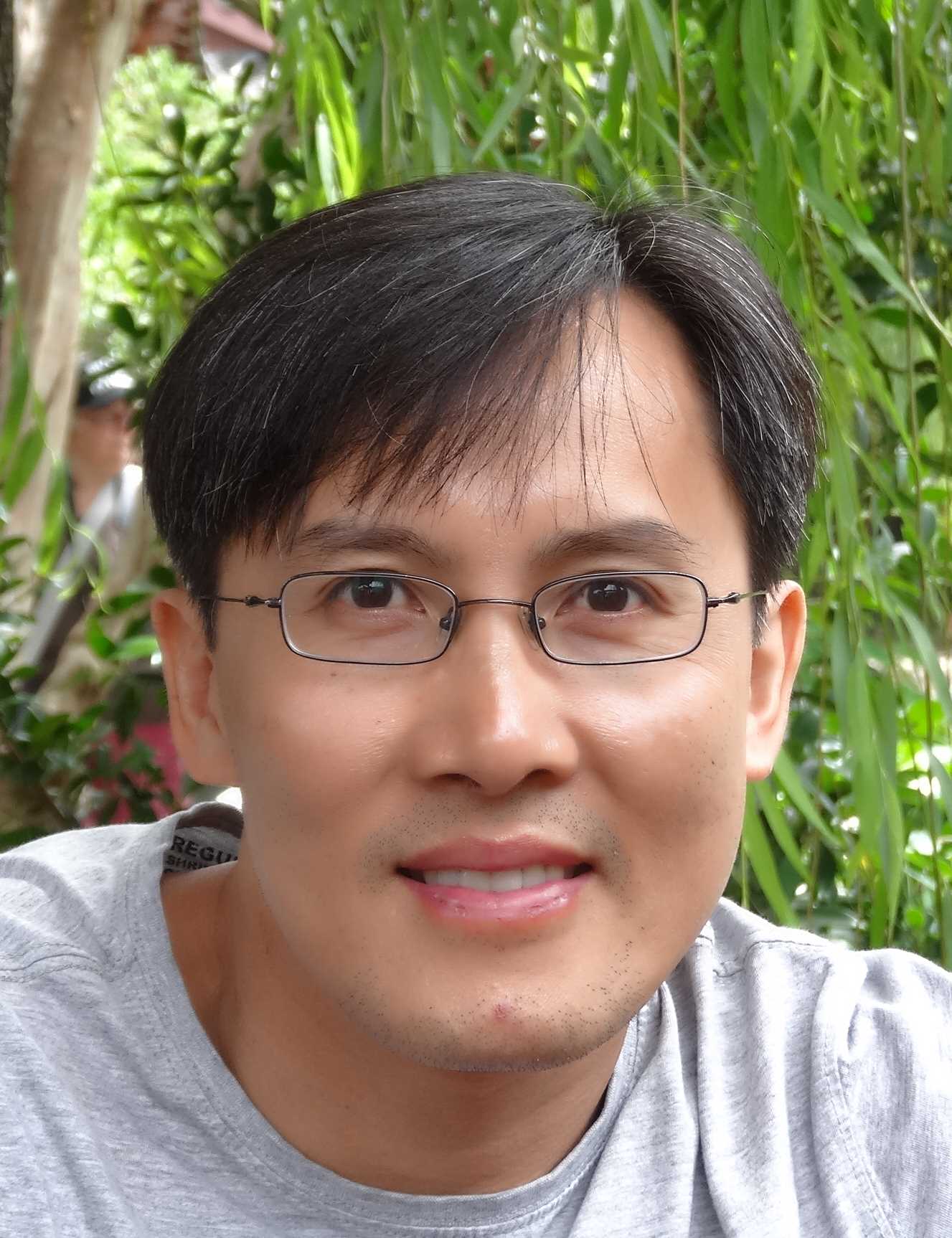}}}]{Ba-Ngu Vo}
 received his Bachelor degrees in Mathematics and Electrical Engineering
with first class honors in 1994, and PhD in 1997. Currently he is
Professor of Signals and Systems at Curtin University. Vo is a Fellow
of the IEEE, and is best known as a pioneer in the stochastic geometric
approach to multi-object system. His research interests are signal
processing, systems theory and stochastic geometry.
\end{IEEEbiography}

\vspace{-1.1cm}
\begin{IEEEbiography}[{{\includegraphics[clip,width=1in,height=1.25in,keepaspectratio]{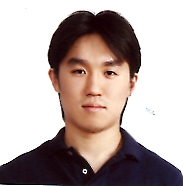}}}]{Du Yong Kim}
 received the B.E. degree in electrical and electronics engineering
from Ajou University, Korea, in 2005. He received the M.S. and Ph.D.
degrees from the Gwangju Institute of Science and Technology, Korea,
in 2006 and 2011, respectively. As a Postdoctoral Researcher, he worked
on statistical signal processing and image processing at the Gwangju
Institute of Science and Technology (2011---2012), and the University
of Western Australia (2012---2014), Curtin University (2014-2018).
He is currently working as a Vice-Chancellor's Research Fellow at
School of Engineering, RMIT University. His main research interests
include Bayesian filtering theory and its applications to machine
learning, computer vision, sensor networks, and automatic control.
\end{IEEEbiography}

\begin{IEEEbiography}[{{\includegraphics[clip,width=1in,height=1.25in,keepaspectratio]{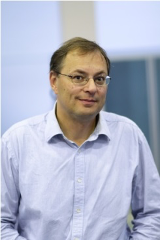}}}]{Sven Nordholm}
 received his PhD in Signal Processing in 1992, Licentiate of engineering
1989 and MscEE (Civilingenjör) 1983 all from Lund University, Sweden.
Since 1999, Nordholm is Professor, Signal Processing, School of Electrical
and Computer Engineering, Curtin University. He is a co-founder of
two start-up companies; Sensear, providing voice communication in
extreme noise conditions and Nuheara a hearables company. He was a
lead editor for a special issue on assistive listing techniques in
IEEE Signal Processing Magazine and several other EURASIP special
issues. He is a former Associate editor for Eurasip Advances in Signal
Processing and Journal of Franklin Institute and at the current time
an Associate Editor IEEE/ACM Transaction on Audio, Speech and Language
Processing. His primary research has encompassed the fields of Speech
Enhancement, Adaptive and Optimum Microphone Arrays, Audio Signal
Processing and WirelessCommunication. He has written more than 200
papers in refereed journals and conference proceedings. He frequently
contributes to book chapters and encyclopedia articles. He is holding
seven patents in the area of speech enhancement and microphone arrays. 
\end{IEEEbiography}

\clearpage{}

\section{Appendix\label{sec:Appendix}}

\pagenumbering{gobble}

\subsection{MS-GLMB recursion \label{subsec:MS-GLMB-recursion}}

Under the standard multi-object model with no occlusions, i.e. $P_{D}^{\left(c\right)}\left(\boldsymbol{x};\boldsymbol{X}\!-\!\{\boldsymbol{x}\}\right)=P_{D}^{\left(c\right)}\left(\boldsymbol{x}\right)$,
hence $\psi_{\boldsymbol{X}\!-\{\boldsymbol{x}\}}^{(\gamma)}(\boldsymbol{x})$
does not depend on $\boldsymbol{X}\!-\!\{\boldsymbol{x}\}$, and as
a result we write $\psi_{\boldsymbol{X}\!-\{\boldsymbol{x}\}}^{(\gamma)}(\boldsymbol{x})=\psi^{(\gamma)}(\boldsymbol{x})$.
In this case, the MS-GLMB recursion\textit{ $\Omega$} taking the
parameter-set 
\[
\boldsymbol{\pi}\triangleq\left\{ \left(w^{\left(I,\xi\right)},p^{\left(\xi\right)}\right):\left(I,\xi\right)\in\mathcal{F}(\mathbb{L})\times\Xi\right\} .
\]
 to the parameter-set
\[
\boldsymbol{\pi}_{+}=\left\{ \left(w_{+}^{\left(I_{+},\xi_{+}\right)},p_{+}^{\left(\xi_{+}\right)}\right)\!:\left(I_{+},\xi_{+}\right)\in\mathcal{F}(\mathbb{L_{+}})\times\Xi_{+}\right\} 
\]
is given by

{\footnotesize{}\vspace*{-0.4cm}}
\begin{equation}
I_{+}=\mathbb{B}_{+}\uplus I,\;\xi_{+}=(\xi,\gamma_{+})\label{e:GLMB_JPU_First-1-1-1-1}
\end{equation}
\begin{equation}
w_{+}^{\left(I_{+},\xi_{+}\right)}=1_{\mathcal{F}(\mathbb{B}_{+}\uplus I)}(\mathcal{L}(\gamma_{+}))w^{(I,\xi)}\left[\omega^{(\xi,\gamma_{+})}\right]^{\mathbb{B}_{+}\uplus I\!}\label{eq:Upda_weight}
\end{equation}

{\footnotesize{}\vspace*{-0.8cm}}
\begin{multline}
\!\!\!\!\!\!\!p_{+}^{\left(\xi_{+}\right)}(x_{+},\ell)\propto\\
\begin{cases}
\langle\Lambda_{S}^{(\gamma_{+})}(x_{+}|\cdot,\ell),p^{(\xi)}(\cdot,\ell)\rangle, & \!\!\ell\in\mathcal{L}(\gamma_{+})-\mathbb{B_{+}}\\
\Lambda_{B}^{(\gamma_{+})}(x_{+},\ell), & \!\!\ell\in\mathcal{L}(\gamma_{+})\cap\mathbb{B_{+}}
\end{cases}\label{eq:_plus_p-1-1}
\end{multline}

\begin{equation}
\!\!\!\!\!\!\!\!\!\!\!\!\!\!\!\omega^{(\xi,\gamma_{+})}(\ell)=\begin{cases}
1-\bar{P}_{S}^{\left(\xi\right)}\left(\ell\right), & \!\!\ell\in\overline{\mathcal{L}(\gamma_{+})}-\mathbb{B_{+}}\\
\overline{\Lambda}_{S}^{(\xi,\gamma_{+})}(\ell), & \!\!\ell\in\mathcal{L}(\gamma_{+})-\mathbb{B_{+}}\\
1-P_{B,+}(\ell), & \!\!\ell\in\overline{\mathcal{L}(\gamma_{+})}\cap\mathbb{B_{+}}\\
\overline{\Lambda}_{B}^{(\gamma_{+})}(\ell), & \!\!\ell\in\mathcal{L}(\gamma_{+})\cap\mathbb{B}_{+}
\end{cases},
\end{equation}

{\footnotesize{}\vspace*{-0.4cm}}{\footnotesize\par}

\begin{align}
\bar{P}_{S}^{\left(\xi\right)}\left(\ell\right)= & \langle P_{S}(\cdot,\ell),p^{(\xi)}(\cdot,\ell)\rangle,\\
\!\!\!\Lambda_{B}^{(\gamma_{+})}(x_{+},\ell)= & \psi^{(\gamma_{+})}\!\!\!\left(x_{+},\ell\right)f_{B,+}(x_{+},\ell)P_{B,+}(\ell),\\
\!\!\!\!\Lambda_{S}^{(\gamma_{+})}(x_{+}|y,\ell)= & \psi^{(\gamma_{+})}\!\!\!\left(x_{+},\ell\right)f_{S,+}(x_{+}|y,\ell)P_{S}(y,\ell),\\
\overline{\Lambda}_{B}^{(\gamma_{+})}(\ell)= & \int\Lambda_{B}^{(\gamma_{+})}(x,\ell)dx,\\
\overline{\Lambda}_{S}^{(\xi,\gamma_{+})}(\ell)= & \int\langle\Lambda_{S}^{(\gamma_{+})}(x|\cdot,\ell),p^{(\xi)}(\cdot,\ell)\rangle dx.
\end{align}
{\footnotesize{}}%
{\footnotesize\par}

\noindent {\footnotesize{}\vspace{-0.1cm}
}{\footnotesize\par}

\subsection{OSPA/OSPA\protect\textsuperscript{(2)} Metrics \label{subsec:OSPA2}}

Consider a space $\mathcal{\mathbb{W}}$ with $\underline{d}:\mathcal{\mathcal{\mathbb{W}}\times}\mathcal{\mathbb{W}}\rightarrow[0;\infty)$
as the\emph{ base-distance} between the elements of $\mathcal{\mathbb{W}}$.
Let $\underline{d}^{(c)}(x,y)=\min\left(c,\underline{d}\left(x,y\right)\right)$,
and $\Pi_{n}$ be the set of permutations of $\left\{ 1,2,...,n\right\} $.
The Optimal Sub-Pattern Assignment (OSPA) distance of order $p\geq1$,
and cut-off $c>0$, between two finite sets of points $X=\{x_{1},...,x_{m}\}$
and $Y=\{y_{1},...,y_{n}\}$ of $\mathcal{\mathbb{W}}$ is defined
by \cite{schuhmacher2008consistent}
\begin{align}
\!\!\!\!d_{\mathtt{O}}^{(p,c)}\!(X\!,\!Y)\!\!=\!\! & \left(\!\frac{1}{n}\!\left(\!\min_{\pi\in\Pi_{n}}\!\sum_{i=1}^{m}\underline{d}^{(c)}\!\!\left(x_{i},y_{\pi(i)}\right)^{p}\!+\!c^{p}\!\left(n-m\right)\!\!\right)\!\!\right)^{\frac{1}{p}}\!\!\!\!\!,\label{eq:OSPA-dist-c}
\end{align}
if $n\geq m>0$, and $d_{\mathtt{O}}^{(p,c)}(X,Y)=d_{\mathtt{O}}^{(p,c)}(Y,X)$
if $m>n>0$. In addition, $d_{\mathtt{O}}^{(p,c)}(X,Y)=c$ if one
of the set is empty, and $d_{\mathtt{O}}^{(p,c)}(\emptyset,\emptyset)=0$.
The integer $p$ plays the same role as the order of the $\ell_{p}$-distance
for vectors. The cut-off parameter $c$ provides a weighting between
cardinality and location errors. A large $c$ emphasizes cardinality
error while a small $c$ emphasizes location error. However, a small
$c$ also decreases the sensitivity to the separation between the
points due to the saturation of $\underline{d}^{(c)}$ at $c$. 

The OSPA\textsuperscript{(2)} distance between two sets of tracks
is the OSPA distance with the following base-distance between two
tracks $f$ and $g$ \cite{beard2020solution}:
\begin{align}
\underline{d}^{\left(c\right)}\left(f,g\right)= & \sum\limits _{t\in\mathcal{D}_{f}\cup\mathcal{D}_{g}}\!\frac{d_{\mathtt{O}}^{\left(c\right)}\left(\left\{ f\left(t\right)\right\} ,\left\{ g\left(t\right)\right\} \right)}{\left|\mathcal{D}_{f}\cup\mathcal{D}_{g}\right|},\label{eq:OSPA-dist-c-1}
\end{align}
if $\mathcal{D}_{f}\cup\mathcal{D}_{g}\neq\emptyset$, and $\underline{d}^{\left(c\right)}\left(f,g\right)=0$
if $\mathcal{D}_{f}\cup\mathcal{D}_{g}=\emptyset$, where $\mathcal{D}_{f}\cup\mathcal{D}_{g}$
denotes the set of instants when at least one of the tracks exists,
and $d_{\mathtt{O}}^{\left(c\right)}\left(\left\{ f\left(t\right)\right\} ,\left\{ g\left(t\right)\right\} \right)$
denotes the OSPA distance between the two sets containing the states
of the two tracks at time $t$ (the set $\left\{ f\left(t\right)\right\} $
(or $\left\{ g\left(t\right)\right\} $) would be empty if the track
$f$ (or $g$) does not exist at time $t$). Note that the order parameter
$p$ of the OSPA distance in (\ref{eq:OSPA-dist-c-1}) is redundant
because only sets of at most one element are considered.

\subsection{Intersection-over-Union (IoU) and Generalized IoU (GIoU) Metrics\label{subsec:Intersection-over-Union-(IoU)-an}}

For bounding boxes $x,y$, the IoU similarity index is given by $IoU(x,y)={\left|x\cap y\right|}/{\left|x\cup y\right|}\in[0;1]$,
where $\left|\cdot\right|$ denotes hyper-volume%BN: I took these out as suggested by Hamid. However, I feel that this is a very abrupt jump
%Note that the IoU similarity between non-overlapping shapes is always zero regardless of their shapes or proximity, which is undesirable for performance evaluation. Generalized IoU (GIoU) introduces an additional term to provide a meaningful measure of similarity for non-overlapping shapes, which takes into account their shapes and proximity. 
, while the Generalized IoU index is given by $GIoU(x,y)=IoU(x,y)-{\left|C(x\cup y)\setminus\left(x\cup y\right)\right|}/{\left|C(x\cup y)\right|}$,
where $C(x\cup y)$ is the convex hull of $x\cup y$ \cite{Rezatofighi_2018_CVPR}.
The metric forms of IoU and GIoU, respectively are $\underline{d}_{IoU}(x,y)=1-IoU(x,y)$
and $\underline{d}_{GIoU}(x,y)=\frac{1-GIoU(x,y)}{2}$, both of which
are bounded by one \cite{Rezatofighi_2018_CVPR}.

\subsection{Monocular Detector Results \label{subsec:Monocular-Detector-Results}}

Table \ref{CMC_CLEARMOT_Det} shows the CLEAR evaluation for detections
on the CMC dataset, which is referenced from Section \ref{subsec:Discussion-on-Ablation}
and Section \ref{subsec:Effectiveness-of-Occlusion}. Table \ref{WILDTRACKS_Det_Results}
shows the CLEAR evaluation for detections on WILDTRACKS dataset, which
is referenced from Section \ref{subsec:WildtrackS_Discussion}. 

\noindent 
\begin{table}[H]
\captionsetup{justification=centering} 

\caption{\small{}CLEAR Evaluation for Detection Results on WILDTRACKS Dataset}

\centering

{\scriptsize{}}%
\begin{tabular}{|c|c|c|c|c|c|}
\hline 
 & {\scriptsize{}Detector} & {\scriptsize{}MODA $\uparrow$} & {\scriptsize{}MODP $\uparrow$} & {\scriptsize{}Precision $\uparrow$} & {\scriptsize{}Recall $\uparrow$}\tabularnewline
\hline 
\multirow{2}{*}{{\scriptsize{}C1}} & {\scriptsize{}YOLOv3} & {\scriptsize{}12.2\%} & {\scriptsize{}70.1\%} & {\scriptsize{}0.55} & {\scriptsize{}0.62}\tabularnewline
 & {\scriptsize{}F-RCNN(VGG16)} & {\scriptsize{}-17.1\%} & {\scriptsize{}69.6\%} & {\scriptsize{}0.44} & {\scriptsize{}0.62}\tabularnewline
\hline 
\multirow{2}{*}{{\scriptsize{}C2}} & {\scriptsize{}YOLOv3} & {\scriptsize{}31.7\%} & {\scriptsize{}68.5\%} & {\scriptsize{}0.68} & {\scriptsize{}0.58}\tabularnewline
 & {\scriptsize{}F-RCNN(VGG16)} & {\scriptsize{}28.4\%} & {\scriptsize{}68.3\%} & {\scriptsize{}0.67} & {\scriptsize{}0.57}\tabularnewline
\hline 
\multirow{2}{*}{{\scriptsize{}C3}} & {\scriptsize{}YOLOv3} & {\scriptsize{}-24.4\%} & {\scriptsize{}69.2\%} & {\scriptsize{}0.42} & {\scriptsize{}0.68}\tabularnewline
 & {\scriptsize{}F-RCNN(VGG16)} & {\scriptsize{}-34.6\%} & {\scriptsize{}69.0\%} & {\scriptsize{}0.40} & {\scriptsize{}0.69}\tabularnewline
\hline 
\multirow{2}{*}{{\scriptsize{}C4}} & {\scriptsize{}YOLOv3} & {\scriptsize{}-272.4\%} & {\scriptsize{}71.1\%} & {\scriptsize{}0.14} & {\scriptsize{}0.57}\tabularnewline
 & {\scriptsize{}F-RCNN(VGG16)} & {\scriptsize{}-300.0\%} & {\scriptsize{}70.1\%} & {\scriptsize{}0.14} & {\scriptsize{}0.57}\tabularnewline
\hline 
\multirow{2}{*}{{\scriptsize{}C5}} & {\scriptsize{}YOLOv3} & {\scriptsize{}-94.4\%} & {\scriptsize{}70.0\%} & {\scriptsize{}0.29} & {\scriptsize{}0.69}\tabularnewline
 & {\scriptsize{}F-RCNN(VGG16)} & {\scriptsize{}-113.0\%} & {\scriptsize{}67.8\%} & {\scriptsize{}0.27} & {\scriptsize{}0.71}\tabularnewline
\hline 
\multirow{2}{*}{{\scriptsize{}C6}} & {\scriptsize{}YOLOv3} & {\scriptsize{}-12.6\%} & {\scriptsize{}63.4\%} & {\scriptsize{}0.44} & {\scriptsize{}0.50}\tabularnewline
 & {\scriptsize{}F-RCNN(VGG16)} & {\scriptsize{}-30.5\%} & {\scriptsize{}65.4\%} & {\scriptsize{}0.39} & {\scriptsize{}0.53}\tabularnewline
\hline 
\multirow{2}{*}{{\scriptsize{}C7}} & {\scriptsize{}YOLOv3} & {\scriptsize{}-79.2\%} & {\scriptsize{}70.1\%} & {\scriptsize{}0.33} & {\scriptsize{}0.77}\tabularnewline
 & {\scriptsize{}F-RCNN(VGG16)} & {\scriptsize{}-100.3\%} & {\scriptsize{}69.3\%} & {\scriptsize{}0.31} & {\scriptsize{}0.77}\tabularnewline
\hline 
{\scriptsize{}All} & {\scriptsize{}Deep-Occlusion} & {\scriptsize{}74.1\%} & {\scriptsize{}53.8\%} & {\scriptsize{}0.95} & {\scriptsize{}0.80}\tabularnewline
\hline 
\end{tabular}{\scriptsize\par}

\label{WILDTRACKS_Det_Results}
\end{table}

\noindent {\footnotesize{}\vspace{0.3cm}
}{\footnotesize\par}

\noindent 
\begin{table}[H]
\captionsetup{justification=centering} 

\caption{\small{}CLEAR Evaluation for Detection Results on \\ CMC1 to CMC5}

\centering

{\scriptsize{}}%
\begin{tabular}{|c|c|c|c|c|c|}
\hline 
{\scriptsize{}CMC1} & {\scriptsize{}Detector} & {\scriptsize{}MODA $\uparrow$} & {\scriptsize{}MODP $\uparrow$} & {\scriptsize{}Prcn $\uparrow$} & {\scriptsize{}Rcll $\uparrow$}\tabularnewline
\hline 
\multirow{2}{*}{{\scriptsize{}Cam1}} & {\scriptsize{}YOLOv3} & {\scriptsize{}20.6\%} & {\scriptsize{}80.2\%} & {\scriptsize{}0.56} & {\scriptsize{}0.97}\tabularnewline
 & {\scriptsize{}F-RCNN(VGG16)} & {\scriptsize{}12.0\%} & {\scriptsize{}80.3\%} & {\scriptsize{}0.53} & {\scriptsize{}0.97}\tabularnewline
\hline 
\multirow{2}{*}{{\scriptsize{}Cam2}} & {\scriptsize{}YOLOv3} & {\scriptsize{}20.5\%} & {\scriptsize{}78.8\%} & {\scriptsize{}0.56} & {\scriptsize{}0.97}\tabularnewline
 & {\scriptsize{}F-RCNN(VGG16)} & {\scriptsize{}12.0\%} & {\scriptsize{}80.1\%} & {\scriptsize{}0.53} & {\scriptsize{}0.98}\tabularnewline
\hline 
\multirow{2}{*}{{\scriptsize{}Cam3}} & {\scriptsize{}YOLOv3} & {\scriptsize{}13.2\%} & {\scriptsize{}79.7\%} & {\scriptsize{}0.53} & {\scriptsize{}0.97}\tabularnewline
 & {\scriptsize{}F-RCNN(VGG16)} & {\scriptsize{}10.1\%} & {\scriptsize{}80.8\%} & {\scriptsize{}0.51} & {\scriptsize{}0.97}\tabularnewline
\hline 
\multirow{2}{*}{{\scriptsize{}Cam4}} & {\scriptsize{}YOLOv3} & {\scriptsize{}12.1\%} & {\scriptsize{}79.7\%} & {\scriptsize{}0.51} & {\scriptsize{}0.96}\tabularnewline
 & {\scriptsize{}F-RCNN(VGG16)} & {\scriptsize{}11.1\%} & {\scriptsize{}80.3\%} & {\scriptsize{}0.41} & {\scriptsize{}0.96}\tabularnewline
\hline 
\end{tabular}{\scriptsize\par}

{\scriptsize{}}%
\begin{tabular}{|c|c|c|c|c|c|}
\hline 
{\scriptsize{}CMC2} & {\scriptsize{}Detector} & {\scriptsize{}MODA $\uparrow$} & {\scriptsize{}MODP $\uparrow$} & {\scriptsize{}Prcn $\uparrow$} & {\scriptsize{}Rcll $\uparrow$}\tabularnewline
\hline 
\multirow{2}{*}{{\scriptsize{}Cam1}} & {\scriptsize{}YOLOv3} & {\scriptsize{}51.2\%} & {\scriptsize{}76.2\%} & {\scriptsize{}0.77} & {\scriptsize{}0.72}\tabularnewline
 & {\scriptsize{}F-RCNN(VGG16)} & {\scriptsize{}37.5\%} & {\scriptsize{}76.5\%} & {\scriptsize{}0.67} & {\scriptsize{}0.73}\tabularnewline
\hline 
\multirow{2}{*}{{\scriptsize{}Cam2}} & {\scriptsize{}YOLOv3} & {\scriptsize{}45.3\%} & {\scriptsize{}76.5\%} & {\scriptsize{}0.72} & {\scriptsize{}0.72}\tabularnewline
 & {\scriptsize{}F-RCNN(VGG16)} & {\scriptsize{}35.5\%} & {\scriptsize{}76.6\%} & {\scriptsize{}0.66} & {\scriptsize{}0.73}\tabularnewline
\hline 
\multirow{2}{*}{{\scriptsize{}Cam3}} & {\scriptsize{}YOLOv3} & {\scriptsize{}43.4\%} & {\scriptsize{}77.2\%} & {\scriptsize{}0.71} & {\scriptsize{}0.72}\tabularnewline
 & {\scriptsize{}F-RCNN(VGG16)} & {\scriptsize{}34.4\%} & {\scriptsize{}77.2\%} & {\scriptsize{}0.66} & {\scriptsize{}0.72}\tabularnewline
\hline 
\multirow{2}{*}{{\scriptsize{}Cam4}} & {\scriptsize{}YOLOv3} & {\scriptsize{}47.3\%} & {\scriptsize{}77.7\%} & {\scriptsize{}0.74} & {\scriptsize{}0.71}\tabularnewline
 & {\scriptsize{}F-RCNN(VGG16)} & {\scriptsize{}37.4\%} & {\scriptsize{}78.0\%} & {\scriptsize{}0.67} & {\scriptsize{}0.72}\tabularnewline
\hline 
\end{tabular}{\scriptsize\par}

{\scriptsize{}}%
\begin{tabular}{|c|c|c|c|c|c|}
\hline 
{\scriptsize{}CMC3} & {\scriptsize{}Detector} & {\scriptsize{}MODA $\uparrow$} & {\scriptsize{}MODP $\uparrow$} & {\scriptsize{}Prcn $\uparrow$} & {\scriptsize{}Rcll $\uparrow$}\tabularnewline
\hline 
\multirow{2}{*}{{\scriptsize{}Cam1}} & {\scriptsize{}YOLOv3} & {\scriptsize{}44.9\%} & {\scriptsize{}76.4\%} & {\scriptsize{}0.79} & {\scriptsize{}0.60}\tabularnewline
 & {\scriptsize{}F-RCNN(VGG16)} & {\scriptsize{}33.1\%} & {\scriptsize{}76.0\%} & {\scriptsize{}0.67} & {\scriptsize{}0.61}\tabularnewline
\hline 
\multirow{2}{*}{{\scriptsize{}Cam2}} & {\scriptsize{}YOLOv3} & {\scriptsize{}39.8\%} & {\scriptsize{}75.3\%} & {\scriptsize{}0.73} & {\scriptsize{}0.62}\tabularnewline
 & {\scriptsize{}F-RCNN(VGG16)} & {\scriptsize{}30.9\%} & {\scriptsize{}75.4\%} & {\scriptsize{}0.66} & {\scriptsize{}0.63}\tabularnewline
\hline 
\multirow{2}{*}{{\scriptsize{}Cam3}} & {\scriptsize{}YOLOv3} & {\scriptsize{}36.1\%} & {\scriptsize{}74.4\%} & {\scriptsize{}0.72} & {\scriptsize{}0.58}\tabularnewline
 & {\scriptsize{}F-RCNN(VGG16)} & {\scriptsize{}29.6\%} & {\scriptsize{}74.0\%} & {\scriptsize{}0.66} & {\scriptsize{}0.61}\tabularnewline
\hline 
\multirow{2}{*}{{\scriptsize{}Cam4}} & {\scriptsize{}YOLOv3} & {\scriptsize{}37.0\%} & {\scriptsize{}74.9\%} & {\scriptsize{}0.72} & {\scriptsize{}0.59}\tabularnewline
 & {\scriptsize{}F-RCNN(VGG16)} & {\scriptsize{}27.6\%} & {\scriptsize{}74.6\%} & {\scriptsize{}0.65} & {\scriptsize{}0.60}\tabularnewline
\hline 
\end{tabular}{\scriptsize\par}

{\scriptsize{}}%
\begin{tabular}{|c|c|c|c|c|c|}
\hline 
{\scriptsize{}CMC4} & {\scriptsize{}Detector} & {\scriptsize{}MODA $\uparrow$} & {\scriptsize{}MODP $\uparrow$} & {\scriptsize{}Prcn $\uparrow$} & {\scriptsize{}Rcll $\uparrow$}\tabularnewline
\hline 
\multirow{2}{*}{{\scriptsize{}Cam1}} & {\scriptsize{}YOLOv3} & {\scriptsize{}86.8\%} & {\scriptsize{}82.0\%} & {\scriptsize{}0.93} & {\scriptsize{}0.93}\tabularnewline
 & {\scriptsize{}F-RCNN(VGG16)} & {\scriptsize{}76.7\%} & {\scriptsize{}82.6\%} & {\scriptsize{}0.84} & {\scriptsize{}0.94}\tabularnewline
\hline 
\multirow{2}{*}{{\scriptsize{}Cam2}} & {\scriptsize{}YOLOv3} & {\scriptsize{}75.2\%} & {\scriptsize{}79.1\%} & {\scriptsize{}0.87} & {\scriptsize{}0.88}\tabularnewline
 & {\scriptsize{}F-RCNN(VGG16)} & {\scriptsize{}68.3\%} & {\scriptsize{}80.3\%} & {\scriptsize{}0.82} & {\scriptsize{}0.88}\tabularnewline
\hline 
\multirow{2}{*}{{\scriptsize{}Cam3}} & {\scriptsize{}YOLOv3} & {\scriptsize{}86.7\%} & {\scriptsize{}84.6\%} & {\scriptsize{}0.93} & {\scriptsize{}0.93}\tabularnewline
 & {\scriptsize{}F-RCNN(VGG16)} & {\scriptsize{}77.3\%} & {\scriptsize{}87.0\%} & {\scriptsize{}0.84} & {\scriptsize{}0.95}\tabularnewline
\hline 
\multirow{2}{*}{{\scriptsize{}Cam4}} & {\scriptsize{}YOLOv3} & {\scriptsize{}81.5\%} & {\scriptsize{}82.7\%} & {\scriptsize{}0.94} & {\scriptsize{}0.87}\tabularnewline
 & {\scriptsize{}F-RCNN(VGG16)} & {\scriptsize{}75.9\%} & {\scriptsize{}82.2\%} & {\scriptsize{}0.82} & {\scriptsize{}0.97}\tabularnewline
\hline 
\end{tabular}{\scriptsize\par}

{\scriptsize{}}%
\begin{tabular}{|c|c|c|c|c|c|}
\hline 
{\scriptsize{}CMC5} & {\scriptsize{}Detector} & {\scriptsize{}MODA $\uparrow$} & {\scriptsize{}MODP $\uparrow$} & {\scriptsize{}Prcn $\uparrow$} & {\scriptsize{}Rcll $\uparrow$}\tabularnewline
\hline 
\multirow{2}{*}{{\scriptsize{}Cam1}} & {\scriptsize{}YOLOv3} & {\scriptsize{}48.7\%} & {\scriptsize{}75.1\%} & {\scriptsize{}0.77} & {\scriptsize{}0.68}\tabularnewline
 & {\scriptsize{}F-RCNN(VGG16)} & {\scriptsize{}50.3\%} & {\scriptsize{}74.8\%} & {\scriptsize{}0.71} & {\scriptsize{}0.69}\tabularnewline
\hline 
\multirow{2}{*}{{\scriptsize{}Cam2}} & {\scriptsize{}YOLOv3} & {\scriptsize{}49.8\%} & {\scriptsize{}75.6\%} & {\scriptsize{}0.66} & {\scriptsize{}0.65}\tabularnewline
 & {\scriptsize{}F-RCNN(VGG16)} & {\scriptsize{}45.3\%} & {\scriptsize{}76.4\%} & {\scriptsize{}0.67} & {\scriptsize{}0.61}\tabularnewline
\hline 
\multirow{2}{*}{{\scriptsize{}Cam3}} & {\scriptsize{}YOLOv3} & {\scriptsize{}50.7\%} & {\scriptsize{}73.1\%} & {\scriptsize{}0.65} & {\scriptsize{}0.66}\tabularnewline
 & {\scriptsize{}F-RCNN(VGG16)} & {\scriptsize{}44.7\%} & {\scriptsize{}74.3\%} & {\scriptsize{}0.65} & {\scriptsize{}0.65}\tabularnewline
\hline 
\multirow{2}{*}{{\scriptsize{}Cam4}} & {\scriptsize{}YOLOv3} & {\scriptsize{}49.8\%} & {\scriptsize{}76.2\%} & {\scriptsize{}0.65} & {\scriptsize{}0.68}\tabularnewline
 & {\scriptsize{}F-RCNN(VGG16)} & {\scriptsize{}46.7\%} & {\scriptsize{}74.1\%} & {\scriptsize{}0.61} & {\scriptsize{}0.69}\tabularnewline
\hline 
\end{tabular}{\scriptsize\par}

\label{CMC_CLEARMOT_Det}
\end{table}

\noindent %

\end{document}